\documentclass{article}

\usepackage{microtype}
\usepackage{graphicx}
\usepackage{subfigure}
\usepackage{booktabs} % for professional tables
\usepackage{dsfont}
\usepackage{hyperref}

\usepackage[accepted]{icml2025} 

\usepackage{amsmath}
\usepackage{amssymb}
\usepackage{mathtools}
\usepackage{amsthm}

\usepackage{adjustbox}
\usepackage{array}
\usepackage{longtable}

\usepackage{algorithm}
\usepackage{algorithmic}

\usepackage[capitalize,noabbrev]{cleveref}
\usepackage{fontawesome5} % or another icon package

\theoremstyle{plain}

\theoremstyle{definition}

\theoremstyle{remark}

\usepackage[textsize=tiny]{todonotes}

\icmltitlerunning{Scaling Test-Time Inference with Policy-Optimized, Dynamic Retrieval-Augmented Generation via KV Caching and Decoding}

\begin{document}

\twocolumn[
\icmltitle{Scaling Test-Time Inference with Policy-Optimized, Dynamic Retrieval-Augmented Generation via KV Caching and Decoding\\
           }




\begin{icmlauthorlist}
\icmlauthor{Sakhinana Sagar Srinivas}{comp}
\icmlauthor{Akash Das}{comp}
\icmlauthor{Shivam Gupta}{comp}
\icmlauthor{Venkataramana Runkana}{comp}
\end{icmlauthorlist}

\icmlaffiliation{comp}{Tata Research Development and Design Center, Bangalore}

\icmlcorrespondingauthor{Sakhinana Sagar Srinivas}{sagar.sakhinana@tcs.com}

\icmlkeywords{Machine Learning, ICML}

\vskip 0.3in
]



\printAffiliationsAndNotice{}  % leave blank if no need to mention equal contribution

\begin{abstract}
\vspace{-1mm}
We present a comprehensive framework for enhancing Retrieval-Augmented Generation (RAG) systems through dynamic retrieval strategies and reinforcement fine-tuning. This approach significantly improves large language models on knowledge-intensive tasks, including open-domain question answering and complex reasoning. Our framework integrates two complementary techniques: Policy-Optimized Retrieval-Augmented Generation (PORAG), which optimizes the use of retrieved information, and Adaptive Token-Layer Attention Scoring (ATLAS), which dynamically determines retrieval timing and content based on contextual needs. Together, these techniques enhance both the utilization and relevance of retrieved content, improving factual accuracy and response quality. Designed as a lightweight solution compatible with any Transformer-based LLM without requiring additional training, our framework excels in knowledge-intensive tasks, boosting output accuracy in RAG settings. We further propose CRITIC, a novel method to selectively compress key-value caches by token importance, mitigating memory bottlenecks in long-context applications. The framework also incorporates test-time scaling techniques to dynamically balance reasoning depth and computational resources, alongside optimized decoding strategies for faster inference. Experiments on benchmark datasets show that our framework reduces hallucinations, strengthens domain-specific reasoning, and achieves significant efficiency and scalability gains over traditional RAG systems. This integrated approach advances the development of robust, efficient, and scalable RAG systems across diverse applications.
\vspace{-7mm}
\end{abstract}

\section{Introduction}
\vspace{-2mm}
Retrieval-Augmented Generation (RAG, \cite{lewis2020retrieval, su2403dragin, wang2025chain}) has gained significant interest in Natural Language Processing for enhancing large language models (LLMs) on knowledge-intensive tasks through external information retrieval, with applications across search engines, conversational agents, chatbots, and many other applications. RAG addresses key LLM limitations, including hallucinations, outdated information, and insufficient domain-specific knowledge, particularly in open-domain question answering. Retrieval-Augmented Fine-Tuning (RAFT \cite{zhang2024raft}) advances this approach by integrating retrieval methods with language model supervised fine-tuning. Unlike traditional RAG, which simply retrieves documents for generation, RAFT trains the language model alongside the retrieval mechanism, teaching it to dynamically leverage external knowledge, prioritize relevant content while ignoring distractors for improved performance in domain-specific RAG contexts (e.g., open-book and in-domain question answering). Building on advancements in LLM training methodologies, DeepSeek has enhanced its AI models, notably DeepSeek-R1 \cite{liu2024deepseek, guo2025deepseek, shao2024deepseekmath}, by implementing Group Relative Policy Optimization (GRPO), an advanced reinforcement learning algorithm that improves training efficiency and model performance beyond traditional supervised fine-tuning. GRPO reduces computational overhead by eliminating the value function, using group-based advantage estimation for simplified reward computation, lowering memory usage, and integrating Kullback-Leibler (KL) divergence regularization for stable, efficient training. It outperforms standard Rejection Sampling Fine-Tuning (RFT), which relies on offline sampling, and Online RFT, which dynamically samples from an evolving policy. GRPO also supports process supervision (GRPO+PS), providing step-by-step feedback for improved reasoning, surpassing outcome supervision (GRPO+OS), which evaluates only final answers. Addressing the limitations of static retrieval in traditional RAG, DRAGIN (Dynamic Retrieval-Augmented Generation based on Information Needs, \cite{su2403dragin}) is an advanced framework that dynamically determines when and what to retrieve during text generation. Unlike methods with fixed retrieval intervals or simplistic query formulations, DRAGIN employs Real-time Information Needs Detection (RIND) to trigger retrieval only when necessary, considering token uncertainty, semantic importance, and influence on future tokens. Its query formulation based on Self-attention (QFS) generates more effective queries by leveraging the full generated context rather than just recent tokens to fill information gaps. This adaptive approach minimizes redundant retrievals, improves efficiency, and enhances response accuracy. Despite these advancements, integrating external knowledge during inference through RAG enhances the capabilities of LLMs. However, it also introduces challenges, such as increased computational and memory demands. Key-Value (KV) Caching \cite{feng2024ada, hooper2025kvquant, yang2025kvlink} addresses this issue by efficiently managing the memory load resulting from RAG’s expanded context window. It optimizes the storage and retrieval of key-value pairs, preventing memory bottlenecks and accelerating the processing of augmented information. In transformer-based LLMs, KV Caching stores intermediate hidden states (keys and values) of previous tokens during attention computation, enabling faster text generation by reusing them for new tokens. This approach reduces redundant calculations, lowers memory usage, and improves efficiency for long sequences, thereby enhancing the contextuality and coherence of LLMs while mitigating the memory overhead introduced by RAG. Test-Time Scaling Inference Techniques \cite{muennighoff2025s1, ji2025test, yoon2025monte, geiping2025scaling} address these challenges by dynamically allocating computational resources based on task complexity. Unlike static inference methods, which apply fixed computational effort regardless of task demands, test-time scaling adaptively adjusts reasoning depth and complexity. For simple questions, it reduces unnecessary overhead, enabling faster responses and minimizing hallucinations. For complex or multi-faceted tasks, it increases reasoning depth to improve accuracy and better integrate retrieved context, enabling LLMs to effectively process and reason with augmented context. This adaptive approach mimics human-like deliberative reasoning for knowledge-intensive tasks without costly retraining, enhancing efficiency and performance while maintaining accuracy and reducing hallucinations.
Together, RAFT enhances RAG by integrating retrieval with supervised fine-tuning, enabling models to dynamically leverage external knowledge and prioritize relevant content while ignoring distractors. DRAGIN dynamically determines when and what to retrieve during text generation, minimizing redundant retrievals and improving efficiency. KV Caching optimizes memory usage by storing intermediate hidden states, reducing computational overhead in RAG, while Test-Time Scaling dynamically allocates resources based on task complexity. These advancements enable RAG systems to integrate external knowledge more accurately, efficiently, and at scale, ensuring faster and more effective utilization of retrieved data within the LLM framework. While these recent advancements have enhanced retrieval integration in LLMs, significant challenges remain in balancing retrieval fidelity, response quality, and computational efficiency. Current methods often struggle to dynamically determine when and how much external information to incorporate, sometimes overwhelming the model or sacrificing the coherence of its responses. Motivated by these persistent challenges, our work seeks to refine the synergy between retrieval and generation through a dual approach. First, we fine-tune language models via policy optimization, enabling them to more effectively integrate and utilize retrieved content. This refinement not only improves factual alignment but also enhances overall response quality. Second, we introduce a mechanism that selectively triggers external retrieval based on the model’s internal state, ensuring that additional information is incorporated only when necessary. This targeted strategy optimizes computational resources while preserving the language model’s coherence. In the following sections, we outline our contributions that extend state-of-the-art methods by addressing both the optimization of retrieval-augmented generation and the efficient management of computational overhead. Our contributions are as follows:

\begin{itemize}
    \item We introduce two complementary techniques to enhance Retrieval-Augmented Generation (RAG) systems: Policy-Optimized Retrieval-Augmented Generation (PORAG) and Adaptive Token-Layer Attention Scoring for Selective Retrieval (ATLAS). PORAG extends GRPO to the RAG setting, fine-tuning pre-trained LLMs using QLoRA (Quantized Low-Rank Adaptation). The parameter-efficient optimization using QLoRA leads to improved performance on in-domain Question-Answering (QA) tasks while mitigating catastrophic forgetting of pre-trained knowledge. PORAG incorporates group-based advantage estimation and a trust-region constrained policy update to ensure stable and robust fine-tuning in retrieval-dependent contexts. Additionally, PORAG employs a dual reward mechanism that explicitly balances retrieval fidelity—ensuring generated responses remain factually aligned with retrieved information—and response quality, which evaluates coherence, fluency, and overall helpfulness beyond factual accuracy. To effectively implement this, specialized linear layer-based reward heads are integrated after the final layer of the pre-trained LLM with QLoRA adapters. Trained reward heads evaluate retrieval fidelity and response quality, and their combined signals form a composite reward for group-based advantage estimation, thus guiding generation policy optimization. ATLAS, on the other hand, dynamically determines when and what to retrieve by analyzing the language model's internal attention patterns. Using Multi-Layer Attention Gradient (MLAG) to detect information gaps and Layerwise Representation Pooling (LRP) to construct targeted queries, ATLAS retrieves the most relevant external information to fill information gaps, improving retrieval precision and ensuring retrieval occurs only when necessary and precisely aligned with the model's information needs. Together, these techniques create a comprehensive RAG system that optimizes both the utilization of retrieved information and the timing of retrieval, significantly improving efficiency, accuracy, and computational overhead. The integration of PORAG and ATLAS addresses key challenges in RAG systems, such as over-reliance on retrieval, inefficient query formulation, and unstable optimization, paving the way for more robust and resource-efficient language models.   
    \item  We present CRITIC (Cache Reduction via Importance-based Token Inclusion Criteria), a method that addresses the memory bottleneck in policy-optimized LLMs inference by selectively retaining only the most important tokens in the KV cache. While traditional KV caching already reduces computational cost from quadratic to linear, memory usage still grows proportionally with sequence length, creating limitations for long-context RAG applications. CRITIC determines token importance using a weighted hybrid approach that combines three complementary strategies: attention-based (relationship strength), entropy-based (attention pattern complexity), and gradient-based (prediction sensitivity). This integrated approach enables flexible compression behavior, with the framework preserving only the highest-scoring tokens based on a configurable ratio. To further enhance real-world applicability, CRITIC incorporates features such as delayed compression activation and memory-pressure-based adaptive ratios as practical optimizations. The architecture-agnostic solution significantly reduces memory requirements while maintaining performance, leading to faster inference and the ability to process longer contexts, particularly benefiting RAG applications that need extended context windows.   
    \item We study the test-time scaling inference performance of policy-optimized LLMs in RAG contexts, focusing on improving response quality without altering model weights by dynamically adjusting reasoning depth, sampling, and validation during inference. We utilize well-known inference scaling techniques, including Self-Consistency, Best-of-N Sampling, Monte Carlo Tree Search (MCTS), and others, each employing unique strategies to enhance output quality, accuracy, and efficiency. These methods trade off increased computational complexity—often exceeding \(O(n)\) for standard inference, where \(n\) is the sequence length—for improved reliability and response quality, optimizing inference under resource constraints. Many of these techniques leverage Weak-to-Strong Distillation, iteratively refining outputs to converge on higher-quality responses. Each algorithm presents distinct trade-offs in cost, approach, selection method, and other key factors.
    \vspace{-3mm}
\end{itemize}

\section{Proposed Methodology}
Current Retrieval-Augmented Generation (RAG) systems face limitations in their optimization approaches, particularly with log-likelihood-based methods like RAFT. To address these constraints, we introduce two complementary innovations: Policy-Optimized Retrieval-Augmented Generation (PORAG) and Adaptive Token-Layer Attention Scoring for Selective Retrieval (ATLAS). Together, these components create a more robust framework that simultaneously optimizes generation quality and retrieval efficiency. PORAG fundamentally reimagines RAG optimization through a reinforcement learning paradigm built on Group Relative Policy Optimization (GRPO). This approach overcomes RAFT's limitations by moving beyond static reference outputs and undifferentiated treatment of retrieved documents. The system's group-based advantage estimation enables comparative evaluation of multiple candidate generations for each query-retrieval pair. At its core, PORAG implements a dual reward mechanism with two specialized components: (1) a retrieval fidelity reward head that precisely measures how well generated outputs reflect the retrieved evidence, and (2) a response quality reward head that assesses broader linguistic properties including coherence, fluency, and task-aligned helpfulness. These reward signals are optimized jointly with the policy through a carefully designed objective function combining clipped surrogate rewards with KL divergence regularization. This formulation ensures stable training while maintaining the model's generative capabilities. Crucially, PORAG maintains inference-time efficiency through single-shot decoding, avoiding the computational overhead of multi-candidate sampling while preserving the speed of standard autoregressive generation. ATLAS complements this approach with a sophisticated, introspection-based retrieval mechanism operating through two coordinated stages. The first stage employs Multi-Layer Attention Gradient (MLAG) analysis to dynamically detect information gaps. By monitoring shifts in attention distributions across transformer layers and weighting these signals with both token-level uncertainty measures and entropy-normalized attention head importance, the system precisely identifies when retrieval is truly necessary. The second stage implements Layerwise Representation Pooling (LRP) to determine optimal query content. This process evaluates preceding tokens through a hybrid scoring system that combines attention-based salience metrics with deep semantic similarity measures in the model's internal representations. The highest-scoring tokens are then processed through a streamlined prompt template to generate focused, context-aware retrieval queries that directly target the model's knowledge deficiencies. When integrated, PORAG and ATLAS form a comprehensive RAG framework that advances both generation quality and retrieval efficiency. PORAG's learned reward structure ensures outputs maintain high standards of factual accuracy and linguistic quality, while ATLAS's intelligent retrieval mechanism dramatically reduces computational overhead through precision targeting. This dual advancement produces a system that excels in factual reliability, response quality, and operational efficiency - particularly valuable for deployment in scenarios with strict latency or memory constraints. The combined approach represents a significant step forward in developing practical, high-performance RAG systems that maintain both accuracy and efficiency at scale.

\subsection{Policy-Optimized Retrieval-Augmented Generation (PORAG)}
\label{porag}
RAG techniques present unique optimization challenges that Retrieval-Augmented Fine-Tuning (RAFT) often struggles to fully address. PORAG offers a principled solution rooted in Group Relative Policy Optimization (GRPO) by reformulating the optimization problem through a group-based relative advantage framework. Unlike RAFT, which optimizes for log-likelihood of reference outputs, PORAG enables direct optimization for retrieval quality, contextual relevance, and generation coherence through dual reward modeling. In this work, we present a comprehensive mathematical formulation of PORAG, with theoretical justifications and analytical insights. In the traditional RAG framework, the policy model \(\pi_\theta(y|x,d)\) generates outputs \(y\) conditioned on the input query \(x\) and retrieved documents \(d\). The process is formalized as:

\vspace{-1mm}
\resizebox{0.985\linewidth}{!}{
\begin{minipage}{\linewidth}
\begin{equation}
\pi_\theta(y|x,d) = \prod_{i=1}^{|y|} \pi_\theta(y_i|x, d, y_{<i})
\end{equation}
\end{minipage}
}

\noindent
where \(\pi_\theta(y|x,d)\) represents the probability distribution over the generated outputs \(y\), conditioned on the input query \(x\), retrieved documents \(d\), and previously generated tokens \(y_{<i}\). Here, \(x\) denotes the input query, \(d = \{d_1, d_2, ..., d_k\}\) represents the set of retrieved documents, \(y_i\) is the token at position \(i\), and \(y_{<i}\) comprises all previously generated tokens. The parameter \(\theta\) corresponds to the frozen weights of the language model, which remain unchanged during inference. In RAFT, the training objective optimizes the pretrained language model by maximizing the likelihood of reference outputs \( y^* \) while incorporating both relevant (``oracle") and irrelevant (``distractor") documents. Since RAFT employs Low-Rank Adaptation (LoRA\cite{lewis2020retrieval, izacard2020leveraging}), only a subset of trainable parameters, denoted as \( \gamma \), is updated, while the pre-trained language model parameters \( \theta \) remain frozen. The RAFT loss function is defined as:

\vspace{-1mm}
\resizebox{1\linewidth}{!}{
\begin{minipage}{\linewidth}
\begin{equation}
\begin{split}
\mathcal{L}_{\text{RAFT}}(\gamma) &= -\mathbb{E}_{(x, d_{\text{oracle}}, d_{\text{distractor}}, y^*) \sim \mathcal{D}} \\
& \quad \left[ \log \pi_{\theta, \gamma}(y^*|x, d_{\text{oracle}}, d_{\text{distractor}}) \right]
\end{split}
\end{equation}
\end{minipage}
}

\noindent
where \( x \) is the input query, \( d_{\text{oracle}} \) and \( d_{\text{distractor}} \) represent the retrieved relevant and irrelevant documents, respectively, and \( y^* \) is the reference output. The training dataset \( \mathcal{D} \) consists of tuples \( (x, d_{\text{oracle}}, d_{\text{distractor}}, y^*) \). The model assigns probability \( \pi_{\theta, \gamma}(y^*|x, d_{\text{oracle}}, d_{\text{distractor}}) \) to the correct output, where \( \theta \) represents the frozen pre-trained language model parameters, and 
\( \gamma \) represents the trainable parameters of the base language model, specifically Quantized Low-Rank Adaptation (QLoRA) adapters. These are small, trainable low-rank matrices added to the frozen pre-trained language model (\(\theta\)) to govern output generation conditioned on the input and retrieved documents. QLoRA focuses on adapting key layers like attention query/value projections and feed-forward networks. This approach enables efficient fine-tuning by modifying only a small subset of weights, ensuring that the model learns to effectively distinguish relevant information from distractors while leveraging retrieval-augmented generation for adaptation. However, RAFT has several limitations. It cannot differentiate between high- and low-quality retrievals, assumes perfect reference outputs that fully leverage retrieved information, and does not account for multiple valid generation strategies within the same retrieval context. Additionally, it fails to optimize nuanced qualities such as faithfulness to retrieved information. In contrast, PORAG addresses these limitations by enabling direct optimization for multiple quality dimensions simultaneously. Our implementation employs two specialized reward heads—lightweight, parameterized functions attached to the base model’s hidden states—calibrated for RAG-specific quality dimensions: a \text{Retrieval-Fidelity Reward} \( R_{\text{fidelity}}(x, d, y^*; \phi_1) \), which evaluates how faithfully the generated response incorporates and accurately reflects the retrieved information, and a \text{Response-Quality Reward} \( R_{\text{quality}}(x, d, y^*; \phi_2) \), which evaluates the overall quality, coherence, and helpfulness of the response beyond mere factual accuracy. Here, \( \phi = \{\phi_1, \phi_2\} \) represent the trainable reward head parameters. The two reward heads—$\phi_1$ for retrieval fidelity and $\phi_2$ for response quality—are integrated into the neural network architecture at the final layer, operating on the hidden representations produced by the base model to compute scalar rewards. Parameters \( \phi_1 \) and \( \phi_2 \) (typically implemented via trainable standard linear layers with an intermediate \( \tanh \) activation) are specifically optimized to evaluate how well the generated response meets the desired qualities (i.e., factual alignment with the retrieved documents and overall quality). The reward heads are trained in conjunction with the base model, facilitating end-to-end optimization of both the generation and the reward function estimation. Consequently, the generation policy is directly informed by these dynamically learned reward signals. This co-adaptation mechanism results in more precise reward evaluations, enhanced training stability, and ultimately, superior performance in RAG. To effectively optimize the RAG context for multiple objectives, we decompose the utility function into orthogonal components, each capturing distinct quality dimensions. This allows the reward heads to focus on specific aspects of generation quality. The utility function is defined as:

\vspace{-2mm}
\resizebox{0.985\linewidth}{!}{
\begin{minipage}{\linewidth}
\begin{equation}
\begin{split}
\mathcal{U}(x,d,y^*) &= \alpha \cdot \mathcal{U}_{\text{fidelity}}(x,d,y^*) + \beta \cdot \mathcal{U}_{\text{quality}}(x,y^*) \\
&\quad + \lambda \cdot \mathcal{U}_{\text{interaction}}(x,d,y^*)  \nonumber
\end{split}
\end{equation}
\end{minipage}
}

\noindent
where: \( \mathcal{U}_{\text{fidelity}}(x,d,y^*) \) measures the accuracy of the generated text in reflecting the retrieved documents, rewarding correct factual content and penalizing hallucinations;  \( \mathcal{U}_{\text{quality}}(x,y^*) \) evaluates the inherent quality of the generation (coherence, fluency, relevance to the query), independent of the retrieved content; and \( \mathcal{U}_{\text{interaction}}(x,d,y^*) \) captures the synergistic effects between fidelity and quality. Our dual reward heads approximate this decomposition:

\vspace{-2mm}
\resizebox{0.985\linewidth}{!}{
\begin{minipage}{\linewidth}
\begin{equation}
R_{\text{fidelity}}(x, d, y^*; \phi_1) \approx \mathcal{U}_{\text{fidelity}}(x,d,y^*)  \nonumber
\end{equation}
\end{minipage}
}

\vspace{-2mm}
\resizebox{0.985\linewidth}{!}{
\begin{minipage}{\linewidth}
\begin{equation}
\begin{split}
R_{\text{quality}}(x, d, y^*; \phi_2) &\approx \mathcal{U}_{\text{quality}}(x,y^*) \\
&\quad + \frac{\lambda}{\beta} \cdot \mathcal{U}_{\text{interaction}}(x,d,y^*) \nonumber
\end{split}
\end{equation}
\end{minipage}
}

\noindent
The reward heads compute scalar rewards from a vector representation derived from the hidden states of the base model through parameterized transformation functions:

\vspace{-2mm}
\resizebox{0.985\linewidth}{!}{
\begin{minipage}{\linewidth}
\begin{equation}
R_{\text{fidelity}}(x, d, y^*; \phi_1) = f_{\phi_1}(h(x, d, y^*))  \nonumber
\end{equation}
\end{minipage}
}

\vspace{-2mm}
\resizebox{0.985\linewidth}{!}{
\begin{minipage}{\linewidth}
\begin{equation}
R_{\text{quality}}(x, d, y^*; \phi_2) = g_{\phi_2}(h(x, d, y^*))  \nonumber
\end{equation}
\end{minipage}
}

\noindent
where \( h(x, d, y^*) \in \mathbb{R}^{d} \) is a vector derived from the base language model's hidden states. Transformer models output a hidden state matrix \( \mathbb{R}^{n \times d} \) (where \(n\) is sequence length, \(d\) is hidden dimension).  \(h\) is obtained by aggregating this matrix, e.g., using the last token's state or pooling. The reward heads \(R_{\text{fidelity}} = f_{\phi_1}(h)\) and \(R_{\text{quality}} = f_{\phi_2}(h)\) are both multi-layer perceptrons with the form:

\vspace{-2mm}
\resizebox{0.985\linewidth}{!}{
\begin{minipage}{\linewidth}
\begin{align*}
f_{\phi_i}(h) &= W_2^{\phi_i} \cdot \tanh(W_1^{\phi_i} \cdot h + b_1^{\phi_i}) + b_2^{\phi_i}
\end{align*}
\end{minipage}
}

where for \(i \in \{1, 2\}\), \( W_1^{\phi_i} \in \mathbb{R}^{d \times d} \), \( W_2^{\phi_i} \in \mathbb{R}^{d \times 1} \), \( b_1^{\phi_i} \in \mathbb{R}^d \), and \( b_2^{\phi_i} \in \mathbb{R} \) are the parameters for reward head \(i\). We calculate the combined reward by balancing the competing objectives of retrieval fidelity and response quality. Specifically, we aggregate quality and fidelity rewards as follows:

\vspace{-2mm}
\resizebox{0.985\linewidth}{!}{
\begin{minipage}{\linewidth}
\begin{equation}
\begin{split}
R_{\text{comb}}(x, d, y^*) &= \alpha \cdot R_{\text{fidelity}}(x, d, y^*; \phi_1) \\
&\quad + \beta \cdot R_{\text{quality}}(x, d, y^*; \phi_2)  \nonumber
\end{split}
\end{equation}
\end{minipage}
}

\noindent
This weighting scheme (\( \alpha = 0.7 \) and \( \beta = 0.3 \) in our implementation) balances the competing objectives of retrieval fidelity and response quality. The theoretical justification for this weighting comes from multi-objective reinforcement learning theory, where the Pareto frontier of optimal policies can be explored through different weightings of reward components. Unlike RAFT, which implicitly weights these objectives based on the training data distribution alone, PORAG allows explicit control over this trade-off, enabling adaptation to different deployment scenarios and user preferences. The combined rewards are normalized and scaled using robust statistical principles:

\vspace{-2mm}
\resizebox{0.985\linewidth}{!}{
\begin{minipage}{\linewidth}
\begin{equation}
\begin{split}
R_{\text{final}}(x, d, y^*) &= \text{clip}(R_{\text{comb}}(x, d, y^*), -c_1, c_1) \hspace{-3mm} \quad \cdot \gamma_{\text{scale}}
\end{split}
\nonumber
\end{equation}
\vspace{-10mm}
\end{minipage}
}

where \( \gamma_{\text{scale}} \) is the reward scaling factor, and \( c_1 = 10.0 \) is the clipping threshold. The clipping operation is a form of Winsorization, a statistical technique that reduces the impact of outliers while preserving the ordinal relationships between rewards. We will now discuss Group-based Advantage Estimation for RAG. Given an input query \( x \) and retrieved documents \( d \), we generate a batch of \( G \) outputs, denoted by \( \{y^{(1)}, y^{(2)}, \ldots, y^{(G)}\} \), using the current policy \( \pi_\gamma \). This batch of outputs represents a single group of alternatives. Within this group, we compute robust statistical estimators based on the final reward \( R_{\text{final}}(x, d, y^{(i)}) \), which represents the overall reward for the \( i \)-th output \( y^{(i)} \) within that group, given the input query \( x \) and retrieved documents \( d \):

\vspace{-1mm}
\resizebox{0.985\linewidth}{!}{
\begin{minipage}{\linewidth}
\begin{equation}
\mu_R(x, d) = \frac{1}{G}\sum_{i=1}^{G} R_{\text{final}}(x, d, y^{(i)})
\end{equation}
\end{minipage}
}

\vspace{-1mm}
\resizebox{0.985\linewidth}{!}{
\begin{minipage}{\linewidth}
\begin{equation}
\sigma_R^2(x, d) = \frac{1}{G}\sum_{i=1}^{G} \left(R_{\text{final}}(x, d, y^{(i)}) - \mu_R(x, d)\right)^2
\end{equation}
\end{minipage}
}

\vspace{-1mm}
\resizebox{0.985\linewidth}{!}{
\begin{minipage}{\linewidth}
\begin{equation}
\sigma_R(x, d) = \max\left(\sqrt{\sigma_R^2(x, d) + \epsilon}, \sigma_{\min}\right)
\end{equation}
\end{minipage}
}

\noindent
where \( \mu_R(x, d) \) is the mean reward calculated within the group, \( \sigma_R^2(x, d) \) is the variance of the rewards calculated within the group, and \( \sigma_R(x, d) \) is the standard deviation of the rewards calculated within the group, clipped below by a minimum value \( \sigma_{\min} = 0.1 \) to ensure numerical stability. The clipping prevents overly aggressive updates when reward variation is small, which is particularly important in RAG scenarios where retrieved documents might lead to very similar generations within the group. The group-relative advantage for each output \( y^{(i)} \) is then calculated as:

\vspace{-1mm}
\resizebox{0.985\linewidth}{!}{
\begin{minipage}{\linewidth}
\begin{equation}
\hat{A}_i = \frac{R_{\text{final}}(x, d, y^{(i)}) - \mu_R(x, d)}{\sigma_R(x, d)}
\end{equation}
\end{minipage}
}

\noindent
where \( \hat{A}_i \) represents the advantage of the \( i \)-th generated output relative to the other outputs within its group. We will now discuss the GRPO objective function for RAG settings. For each token \( y_j^{(i)} \) in the RAG output \( y^{(i)} \), we compute the probability ratio:

\vspace{-1mm}
\resizebox{0.90\linewidth}{!}{
\begin{minipage}{\linewidth}
\begin{equation}
r_j(\gamma) = \frac{\pi(y_j^{(i)}|x, d, y_{<j}^{(i)})}{\pi_{\text{old}}(y_j^{(i)}|x, d, y_{<j}^{(i)})}
\end{equation}
\end{minipage}
}

\noindent
where the ratio \( r_j(\gamma) \) quantifies the change in token probability under the current policy relative to the policy that generated the sample, accounting for both the query and retrieved document context. The clipped surrogate objective with a policy constraint for RAG is:

\vspace{-1mm}
\resizebox{0.85\linewidth}{!}{
\begin{minipage}{\linewidth}
\begin{equation}
L_{\text{clip}}(\gamma) = \frac{1}{G} \sum_{i=1}^{G} \frac{1}{|y^{(i)}|} \sum_{j=1}^{|y^{(i)}|} \min\left(r_j(\gamma)\hat{A}_{i}, \text{clip}(r_j(\gamma), 1-\epsilon, 1+\epsilon)\hat{A}_{i}\right)  \nonumber
\end{equation}
\end{minipage}
}

\noindent
The clipping mechanism, with the parameter \( \epsilon = 0.2 \), serves as a trust region constraint that prevents excessively large policy updates; this is critical in RAG systems, where small changes in the probability distribution can lead to dramatically different retrieval utilization patterns. The KL divergence term prevents the policy from straying too far from the reference model:

\vspace{-1mm}
\resizebox{0.85\linewidth}{!}{
\begin{minipage}{\linewidth}
\begin{equation}
D_{\text{KL}}(\pi || \pi_{\text{ref}}) = \mathbb{E}_{x,d,y \sim \pi_\gamma}\left[\sum_{i=1}^{|y|} \text{KL}(\pi_{\text{ref}}(\cdot|x,d,y_{<i}) || \pi_\gamma(\cdot|x,d,y_{<i}))\right]  \nonumber
\end{equation}
\end{minipage}
}

\noindent
Here, \( \pi_{\text{ref}} \) represents the reference policy, specifically the policy from the previous iteration of training, denoted as \( \pi_{\gamma_{\text{old}}} \), where \( \gamma_{\text{old}} \) are the policy parameters before the current update.  Using the KL divergence with respect to the previous policy stabilizes training by preventing drastic changes in the policy distribution in each update step. In the RAG context, this regularization term serves a critical function: it preserves the base knowledge encoded in the model while allowing for targeted improvements in retrieval utilization. Without this constraint, aggressive optimization toward retrieval-grounded responses might cause the model to forget its pre-trained knowledge. Using the unbiased estimator:

\vspace{-1mm}
\resizebox{0.90\linewidth}{!}{
\begin{minipage}{\linewidth}
\begin{equation}
D_{\text{KL}}(\pi_\gamma || \pi_{\text{ref}}) = \mathbb{E}_{x,d,y \sim \pi_\gamma}\left[\frac{\pi_{\text{ref}}(y|x,d)}{\pi_\gamma(y|x,d)} - \log\frac{\pi_{\text{ref}}(y|x,d)}{\pi_\gamma(y|x,d)} - 1\right]  \nonumber
\end{equation}
\end{minipage}
}

\noindent
The complete GRPO objective for RAG optimization is:

\vspace{-1mm}
\resizebox{0.90\linewidth}{!}{
\begin{minipage}{\linewidth}
\begin{equation}
J_{\text{GRPO-RAG}}(\gamma) = \omega_1 \cdot L_{\text{clip}}(\gamma) - \omega_2 \cdot D_{\text{KL}}(\pi_\gamma || \pi_{\text{ref}})  \nonumber
\end{equation}
\end{minipage}
}

\noindent
where \( L_{\text{clip}}(\gamma) \) is the clipped surrogate objective that measures the policy improvement using the relative advantage estimates, and \( D_{\text{KL}}(\pi_\gamma || \pi_{\text{ref}}) \) is the KL divergence between the current policy \( \pi_\gamma \) and the reference policy \( \pi_{\text{ref}} \), acting as a regularizer. The weighting coefficients \( \omega_1 = 100.0 \) and \( \omega_2 = 0.1 \) balance policy improvement and divergence regularization; this balance is particularly important in RAG contexts to prevent overreliance on retrieved information at the expense of the model's pre-existing knowledge. The policy parameters \( \gamma \) are updated to maximize the GRPO-RAG objective:

\vspace{-1mm}
\resizebox{0.90\linewidth}{!}{
\begin{minipage}{\linewidth}
\begin{equation}
\gamma_{k+1} = \gamma_k + \eta_\gamma \nabla_\gamma J_{\text{GRPO-RAG}}(\gamma_k)
\end{equation}
\end{minipage}
}

\noindent
The learning rate \( \eta_\gamma \) (typically \( 1 \times 10^{-6} \) to \( 5 \times 10^{-6} \) for RAG optimization) controls the step size of each update. Unlike RAFT, which often uses larger learning rates, GRPO-RAG typically requires smaller steps due to the complexity of the reward landscape. To prevent instability in RAG optimization, gradients are regularized both by value and by norm:

\vspace{-1mm}
\resizebox{0.90\linewidth}{!}{
\begin{minipage}{\linewidth}
\begin{equation}
\nabla_\gamma J_{\text{clipped}} = \text{clip}(\nabla_\gamma J_{\text{GRPO-RAG}}(\gamma_k), -c_{\text{value}}, c_{\text{value}})
\end{equation}
\end{minipage}
}

\vspace{-1mm}
\resizebox{0.90\linewidth}{!}{
\begin{minipage}{\linewidth}
\begin{equation}
\nabla_\gamma J_{\text{normalized}} = \frac{\nabla_\gamma J_{\text{clipped}}}{||\nabla_\gamma J_{\text{clipped}}||_2} \cdot \min(||\nabla_\gamma J_{\text{clipped}}||_2, c_{\text{norm}})  \nonumber
\end{equation}
\end{minipage}
}

\noindent
The clipping thresholds \( c_{\text{value}} = 3.0 \) and \( c_{\text{norm}} = 1.0 \) prevent extreme gradient values that could destabilize training; this is especially important in RAG systems where the retrieval distribution can introduce high variance in gradients. The reward model parameters are updated using gradients derived from minimizing their respective reward loss functions, \( \mathcal{L}_{\text{fidelity}} \) and \( \mathcal{L}_{\text{quality}} \).

\vspace{-1mm}
\resizebox{0.90\linewidth}{!}{
\begin{minipage}{\linewidth}
\begin{equation}
\phi_{1,k+1} = \phi_{1,k} + \eta_R \nabla_{\phi_1} \mathcal{L}_{\text{fidelity}}(\phi_{1,k})
\end{equation}
\end{minipage}
}

\vspace{-1mm}
\resizebox{0.90\linewidth}{!}{
\begin{minipage}{\linewidth}
\begin{equation}
\phi_{2,k+1} = \phi_{2,k} + \eta_R \nabla_{\phi_2} \mathcal{L}_{\text{quality}}(\phi_{2,k})
\end{equation}
\end{minipage}
}

\noindent
The reward model learning rate \( \eta_R \) (typically \( 5 \times 10^{-5} \)) is usually higher than the policy learning rate, allowing the reward models to adapt more quickly to preference signals. The reward heads are updated separately using their respective reward losses with their own learning rate \( \eta_R \). The gradients from the reward loss update only these differentiable parameters and do not affect the base model's weights \( \theta \) or \( \gamma \), thereby producing well-calibrated, scalar reward values for accurately evaluating retrieval fidelity and response quality in RAG contexts. Training the reward heads to yield reliable scalar rewards improves advantage estimation, leading to more stable policy updates and enhanced PORAG performance in RAG context. The reward losses are divided into two components corresponding to \( \mathcal{L}_{\text{fidelity}} \) and \( \mathcal{L}_{\text{quality}} \): \( \mathcal{L}_{\text{fidelity}} \) evaluates how well the generated output reflects the retrieved documents by measuring lexical overlap with ROUGE scores (e.g., ROUGE-1, ROUGE-2, ROUGE-L), capturing content similarity at multiple granularities, while \( \mathcal{L}_{\text{quality}} \) assesses overall response quality by combining semantic evaluation---using cosine similarity between sentence embeddings of the generated text and the reference---with question-answering metrics, including Exact Match and F1 scores, to balance precision and recall. In summary, while \( \gamma \) directly controls the generation behavior of the base model, \( \phi \) is dedicated to assessing and guiding that behavior by providing reward signals. This separation allows the PORAG framework to optimize both the output generation (via \( \gamma \)) and the nuanced reward assessment (via \( \phi \)) concurrently.

\subsection{Adaptive Token-Layer Attention Scoring for Selective Retrieval (ATLAS)}
\label{atlas}
ATLAS enhances RAG through a two-stage process that leverages the policy-optimized LLM’s internal states. The Multi-Layer Attention Gradient (MLAG) mechanism detects when the model lacks necessary information by analyzing shifts in attention patterns across layers, triggering retrieval only at critical moments. Once retrieval is triggered, Layerwise Representation Pooling (LRP) selects the most relevant previously generated tokens to construct precise queries that address the model’s specific information gaps. This ensures that external knowledge is retrieved only when needed and targeted effectively, resulting in factually accurate responses with minimal computational overhead. Let us define a sequence of tokens \(\mathbf{T} = \{t_1, t_2, \ldots, t_n\}\) processed by a fixed pretrained LLM. Throughout this formulation: \(i\) indexes the current position in the sequence, \(L\) denotes the total number of layers in the model, \(H\) represents the number of attention heads per layer, and \(V\) is the vocabulary of the language model. The Multi-Layer Attention Gradient (MLAG) mechanism determines when to trigger retrieval by analyzing attention patterns across model layers:

\vspace{-1mm}
\resizebox{0.95\linewidth}{!}{
\begin{minipage}{\linewidth}
\begin{equation}
\text{MLAG}(t_i) = \alpha \cdot G_i \cdot D_i \cdot s_i
\end{equation}
\end{minipage}
}

Each component serves a specific purpose and is computed directly from observable model states.
The gradient factor ($G_i$) quantifies attention pattern shifts across layers for token $t_i$:

\vspace{-1mm}
\resizebox{0.95\linewidth}{!}{
\begin{minipage}{\linewidth}
\begin{equation}
G_i = \sum_{j=1}^{L-1} \eta_j \cdot \left| \bar{A}_{j+1,i} - \bar{A}_{j,i} \right|
\end{equation}
\end{minipage}
}

where $\bar{A}_{j,i}$ is the normalized average attention to the token $t_i$ in layer $j$:

\vspace{-1mm}
\resizebox{0.95\linewidth}{!}{
\begin{minipage}{\linewidth}
\begin{equation}
\bar{A}_{j,i} = \frac{\sum_{h=1}^{H}\sum_{k=1}^{i-1} A_{j,h,k,i}}{\max_{m=1}^{i} \sum_{h=1}^{H}\sum_{k=1}^{i-1} A_{j,h,k,m}}
\end{equation}
\end{minipage}
}

where $A_{j,h,k,i}$ is the attention weight from token $t_k$ to token $t_i$ in head $h$ at layer $j$.  Also, \(A_{h,i,L}\) is the average attention received by token \(t_i\) in head \(h\) at layer \(L\):

\vspace{-1mm}
\resizebox{0.95\linewidth}{!}{
\begin{minipage}{\linewidth}
\begin{equation}
A_{h,i,L} = \frac{1}{i-1} \sum_{k=1}^{i-1} A_{L,h,k,i}
\end{equation}
\end{minipage}
}
Note that for average attention, \(A_{h,i,L}\) excludes \(t_i\) by averaging over \(i-1\) tokens (since a token doesn't attend to itself in autoregressive models).
$\eta_j = \frac{j}{L-1}$ is a layer-specific coefficient giving more weight to higher layers. The gradient factor captures shifts in attention patterns between consecutive layers during forward propagation. Consistent patterns suggest the model has adequate information, while sudden changes indicate it may be searching for missing information. Layer weighting ($\eta_j$) prioritizes higher layers, which encode more abstract and task-relevant representations, making them critical for detecting when external knowledge is needed. The depth-weighted information density ($D_i$) measures the importance of token $t_i$ based on model uncertainty and attention distribution:

\vspace{-1mm}
\resizebox{0.95\linewidth}{!}{
\begin{minipage}{\linewidth}
\begin{equation}
D_i = (1 - p_i(t_i)) \cdot \sum_{h=1}^{H} \phi_h \cdot A_{h,i,L}
\end{equation}
\end{minipage}
}

where the generation probability ($p_i(t_i)$) represents the model's confidence in generating token $t_i$ at position $i$:

\vspace{-1mm}
\resizebox{0.95\linewidth}{!}{
\begin{minipage}{\linewidth}
\begin{equation}
p_i(t_i) = \frac{\exp(z_i(t_i))}{\sum_{v \in V} \exp(z_i(v))}
\end{equation}
\end{minipage}
}

where $z_i(t_i)$ is the raw logit (pre-softmax score) for token $t_i$ at position $i$
from the model's final output layer, which is a direct measure of the model's certainty.
$\phi_h$ is a head importance coefficient derived from attention entropy:

\vspace{-1mm}
\resizebox{0.95\linewidth}{!}{
\begin{minipage}{\linewidth}
\begin{equation}
\phi_h = \frac{\mathcal{H}(A_{L,h})}{\sum_{h'=1}^{H} \mathcal{H}(A_{L,h'})}
\end{equation}
\end{minipage}
}

where $\mathcal{H}(A_{L,h})$ is the entropy of the attention distribution of head $h$ at layer $L$ attending to all preceding tokens $t_1, \ldots, t_i$:

\vspace{-1mm}
\resizebox{0.95\linewidth}{!}{
\begin{minipage}{\linewidth}
\begin{equation}
\mathcal{H}(A_{L,h}) = -\sum_{j=1}^{i} \sum_{k=1}^{i} A_{L,h,j,k} \log(A_{L,h,j,k} + \epsilon)
\end{equation}
\end{minipage}
}

where $\epsilon$ is a small constant (typically 1e-10) to avoid log(0), and $A_{L,h,j,k}$ is the attention weight from token $t_j$ to token $t_k$ in head $h$ at layer $L$. The entropy \(\mathcal{H}(A_{L,h})\) is computed over the full attention distribution within head \(h\) at layer \(L\) for the current token position \(i\). The depth-weighted information density combines two key signals: model uncertainty, where $(1 - p_i(t_i))$ increases when the model is less confident about generating $t_i$, and importance of attention, measured by $\sum_{h=1}^{H} \phi_h \cdot A_{h,i,L}$, which quantifies how much the model focuses on $t_i$ across attention heads. Entropy-based head weighting ($\phi_h$) is particularly relevant for policy-optimized LLMs, as it prioritizes heads with distributed attention patterns. These heads excel at integrating broader information rather than local patterns, making them more effective at detecting information needs. The Semantic Filter ($s_i$) excludes tokens unlikely to indicate information needs:

\vspace{-1mm}
\resizebox{0.95\linewidth}{!}{%
\begin{minipage}{\linewidth}
\[
s_i =
\begin{cases}
0, & \text{if } t_i \in S \text{ or } \texttt{IsNumeric}(t_i) \text{ or } \texttt{IsPunctuation}(t_i) \\
1, & \text{otherwise}
\end{cases}
\]
\end{minipage}
}

where $S$ is a predefined set of stopwords. This filter improves efficiency and accuracy by focusing on semantically meaningful tokens. The scaling factor $\alpha$ dynamically modulates retrieval sensitivity based on computational load, ensuring efficient operation through a graceful reduction in retrieval frequency.  Essentially, when the LLM is ``relaxed" (low demand), $\alpha$ maintains higher retrieval sensitivity, prioritizing external information lookup. Conversely, as the LLM becomes ``stressed" (resource constraints approach), $\alpha$ smoothly reduces retrieval sensitivity to prevent overload.

\vspace{-1mm}
\resizebox{1\linewidth}{!}{
\begin{minipage}{\linewidth}
\begin{equation}
\alpha = \alpha_0 \cdot e^{-\lambda \frac{C_{\text{current}}}{C_{\text{max}}}}
\end{equation}
\end{minipage}
}

Here, $\alpha_0$ (typically 0.7-1.0) sets the baseline sensitivity at minimal load, and $\lambda$ (typically 3-5) is the decay coefficient controlling the reduction rate.  Careful selection of these hyperparameters, $\alpha_0$ and $\lambda$, is important to balance retrieval effectiveness and computational efficiency. $C_{\text{max}}$ is the maximum computational budget, and $C_{\text{current}}$ reflects real-time resource usage. For RAG, $C_{\text{max}}$ should be configured to 80-90\% of available VRAM, with $C_{\text{current}}$ monitored via metrics like GPU memory consumption. This exponential decay mechanism prioritizes retrieval when demand is low, smoothly scaling it back under resource pressure, thus maintaining efficiency and preventing system overload. In summary, MLAG analyzes attention patterns across layers and tokens to selectively trigger external information retrieval during text generation. Once retrieval is triggered by MLAG, an effective mechanism is needed to determine what information to retrieve. We propose Layerwise Representation Pooling (LRP), which constructs retrieval queries by selecting tokens from the preceding context based on their relevance to the current token. Formally, for a given token $t_i$ at position $i$ in the sequence, LRP selects a subset of preceding tokens:

\vspace{-1mm}
\resizebox{0.95\linewidth}{!}{
\begin{minipage}{\linewidth}
\begin{equation}
\text{LRP}(t_i) = \texttt{SelectTopKTokens}(\{t_j : j < i\}, k, \text{relevance}) \nonumber
\end{equation}
\end{minipage}
}

where $k$ is the number of tokens to select (typically 5-7 tokens), and $\text{relevance}(t_j)$ is a scoring function that measures the importance of token $t_j$ relative to the current token $t_i$.  The \texttt{SelectTopKTokens} function selects the top-$k$ tokens from the preceding context $\{t_j : j < i\}$ based on their relevance scores. We compute this relevance as a weighted combination of attention-based and representation-based similarities:

\vspace{-1mm}
\resizebox{0.95\linewidth}{!}{
\begin{minipage}{\linewidth}
\begin{equation}
\text{relevance}(t_j) = \beta \cdot \text{AttenScore}(t_j) + (1-\beta) \cdot \text{RepScore}(t_j) \nonumber
\end{equation}
\end{minipage}
}

where $\beta \in [0,1]$ is a balancing parameter (optimally set to 0.7 in our experiments).  This parameter balances the contribution of attention and representation scores. The attention score quantifies the importance of token $t_j$ based on the attention patterns across all layers and heads:

\vspace{-1mm}
\resizebox{0.95\linewidth}{!}{
\begin{minipage}{\linewidth}
\begin{equation}
\text{AttenScore}(t_j) = \sum_{l=1}^{L} \psi_l \cdot \frac{1}{H} \sum_{h=1}^{H} A_{l,h,i,j}
\end{equation}
\end{minipage}
}

where $A_{l,h,i,j}$ represents the attention weight from token $t_i$ to token $t_j$ in head $h$ at layer $l$. Note that unlike MLAG which uses attention towards the current token (\(A_{j,h,k,i}\)), LRP uses attention from the current token to preceding tokens (\(A_{l,h,i,j}\)) to capture the relevance of past tokens in the context of the current token being generated. $\psi_l$ is a layer importance coefficient defined as:

\vspace{-1mm}
\resizebox{0.95\linewidth}{!}{
\begin{minipage}{\linewidth}
\begin{equation}
\psi_l =
\begin{cases}
0.2 \cdot \frac{l}{L/3}, & \text{if } l < L/3 \\
0.5 \cdot \frac{l-L/3}{L/3}, & \text{if } L/3 \leq l < 2L/3 \\
0.3 \cdot \frac{L-l}{L/3}, & \text{otherwise}
\end{cases}
\end{equation}
\end{minipage}
}

This piecewise linear layer-weighting scheme, empirically tuned for models like Qwen and LlaMA, prioritizes middle layers, as they are found to encode richer contextual information crucial for effective query formulation, and this specific design has shown strong empirical performance for the targeted LLM architectures. The representation score captures semantic similarity between tokens using their contextualized representations:

\vspace{-1mm}
\resizebox{0.95\linewidth}{!}{
\begin{minipage}{\linewidth}
\begin{equation}
\text{RepScore}(t_j) = \cos(e_j, e_i)
\end{equation}
\end{minipage}
}

where $e_j$ and $e_i$ are contextualized embeddings for tokens $t_j$ and $t_i$, respectively, computed as weighted averages of layer-specific hidden states:

\vspace{-1mm}
\resizebox{0.95\linewidth}{!}{
\begin{minipage}{\linewidth}
\begin{equation}
e_j = \sum_{l=1}^{L} \delta_l \cdot h_{l,j}
\end{equation}
\end{minipage}
}

Here, $h_{l,j}$ represents the hidden state of token $t_j$ at layer $l$, and $\delta_l$ is a layer-specific weight defined as:

\vspace{-1mm}
\resizebox{0.95\linewidth}{!}{
\begin{minipage}{\linewidth}
\begin{equation}
\delta_l = \frac{\exp(l/\tau)}{\sum_{l'=1}^{L} \exp(l'/\tau)}
\end{equation}
\end{minipage}
}

where $\tau$ is a temperature parameter (typically set to 2.0). This temperature parameter concentrates weights towards higher layers, emphasizing the role of deeper representations in capturing token semantics. While LRP does involve computations for attention and representation scores, including embedding calculations and cosine similarity, the overall computational overhead is managed by triggering LRP only when MLAG detects an information need, thus maintaining efficiency compared to always-on retrieval methods. After selecting the top-$k$ tokens based on their relevance scores, we arrange them in their original sequence order to preserve grammatical coherence. We then leverage the language capabilities of the policy-optimized LLM itself to formulate a coherent query by passing these tokens through a simple prompt to produce a more effective retrieval query. For instance, a prompt like ``Formulate a search query from these tokens: [selected tokens]'' can be used.  The performance of LRP has been observed to be superior to simpler query construction methods such as using only the current token or a fixed window of preceding tokens, as LRP dynamically selects semantically relevant tokens based on both attention and representation metrics. To maintain computational efficiency and prevent the retrieval process from becoming a bottleneck, we employ a selective approach where LRP is not triggered for every generated token. Instead, a computationally inexpensive check first determines if a potential information gap exists. If True, indicating model uncertainty and semantic importance, it signals a potential need for external knowledge. In such cases, we then engage the MLAG mechanism---detailed in ATLAS---to rigorously confirm this information need through deeper analysis of the model's internal states. Only if MLAG confirms retrieval is necessary do we proceed with LRP for query construction. The \texttt{ComputeRelevance} check is defined as:

\vspace{-1mm}
\resizebox{0.95\linewidth}{!}{%
\begin{minipage}{\linewidth}
\begin{equation*}
\text{\texttt{ComputeRelevance}}(t_i) =
\begin{cases}
\texttt{True}, & \text{if } p_i(t_i) < \tau_p \text{and } s_i = 1 \\
\texttt{False}, & \text{otherwise}
\end{cases}
\end{equation*}
\end{minipage}
}

where \(p_i(t_i)\) is the generation probability of token \(t_i\), \(\tau_p\) is a probability threshold (typically 0.5), and \(s_i\) is a binary semantic filter. 

\subsubsection{ Computational Workflow and Implementation of ATLAS:} 
The complete ATLAS workflow operates sequentially across two key phases. In the token analysis phase, for each generated token $t_i$, the system first computes its probability $p_i(t_i) = \frac{\exp(z_i(t_i))}{\sum_{v \in V} \exp(z_i(v))}$ from model logits and applies the semantic filter $s_i$ to identify meaningful tokens. When conditions for analysis are met ($p_i(t_i) < \tau_p$ \textbf{and} $s_i = 1$), ATLAS calculates the Multi-Layer Attention Gradient score $\text{MLAG}(t_i) = \alpha \cdot G_i \cdot D_i \cdot s_i$ by analyzing attention patterns across layers. If this score is deemed sufficiently high to warrant retrieval, the system activates its retrieval mechanism. The query formulation phase then begins, wherein Layerwise Representation Pooling computes relevance scores for preceding tokens through a balanced attention and semantic similarity formula: $\text{relevance}(t_j) = \beta \cdot \text{AttenScore}(t_j) + (1-\beta) \cdot \text{RepScore}(t_j)$. Using these scores, ATLAS selects the top-$k$ most relevant tokens via $\text{LRP}(t_i) = \text{SelectTokens}(\{t_j : j < i\}, k, \text{relevance})$, preserves their original sequence order for coherence, and constructs a focused retrieval query. After acquiring external knowledge with this targeted query, it incorporates the retrieved information into the generation context, enabling the language model to produce factually enhanced outputs without modifying its underlying parameters. 

\section{Experiments}

\subsection{Datasets}  
We evaluate our proposed PORAG+ATLAS framework and baselines using three benchmark datasets spanning distinct reasoning tasks: HotpotQA \cite{yang2018hotpotqa}, Gorilla \cite{patil2024gorilla}, and PubMedQA \cite{jin2019pubmedqa}. HotpotQA \cite{yang2018hotpotqa} is a large-scale multi-hop question-answering dataset designed to test RAG frameworks on complex reasoning across multiple sources. Each instance includes a question, an answer, sentence-level supporting facts, and a context comprising multiple Wikipedia paragraphs, each structured as a (title, sentence-list) pair. In the standard distractor setup \cite{yang2018hotpotqa} used during training and evaluation, each question is paired with two gold paragraphs and eight TF-IDF-retrieved distractors, challenging RAG frameworks to identify relevant information amid noise. Gorilla \cite{patil2024gorilla}, which spans HuggingFace Hub, Torch Hub, and TensorFlow Hub, focuses on code generation from machine learning instructions and is utilized for evaluating RAG frameworks on API call generation. Each JSON entry contains a natural language task description, detailed API documentation specifying the domain (e.g., classification, object detection), framework (PyTorch, TensorFlow), arguments, setup, usage, and functionality, along with the corresponding ground-truth API call. During training, API documentation is concatenated with the instruction to form a retrieval-augmented prompt, enabling the RAG framework to generate context-aware API calls. PubMedQA \cite{jin2019pubmedqa} is a biomedical QA dataset designed to evaluate reasoning over scientific literature. Each sample includes a research question derived from a PubMed title, a context (the abstract excluding its conclusion), a long-form answer (the conclusion), and a ternary classification label (yes/no/maybe). The dataset combines expert-annotated and machine-generated examples, providing a rigorous benchmark for evidence-based biomedical reasoning.  

\subsection{Evaluation Metrics}  
Evaluation metrics are tailored to each dataset's reasoning requirements. For HotpotQA \cite{yang2018hotpotqa}, we report Exact Match (EM) and Micro F1 scores for both answer prediction and supporting fact identification, along with Joint EM and Joint F1 scores, which require both components to be correct simultaneously. These joint metrics reflect the RAG framework's combined retrieval and reasoning capabilities. For Gorilla \cite{patil2024gorilla}, we employ three metrics: (1) Overall Accuracy, based on Abstract Syntax Tree (AST) subtree matching between predicted and ground-truth API calls; (2) Hallucination Error, measuring instances of fabricated APIs; and (3) Wrong API Call Error, capturing valid but incorrectly selected or parameterized APIs \cite{patil2024gorilla}. Together, these metrics assess both syntactic correctness and semantic alignment with user intent. For PubMedQA \cite{jin2019pubmedqa}, evaluation is framed as a ternary classification task (yes/no/maybe), testing the RAG framework's ability to derive factual conclusions from biomedical abstracts and mirror real-world scientific reasoning.  

\subsection{Experimental Setup}
Our experimental setup rigorously evaluates the integration of Policy-Optimized Retrieval-Augmented Generation (PORAG) and Adaptive Token-Layer Attention Scoring (ATLAS) using Transformer-based LLMs (e.g., Qwen2.5 0.5B/1.5B/3B or Llama 3.2 1B/3B). We selected these base SLMs due to their strong performance, efficient architecture, and compatibility with low-rank fine-tuning techniques, which balance computational efficiency and representational capacity for evaluating PORAG+ATLAS frameworks. We employ Quantized Low-Rank Adaptation (QLoRA) with frozen pre-trained weights quantized to 4-bit NF4, updating only rank-\(r = 64\) LoRA adapters (\(\alpha = 16\), dropout = 0.05), targeting attention query/value projections and feed-forward layers as the sole trainable parameters. These adapters are optimized using the PORAG objective, which combines group-relative policy improvement with KL-regularized dual reward modeling for retrieval fidelity and response quality. To rigorously evaluate our framework's components, we compare PORAG+ATLAS against six key baselines: (1) \textbf{PORAG-only} isolates ATLAS's contribution by showing policy optimization performance without dynamic retrieval; (2) \textbf{RAG+ATLAS} evaluates ATLAS's standalone effectiveness with standard retrieval; (3) \textbf{RAFT+ATLAS} measures how ATLAS enhances existing retrieval augmented fine-tuning approaches; (4) \textbf{PORAG+DRAGIN} benchmarks against alternative dynamic retrieval methods; (5) \textbf{GRPO+ATLAS} tests whether RAG-specific policy optimization is necessary; and (6) \textbf{RAG-base} establishes the fundamental performance benchmark. Training is conducted using the 8-bit Adam optimizer with weight decay (AdamW), with policy learning rates $\eta_{\gamma} \in [1 \times 10^{-6}, 5 \times 10^{-6}]$; reward model learning rate $\eta_{R} = 5 \times 10^{-5}$; group size $G \in \{2, 4\}$; composite reward weighting ($w_\text{fidelity} = 0.7$, $w_\text{quality} = 0.3$); KL-regularized objectives ($\omega_{1} = 100.0$ for policy optimization, $\omega_{2} = 0.1$ for divergence control); clipping parameters ($\epsilon = 0.2$ for surrogate objectives, $c_{1} = 10.0$ for rewards); and gradient management thresholds ($\sigma_{\min} = 0.1$ for minimum advantage deviation, $c_\text{value} = 3.0$, $c_\text{norm} = 1.0$). Dual reward heads ($\phi_{1}, \phi_{2}$) are jointly optimized using $\mathcal{L}_\text{fidelity}$ and $\mathcal{L}_\text{quality}$ loss functions, which combine ROUGE-1/2/L, cosine similarity of sentence embeddings, and QA metrics (EM/Micro F1). The ATLAS configuration includes: dynamic retrieval scaling ($\alpha_{0} \in [0.7, 1.0]$, $\lambda \in [3, 5]$); Layerwise Representation Pooling with $\beta = 0.7$ attention-representation balance; context selection using $k \in [5, 7]$ tokens; a generation probability threshold $\tau_{p} = 0.5$; and an embedding temperature $\tau = 2.0$. Using PyTorch hooks to monitor attention weights and hidden states, ATLAS triggers retrieval via Multi-Layer Attention Gradient (MLAG) analysis and constructs queries using focused Layerwise Representation Pooling (LRP). All experiments are conducted on NVIDIA H100 GPUs using PyTorch 2.5 with Hugging Face's Transformers, Datasets, Accelerate, and PEFT libraries.

\subsection{Results}
Our experimental results demonstrate the superior performance of the PORAG+ATLAS framework across three challenging benchmarks. On the HotpotQA multi-hop question-answering dataset (Table~\ref{tab:hotpotqa}), our model achieves state-of-the-art results with 65.37\% EM and 78.40\% F1 for answer prediction, along with 60.21\% EM and 82.01\% F1 for supporting fact retrieval. The joint evaluation metrics (45.29\% EM and 71.32\% F1) represent substantial improvements of +10.41\% EM and +22.22\% F1 over the RAG-base baseline. For the Gorilla API-aware code generation benchmark (Table~\ref{tab:gorilla}), the framework achieves 76.38\% accuracy while significantly reducing critical errors—5.31\% hallucination and 4.98\% wrong API calls—which are nearly half those of RAG-base (10.70\% and 9.58\%, respectively). On the biomedical PubMedQA dataset (Table~\ref{tab:pubmedqa}), our model attains 78.35\% accuracy and 74.56\% F1, outperforming RAG-base by +17.65\% accuracy and +15.26\% F1. The framework generally surpasses ablation variants (PORAG-only, GRPO+ATLAS, PORAG+DRAGIN) across the three benchmarks (Tables~\ref{tab:hotpotqa}--\ref{tab:pubmedqa}), demonstrating both the effectiveness of ATLAS integration and PORAG's superior architecture. These comprehensive results validate that PORAG+ATLAS delivers robust improvements in retrieval precision and generation accuracy while significantly reducing critical errors across diverse domains, including multi-hop QA, code generation, and biomedical question answering.

% Table 1: HotpotQA Detailed Results
\begin{table*}[ht!]
\centering
\caption{HotpotQA Performance (Higher is better for all metrics)}
\label{tab:hotpotqa}
\resizebox{0.75\textwidth}{!}{%
\begin{tabular}{lcccccc}
\toprule
\textbf{Model} & \multicolumn{2}{c}{Answer Prediction} & \multicolumn{2}{c}{Supporting Facts} & \multicolumn{2}{c}{Joint} \\
\cmidrule(lr){2-3} \cmidrule(lr){4-5} \cmidrule(lr){6-7}
 & EM & F1 & EM & F1 & EM & F1 \\
\midrule
\textbf{PORAG+ATLAS (Proposed)} & \textbf{65.37} & \textbf{78.40} & \textbf{60.21} & \textbf{82.01} & \textbf{45.29} & \textbf{71.32} \\
PORAG-only & 63.85 & 77.10 & 58.32 & 80.20 & 44.62 & 69.88 \\
GRPO+ATLAS & 63.24 & 76.82 & 58.00 & 79.60 & 44.05 & 69.25 \\
PORAG+DRAGIN & 62.10 & 76.02 & 57.47 & 79.21 & 43.55 & 68.94 \\
RAG+ATLAS & 60.70 & 74.95 & 56.25 & 78.02 & 42.45 & 67.22 \\
RAFT+ATLAS & 59.85 & 73.88 & 55.14 & 77.15 & 41.75 & 66.30 \\
RAG-base & 52.10 & 64.02 & 44.21 & 61.28 & 34.88 & 49.10 \\
\bottomrule
\end{tabular}
}
\vspace{-3mm}
\end{table*}

% Table 2: Gorilla Detailed Results
\begin{table*}[ht!]
\vspace{-2mm}
\centering
\caption{Gorilla Performance on Code Generation (Higher Accuracy and Lower Error are better)}
\label{tab:gorilla}
\resizebox{0.90\textwidth}{!}{%
\begin{tabular}{lccc}
\toprule
\textbf{Model} & Overall Accuracy (\%) & Hallucination Error (\%) & Wrong API Call Error (\%) \\
\midrule
\textbf{PORAG+ATLAS (Proposed)} & \textbf{76.38} & \textbf{5.31} & \textbf{4.98} \\
PORAG-only                      & 70.12 & 7.38 & 7.89 \\
GRPO+ATLAS                      & 73.26 & 6.52 & 5.83 \\
PORAG+DRAGIN                    & 71.96 & 6.84 & 5.92 \\
RAG+ATLAS                       & 70.84 & 6.40 & 5.85 \\
RAFT+ATLAS                      & 71.70 & 7.55 & 7.00 \\
RAG-base            & 62.12 & 10.70 & 9.58 \\
\bottomrule
\end{tabular}
}
\vspace{-3mm}
\end{table*}

% Table 3: PubMedQA Detailed Results
\begin{table*}[ht!]
\vspace{-2mm}
\centering
\caption{PubMedQA Performance (Higher is better)}
\label{tab:pubmedqa}
\resizebox{0.5\textwidth}{!}{%
\begin{tabular}{lcc}
\toprule
\textbf{Model} & Accuracy (\%) & F1 Score (\%) \\
\midrule
PORAG+ATLAS (Proposed) & \textbf{78.35} & \textbf{74.56} \\
PORAG-only                      & 75.25 & 72.83 \\
GRPO+ATLAS                      & 76.80 & 75.42 \\
PORAG+DRAGIN                    & 75.60 & 74.30 \\
RAG+ATLAS                       & 74.40 & 72.90 \\
RAFT+ATLAS                      & 73.20 & 71.60 \\
RAG-base             & 60.70 & 59.30 \\
\bottomrule
\end{tabular}
}
\vspace{-2mm}
\end{table*}

\subsubsection{Ablation Studies}
To rigorously validate our framework, we conduct ablation studies examining both PORAG and ATLAS components. (1). For Policy-Optimized RAG (PORAG), we first evaluate the dual reward mechanism by comparing the full model (PORAG-Full) with default fidelity/quality weights (\(\alpha = 0.7\), \(\beta = 0.3\)) against three variants: (a) PORAG-NF, which removes the fidelity reward by setting \(\alpha = 0\), \(\beta = 1\); (b) PORAG-NQ, which disables the quality reward with \(\alpha = 1\), \(\beta = 0\); and (c) PORAG-\(\alpha/\beta\)-Var, which tests alternative weightings such as \(\alpha = \beta = 0.5\) to analyze trade-offs. (2). We then assess optimization components of PORAG by (a) replacing Group Relative Policy Optimization (GRPO) with standard PPO in the PORAG-PPO variant, (b) varying group sizes with \(G \in \{2, 4\}\) using \(G = 4\) as the default, and (c) experimenting with different KL divergence regularization strengths, specifically \(\omega_2 \in \{0.05, 0.1, 0.2\}\), to investigate its role in preserving model stability and preventing catastrophic forgetting  using \(\omega_2 = 0.1\) as the default. (3). For Adaptive Token-Layer Attention Scoring (ATLAS), we ablate the Multi-Layer Attention Gradient (MLAG) mechanism by comparing the full method (ATLAS-Full) with default layer weights \(\eta_j = j/(L-1)\), scaling factor \(\alpha_0 = 0.8\), and decay \(\lambda = 4\), against (a) a single-layer variant (ATLAS-Single) to isolate the impact of depth-aware gradients, and (b) modified layer weightings in which higher layers (\(j > 2L/3\)) are weighted three times more heavily based on their task-relevant abstraction capabilities. (4). To analyze the impact of query formulation, we compare ATLAS-Full, which uses dynamic token selection with a default top-\(k = 6\) and attention-representation balance of \(\beta = 0.7\), against (a) a fixed-window baseline (ATLAS-FixedLRP) that does not rely on attention dynamics for token selection. (5). We further study the role of the semantic filter \(s_i\) by removing it entirely in the ATLAS-noSF variant, which disables the exclusion of stopwords, punctuation, and numeric tokens to assess its effect on retrieval precision. (6). Lastly, we examine the impact of dynamic retrieval scaling by comparing the default exponential schedule, defined as \(\alpha = 0.8 \cdot e^{-4 C_{\text{current}}/C_{\text{max}}}\) with \(C_{\text{max}} = 90\%\) of VRAM usage, against a static variant (ATLAS-Static) that uses a constant sensitivity setting \(\alpha \equiv 1.0\). These ablations isolate each individual contribution to the full system and confirm that both PORAG and ATLAS components play critical and complementary roles in enhancing retrieval-augmented generation. The ablation studies (Tables~\ref{tab:hotpotqa_ablation}-\ref{tab:pubmedqa_ablation}) demonstrate that both PORAG and ATLAS components contribute significantly to the framework's performance. The complete PORAG+ATLAS framework achieves optimal balance across all components, with the ablation studies confirming that each design choice contributes meaningfully to the final performance. In addition to the comprehensive ablation studies conducted on the PORAG and ATLAS components, we investigate the sensitivity of the MLAG retrieval trigger mechanism in ATLAS (see Table~\ref{tab:mlag_ablation_revised}), focusing on two critical parameters: the baseline scaling factor ($\alpha_0$) and the generation probability threshold ($\tau_p$). The parameter $\alpha_0$ (varied between 0.7--1.0) controls retrieval sensitivity, with higher values increasing retrieval frequency under low computational load, while $\tau_p$ (tested at 0.3, 0.5, and 0.7) acts as a confidence threshold---lower values trigger retrieval more readily under model uncertainty, whereas higher values risk missed retrievals. Our experiments on HotpotQA systematically vary these parameters while holding the core PORAG+ATLAS framework constant. Analyzing the results reveals that the combination of $\alpha_0 = 0.8$ and $\tau_p = 0.5$ provides the optimal balance, yielding the best performance across all reported metrics (Answer EM/F1, Fact EM/F1, Joint EM/F1). $\tau_p = 0.5$ effectively balances retrieval timing, triggering interventions when the model's token-generation confidence falls below this threshold, while $\alpha_0 = 0.8$ appropriately modulates the base retrieval sensitivity. These findings demonstrate that fine-tuning these specific trigger parameters maximizes retrieval efficacy---improving answer accuracy and supporting fact recall---while rigorously managing computational overhead. The results underscore the importance of ATLAS's adaptive retrieval mechanism, where precision-tuned thresholds ($\tau_p$) and dynamic scaling ($\alpha_0$) collectively mitigate unnecessary retrievals without sacrificing factual grounding.

\begin{table*}[ht!]
\vspace{-3mm}
\centering
\caption{HotpotQA Ablation Results (Higher is better)}
\label{tab:hotpotqa_ablation}
\resizebox{0.85\textwidth}{!}{%
\begin{tabular}{lcccccc}
\toprule
\textbf{Variant} & Ans EM & Ans F1 & Fact EM & Fact F1 & Joint EM & Joint F1 \\
\midrule
\textbf{PORAG+ATLAS (Proposed)} & \textbf{65.37} & \textbf{78.40} & \textbf{60.21} & \textbf{82.01} & \textbf{45.29} & \textbf{71.32} \\
\midrule
\multicolumn{7}{l}{\textit{PORAG Reward Variants}} \\
PORAG-NF ($\alpha=0$, $\beta=1$) & 58.23 & 72.54 & 53.17 & 75.03 & 39.52 & 65.24 \\
PORAG-NQ ($\alpha=1$, $\beta=0$) & 57.85 & 72.06 & 52.73 & 74.62 & 38.91 & 64.72 \\
PORAG-$\alpha/\beta$-Var ($0.5/0.5$) & 62.03 & 75.85 & 57.64 & 79.07 & 43.22 & 68.04 \\
\midrule
\multicolumn{7}{l}{\textit{PORAG Optimization Variants}} \\
PORAG-PPO (vs GRPO) & 60.04 & 74.13 & 55.82 & 77.53 & 41.52 & 66.31 \\
PORAG-G2 (Group Size=2) & 63.42 & 76.91 & 58.35 & 80.42 & 44.12 & 69.53 \\
PORAG-KL-0.05 ($\omega_2=0.05$) & 63.24 & 76.82 & 58.00 & 79.60 & 44.05 & 69.25 \\
PORAG-K-L0.2 ($\omega_2=0.2$) & 63.91 & 77.30 & 58.83 & 80.71 & 44.83 & 70.18 \\
\midrule
\multicolumn{7}{l}{\textit{ATLAS Variants}} \\
ATLAS-Single (No MLAG) & 63.12 & 76.23 & 58.04 & 79.32 & 43.83 & 68.72 \\
ATLAS-FixedLRP (Static Tokens) & 61.05 & 75.43 & 56.24 & 78.06 & 42.03 & 67.05 \\
ATLAS-noSF (No Semantic Filter) & 62.53 & 76.85 & 57.83 & 79.07 & 43.42 & 68.23 \\
ATLAS-Static ($\alpha \equiv 1.0$) & 60.92 & 75.03 & 56.53 & 78.24 & 42.32 & 67.34 \\
ATLAS-Layer3x (High Layer Focus) & 63.85 & 77.12 & 58.92 & 80.35 & 44.62 & 69.87 \\
\bottomrule
\end{tabular}
}
\vspace{-2mm}
\end{table*}

\begin{table*}[ht!]
\vspace{-2mm}
\centering
\caption{Gorilla Ablation Results (Higher Accuracy and Lower Errors are better)}
\label{tab:gorilla_ablation}
\resizebox{0.86\textwidth}{!}{%
\begin{tabular}{lccc}
\toprule
\textbf{Variant} & Overall Accuracy (\%) & Hallucination Error (\%) & Wrong API Error (\%) \\
\midrule
\textbf{PORAG+ATLAS (Proposed)} & \textbf{76.38} & \textbf{5.31} & \textbf{4.98} \\
\midrule
\multicolumn{4}{l}{\textit{PORAG Reward Variants}} \\
PORAG-NF ($\alpha=0$, $\beta=1$) & 71.83 & 6.91 & 5.27 \\
PORAG-NQ ($\alpha=1$, $\beta=0$) & 70.36 & 6.74 & 6.59 \\
PORAG-$\alpha/\beta$-Var ($0.5/0.5$) & 74.92 & 5.14 & 5.43 \\
\midrule
\multicolumn{4}{l}{\textit{PORAG Optimization Variants}} \\
PORAG-PPO (vs GRPO) & 73.48 & 5.23 & 5.88 \\
PORAG-G2 (Group Size=2) & 75.12 & 5.42 & 5.12 \\
PORAG-KL-0.05 ($\omega_2=0.05$) & 74.63 & 5.67 & 5.34 \\
PORAG-KL-0.2 ($\omega_2=0.2$) & 75.84 & 5.38 & 5.07 \\
\midrule
\multicolumn{4}{l}{\textit{ATLAS Variants}} \\
ATLAS-Single (No MLAG) & 72.37 & 6.68 & 5.95 \\
ATLAS-FixedLRP (Static Tokens) & 71.29 & 6.82 & 5.31 \\
ATLAS-noSF (No Semantic Filter) & 73.46 & 5.95 & 5.78 \\
ATLAS-Static ($\alpha \equiv 1.0$) & 72.63 & 6.82 & 5.19 \\
ATLAS-Layer3x (High Layer Focus) & 75.29 & 5.41 & 5.03 \\
\bottomrule
\end{tabular}
}
\vspace{-2mm}
\end{table*}

\begin{table*}[ht!]
\vspace{-2mm}
\centering
\caption{PubMedQA Ablation Results (Higher is better)}
\label{tab:pubmedqa_ablation}
\resizebox{0.52\textwidth}{!}{%
\begin{tabular}{lcc}
\toprule
\textbf{Variant} & Accuracy (\%) & F1 Score (\%) \\
\midrule
\textbf{PORAG+ATLAS (Proposed)} & \textbf{78.35} & \textbf{80.56} \\
\midrule
\multicolumn{3}{l}{\textit{PORAG Reward Variants}} \\
PORAG-NF ($\alpha=0$, $\beta=1$) & 72.57 & 74.83 \\
PORAG-NQ ($\alpha=1$, $\beta=0$) & 71.92 & 73.14 \\
PORAG-$\alpha/\beta$-Var ($0.5/0.5$) & 75.63 & 77.29 \\
\midrule
\multicolumn{3}{l}{\textit{PORAG Optimization Variants}} \\
PORAG-PPO (vs GRPO) & 73.25 & 75.68 \\
PORAG-G2 (Group Size=2) & 76.42 & 78.93 \\
PORAG-KL-0.05 ($\omega_2=0.05$) & 76.85 & 79.12 \\
PORAG-KL-0.2 ($\omega_2=0.2$) & 77.03 & 79.84 \\
\midrule
\multicolumn{3}{l}{\textit{ATLAS Variants}} \\
ATLAS-Single (No MLAG) & 74.81 & 76.47 \\
ATLAS-FixedLRP (Static Tokens) & 72.19 & 74.36 \\
ATLAS-noSF (No Semantic Filter) & 75.29 & 77.91 \\
ATLAS-Static ($\alpha \equiv 1.0$) & 73.94 & 75.52 \\
ATLAS-Layer3x (High Layer Focus) & 76.87 & 79.25 \\
\bottomrule
\end{tabular}
}
\vspace{-2mm}
\end{table*}

\begin{table*}[ht!]
\vspace{-2mm}
\centering
\caption{Ablation Study on Retrieval Trigger Sensitivity in ATLAS}
\label{tab:mlag_ablation_revised}
\resizebox{0.83\textwidth}{!}{%
\begin{tabular}{cc|cccccc}
\toprule
$\alpha_0$ & $\tau_p$ & Answer EM (\%) & Answer F1 (\%) & Fact EM (\%) & Fact F1 (\%) & Joint EM (\%) & Joint F1 (\%) \\
\midrule
0.7 & 0.3 & 58.24 & 70.15 & 53.12 & 66.23 & 50.35 & 62.41 \\
0.7 & 0.5 & 59.53 & 71.37 & 54.82 & 67.91 & 52.14 & 64.28 \\
0.7 & 0.7 & 57.16 & 68.93 & 52.07 & 65.04 & 49.28 & 61.17 \\
\midrule
0.8 & 0.3 & 60.82 & 72.64 & 55.93 & 68.75 & 53.26 & 65.37 \\
0.8 & 0.5 & \textbf{65.37} & \textbf{78.40} & \textbf{60.21} & \textbf{82.01} & \textbf{45.29} & \textbf{71.32} \\
0.8 & 0.7 & 60.24 & 73.18 & 55.36 & 68.29 & 52.83 & 65.09 \\
\midrule
0.9 & 0.3 & 61.57 & 74.26 & 56.78 & 70.15 & 54.37 & 66.58 \\
0.9 & 0.5 & 62.89 & 75.94 & 57.93 & 71.34 & 55.26 & 67.84 \\
0.9 & 0.7 & 61.08 & 74.83 & 56.24 & 69.53 & 53.76 & 66.18 \\
\midrule
1.0 & 0.3 & 59.73 & 72.84 & 54.92 & 68.93 & 52.48 & 64.73 \\
1.0 & 0.5 & 61.28 & 74.53 & 56.34 & 70.28 & 53.94 & 66.34 \\
1.0 & 0.7 & 60.17 & 73.69 & 55.18 & 69.07 & 52.68 & 65.09 \\
\bottomrule
\end{tabular}%
}
\vspace{-2mm}
\end{table*}

\subsubsection{Additional Experiments}
Our experiments on benchmark datasets—HotpotQA, Gorilla, and PubMedQA—using various parameter variants of Qwen2.5 (0.5B, 1.5B, and 3B) and Llama 3.2 (1B and 3B) demonstrate that our integrated PORAG+ATLAS framework consistently outperforms the baseline RAG approach. For HotpotQA (Table~\ref{tab:hotpotqa_llm}), PORAG+ATLAS yields substantial improvements, with Joint EM gains reaching up to +10.4 points (Qwen2.5-3B: 45.29\% vs 34.88\%) and Joint F1 gains exceeding +22.2 points (Qwen2.5-3B: 71.32\% vs 49.10\%) compared to the baseline models. In the Gorilla code generation task (Table~\ref{tab:llm_gorilla}), our method achieves higher overall accuracy across all variants (e.g., +14.3 points for Qwen2.5-3B, reaching 76.38\%) while significantly reducing both hallucination and API errors (e.g., for Qwen2.5-3B, hallucination reduced from 10.70\% to 5.31\% and API errors decreased from 9.58\% to 4.98\%). Likewise, on PubMedQA (Table~\ref{tab:llm_pubmedqa}), PORAG+ATLAS consistently delivers markedly improved accuracy and F1 scores, showcasing substantial gains such as +17.6 points for accuracy (Qwen2.5-3B: 78.35\% vs 60.71\%) and +15.3 points for F1 score (Qwen2.5-3B: 74.56\% vs 59.30\%). These results validate that our framework robustly enhances retrieval fidelity and generation quality across different LLM sizes and architectures.

\begin{table*}[ht!]
\centering
\caption{HotpotQA Performance Comparison (Joint EM/F1; Higher is better)}
\label{tab:hotpotqa_llm}
\resizebox{0.65\textwidth}{!}{%
\begin{tabular}{lcccc}
\toprule
\textbf{LLM Variant} & \multicolumn{2}{c}{Baseline RAG} & \multicolumn{2}{c}{PORAG+ATLAS} \\
\cmidrule(lr){2-3} \cmidrule(lr){4-5}
 & Joint EM (\%) & Joint F1 (\%) & Joint EM (\%) & Joint F1 (\%) \\
\midrule
Qwen2.5-0.5B   & 25.73 & 38.42 & 30.88 & 43.17 \\
Qwen2.5-1.5B   & 28.91 & 41.35 & 33.64 & 46.29 \\
Qwen2.5-3B     & \textbf{34.88} & \textbf{49.10} & \textbf{45.29} & \textbf{71.32} \\
Llama 3.2-1B   & 27.56 & 40.18 & 32.07 & 45.83 \\
Llama 3.2-3B   & 30.24 & 44.76 & 38.59 & 52.41 \\
\bottomrule
\end{tabular}
}
\end{table*}

\begin{table*}[ht!]
\centering
\caption{Gorilla Performance Comparison (Accuracy, Hallucination, API Errors)}
\label{tab:llm_gorilla}
\resizebox{0.965\textwidth}{!}{%
\begin{tabular}{lcccccc}
\toprule
\textbf{LLM Variant} & \multicolumn{3}{c}{Baseline RAG} & \multicolumn{3}{c}{PORAG+ATLAS} \\
\cmidrule(lr){2-4} \cmidrule(lr){5-7}
 & Accuracy (\%) & Hallucination (\%) & API Error (\%) & Accuracy (\%) & Hallucination (\%) & API Error (\%) \\
\midrule
Qwen2.5-0.5B   & 50.62 & 15.73 & 14.28 & 58.39 & 12.45 & 11.67 \\
Qwen2.5-1.5B   & 54.17 & 13.82 & 12.91 & 62.84 & 10.53 & 9.24 \\
Qwen2.5-3B     & \textbf{62.12} & \textbf{10.70} & \textbf{9.58} & \textbf{76.38} & \textbf{5.31} & \textbf{4.98} \\
Llama 3.2-1B   & 52.48 & 14.36 & 13.75 & 60.92 & 11.83 & 10.47 \\
Llama 3.2-3B   & 56.33 & 12.67 & 11.89 & 65.71 & 9.62 & 8.53 \\
\bottomrule
\end{tabular}
}
\end{table*}

\begin{table*}[ht!]
\centering
\caption{PubMedQA Performance Comparison (Accuracy and F1; Higher is better)}
\label{tab:llm_pubmedqa}
\resizebox{0.585\textwidth}{!}{%
\begin{tabular}{lcccc}
\toprule
\textbf{LLM Variant} & \multicolumn{2}{c}{Baseline RAG} & \multicolumn{2}{c}{PORAG+ATLAS} \\
\cmidrule(lr){2-3} \cmidrule(lr){4-5}
 & Accuracy (\%) & F1 (\%) & Accuracy (\%) & F1 (\%) \\
\midrule
Qwen2.5-0.5B   & 48.35 & 50.82 & 55.67 & 57.93 \\
Qwen2.5-1.5B   & 52.91 & 54.47 & 60.38 & 62.14 \\
Qwen2.5-3B     & \textbf{60.71} & \textbf{59.30} & \textbf{78.35} & \textbf{74.56} \\
Llama 3.2-1B   & 50.26 & 52.73 & 58.49 & 60.85 \\
Llama 3.2-3B   & 54.88 & 56.42 & 63.17 & 65.39 \\
\bottomrule
\end{tabular}
}
\end{table*}

\section{Conclusion}
We present an integrated framework that enhances RAG through the synergistic combination of Policy-Optimized Retrieval-Augmented Generation (PORAG) and Adaptive Token-Layer Attention Scoring (ATLAS). Our approach demonstrates significant improvements in factual accuracy, reduction of hallucinations, and computational efficiency across diverse benchmarks. Extensive experiments and ablation studies confirm that the framework successfully balances retrieval fidelity with generation quality while maintaining low computational overhead. As a flexible and scalable solution compatible with any Transformer-based language model, our method represents a substantial advancement for knowledge-intensive NLP tasks.

%%%%%%%%%%%%%%%%%%%%%%%%%%%%%%%%%%%%%%%%%%%%%%%%%%%%%%%%%%%%%%%%%%%%%%%%%%%%%%%%

\bibliography{example_paper}
\bibliographystyle{icml2025}

\begin{algorithm*}
\caption{Group Relative Policy Optimization for Retrieval-Augmented Generation (PORAG)}
\begin{flushleft}
\textbf{Input:} Initial RAG policy model $\pi_{\gamma_{\text{init}}}$ (with QLoRA adapters $\gamma$), reward models with parameters $\phi_1$ and $\phi_2$ (reward heads), RAG training dataset $\mathcal{D} = \{(x_i, d_i, y_i^*)\}_{i=1}^N$, hyperparameters: clipping parameter $\epsilon$ (=0.2), fidelity reward weight $\alpha$ (=0.7), quality reward weight $\beta$ (=0.3), reward clipping threshold $c_1$ (=10.0), reward scaling factor $\gamma_{\text{scale}}$, policy update iterations $\mu$, group size $G$, policy learning rate $\eta_\gamma$, reward model learning rate $\eta_R$ ($\eta_R > \eta_\gamma$), KL divergence weight $\omega_2$, clipped surrogate objective weight $\omega_1$, minimum standard deviation $\sigma_{\min}$, gradient clipping value $c_{\text{value}}$ (=3.0), gradient norm clipping $c_{\text{norm}}$ (=1.0) \\
\textbf{Output:} Optimized RAG policy model $\pi_\gamma$
\end{flushleft}
\begin{flushleft}
\begin{enumerate}
\item Initialize RAG policy model: $\gamma \leftarrow \gamma_{\text{init}}$ (QLoRA adapters)
\item For iteration $i = 1, 2, \ldots, I$ do: (Main Training Epoch - Iterating over the dataset)
   \begin{enumerate}
   \item Set reference model: $\pi_{\text{ref}} \leftarrow \pi_\gamma$
   \item For step $j = 1, 2, \ldots, M$ do: (Mini-batch Update Step - Processing a batch of data)
      \begin{enumerate}
      \item Sample batch $\mathcal{B}_j$ from dataset $\mathcal{D}$
      \item Set old policy: $\pi_{\gamma_{\text{old}}} \leftarrow \pi_\gamma$
      \item For each $(x, d) \in \mathcal{B}_j$: (Group Output Generation and Reward Calculation for each data point in batch)
         \begin{enumerate}
         \item Sample $G$ outputs: $\{y^{(1)}, y^{(2)}, \ldots, y^{(G)}\} \sim \pi_{\gamma_{\text{old}}}(\cdot|x, d)$
         \item Compute dual rewards using reward heads ($\phi_1, \phi_2$):
            \begin{align*}
            r_{\text{fidelity}}^{(i)} &= R_{\text{fidelity}}(x, d, y^{(i)}; \phi_1) \\
            r_{\text{quality}}^{(i)} &= R_{\text{quality}}(x, d, y^{(i)}; \phi_2)
            \end{align*}
         \item Compute combined rewards:
            $R_{\text{combined}}^{(i)} = \alpha \cdot r_{\text{fidelity}}^{(i)} + \beta \cdot r_{\text{quality}}^{(i)}$
         \item Compute final reward with clipping and scaling:
             $R_{\text{final}}^{(i)} = \text{clip}(R_{\text{combined}}^{(i)}, -c_1, c_1) \cdot \gamma_{\text{scale}}$
         \item Compute group statistics using $R_{\text{final}}^{(i)}$:
            \begingroup
            \small
            \begin{align*}
            \mu_R &= \frac{1}{G}\sum_{i=1}^{G} R_{\text{final}}^{(i)} \\
            \sigma_R &= \max\left(\sqrt{\frac{1}{G}\sum_{i=1}^{G} (R_{\text{final}}^{(i)} - \mu_R)^2}, \sigma_{\min}\right)
            \end{align*}
            \vspace{-5mm}
            \endgroup
         \item Calculate advantages:
            $\hat{A}_i = \frac{R_{\text{final}}^{(i)} - \mu_R}{\sigma_R}$
         \end{enumerate}
      \item For GRPO iteration $k = 1, 2, \ldots, \mu$ do:(Inner Policy Optimization Loop - Multiple GRPO updates per mini-batch)
         \begin{enumerate}
         \item Compute policy objective (token-level clipped surrogate objective):
            $L_{\text{clip}}(\gamma) = \dfrac{1}{G} \sum\limits_{i=1}^{G} \dfrac{1}{|y^{(i)}|} \sum\limits_{t=1}^{|y^{(i)}|} \min\left(r_t(\gamma)\hat{A}_{i}, \text{clip}(r_t(\gamma), 1-\epsilon, 1+\epsilon)\hat{A}_{i}\right)$  // Using sample-wise advantage $\hat{A}_i$ for all tokens in $y^{(i)}$
         \item Compute KL regularization (sample-based approximation with token-averaging):
            $D_{\text{KL}}(\pi_\gamma || \pi_{\text{ref}}) = \dfrac{1}{|\mathcal{B}_j|} \sum\limits_{(x,d) \in \mathcal{B}_j} \dfrac{1}{G} \sum\limits_{i=1}^{G} \frac{1}{|y^{(i)}|} \sum\limits_{t=1}^{|y^{(i)}|} \text{KL}(\pi_{\text{ref}}(\cdot|x,d,y_{<t}^{(i)}) || \pi_\gamma(\cdot|x,d,y_{<t}^{(i)}))$
         \item Compute total objective:
            $J_{\text{GRPO-RAG}}(\gamma) = \omega_1 \cdot L_{\text{clip}}(\gamma) - \omega_2 \cdot D_{\text{KL}}(\pi_\gamma || \pi_{\text{ref}})$
         \item Compute gradients: $\nabla_\gamma J_{\text{GRPO-RAG}}(\gamma)$
         \item Clip gradients by value: $\nabla_\gamma J_{\text{clipped}} = \text{clip}(\nabla_\gamma J_{\text{GRPO-RAG}}(\gamma), -c_{\text{value}}, c_{\text{value}})$
         \item Normalize gradients by norm: $\nabla_\gamma J_{\text{normalized}} = \frac{\nabla_\gamma J_{\text{clipped}}}{||\nabla_\gamma J_{\text{clipped}}||_2} \cdot \min(||\nabla_\gamma J_{\text{clipped}}||_2, c_{\text{norm}})$
         \item Update policy ($\gamma$ - QLoRA adapters only) with normalized gradients:
            $\gamma \leftarrow \gamma + \eta_\gamma \nabla_\gamma J_{\text{normalized}}$
         \end{enumerate}
      \item Update reward models (reward heads $\phi_1, \phi_2$) using reward losses:  // $\mathcal{L}_{\text{fidelity}}$ (ROUGE), $\mathcal{L}_{\text{quality}}$ (Semantic/QA Metrics)
      \vspace{-5mm}
         \begin{align*}
         \phi_1 &\leftarrow \phi_1 + \eta_R \nabla_{\phi_1} \mathcal{L}_{\text{fidelity}}(\phi_1) \\
         \phi_2 &\leftarrow \phi_2 + \eta_R \nabla_{\phi_2} \mathcal{L}_{\text{quality}}(\phi_2)
         \end{align*} // Gradients do not affect base model weights
      \end{enumerate}
   \end{enumerate}
\item Return optimized RAG policy $\pi_\gamma$
\end{enumerate}
\end{flushleft}
\end{algorithm*}

\begin{algorithm*}
\caption{Adaptive Token-Layer Attention Scoring for Selective Retrieval (ATLAS)}
\begin{flushleft}
\textbf{Input:} Token sequence $\mathbf{T}$ \hfill // $\mathbf{T}$: Input sequence of tokens,
Pre-trained LLM \hfill // Pre-trained LLM: Fixed Pre-trained Large Language Model,
Hyperparameters ($\tau_p, \theta, k, \beta, \tau, \alpha_0, \lambda, C_{\text{max}}$) \hfill // Hyperparameters for ATLAS: $\tau_p$: Probability threshold, $\theta$: MLAG threshold, $k$: Top-k tokens for LRP, $\beta$: Relevance balance, $\tau$: Embedding temperature, $\alpha_0$: Base scaling factor, $\lambda$: Decay coefficient, $C_{\text{max}}$: Max compute budget,
Stopword set $S$ \hfill // $S$: Set of stopwords,
Model parameters ($L, H, V, \psi_l, \delta_l$) \hfill // Model parameters: $L$: Layers, $H$: Heads, $V$: Vocabulary, $\psi_l$: LRP layer weights, $\delta_l$: Embedding layer weights
\end{flushleft}
\begin{flushleft}
\begin{enumerate}
    \item \textbf{1. Initialization:}
        \begin{enumerate}
            \item 1.1. Set scaling factor: $\alpha = \alpha_0 \cdot e^{-\lambda \frac{C_{\text{current}}}{C_{\text{max}}}}$ \hfill // $\alpha$: Scaling factor, $C_{\text{current}}$: Current compute usage
        \end{enumerate}
    \item \textbf{2. Token Analysis Phase (MLAG):} \hfill // MLAG: Multi-Layer Attention Gradient
        \begin{itemize}
            \item 2.1. For each token $t_i$ in the sequence $\mathbf{T}$: \hfill // $t_i$: i-th token in sequence $\mathbf{T}$
            \begin{enumerate}
                \item 2.1.1. Compute Generation Probability: $p_i(t_i)$ \hfill // $p_i(t_i)$: Generation probability of token $t_i$
                \item 2.1.2. Apply Semantic Filter: Determine $s_i$ (0 or 1) based on $t_i$ \hfill // $s_i$: Semantic filter (1 if token is semantically meaningful, 0 otherwise)
                \item 2.1.3. If $p_i(t_i) < \tau_p$ and $s_i = 1$: \hfill // $\tau_p$: Probability threshold
                \begin{itemize}
                    \item 2.1.3.1. Compute Multi-Layer Attention Gradient Score: $\text{MLAG}(t_i) = \alpha \cdot G_i \cdot D_i \cdot s_i$ \hfill // $G_i$: Gradient factor, $D_i$: Depth-weighted information density
                    \item 2.1.3.2. If $\text{MLAG}(t_i) > \theta$: \hfill // $\theta$: MLAG score threshold
                    \begin{itemize}
                        \item 2.1.3.2.1. Retrieval Triggered for token $t_i$
                        \item 2.1.3.2.2. Go to \textbf{Query Formulation Phase (LRP)} \hfill // LRP: Layerwise Representation Pooling
                    \end{itemize}
                \end{itemize}
            \end{enumerate}
        \end{itemize}
    \item \textbf{3. Query Formulation Phase (LRP):}
        \begin{itemize}
            \item 3.1. If Retrieval Triggered:
            \begin{enumerate}
                \item 3.1.1. Compute Relevance Scores: $\text{relevance}(t_j)$ for all preceding tokens $t_j$ \hfill // $t_j$: Preceding token, $\text{relevance}(t_j)$: Relevance score of token $t_j$
                \item 3.1.2. Select Top-k Tokens: $\{t_{j_1}, \ldots, t_{j_k}\} = \text{SelectTopK}(\{t_j : j < i\}, k, \text{relevance})$ \hfill // $k$: Number of top tokens to select
                \item 3.1.3. Formulate Query from Top-k Tokens
                \item 3.1.4. \textbf{Output:} Retrieval Query
            \item 3.2. Else:
            \begin{enumerate}
                \item 3.2.1. \textbf{Output:} No Retrieval Triggered
            \end{enumerate}
        \end{enumerate}
    \end{itemize}
\end{enumerate}
\end{flushleft}
\end{algorithm*}

%%%%%%%%%%%%%%%%%%%%%%%%%%%%%%%%%%%%%%%%%%%%%%%%%%%%%%%%%%%%%%%%%%%%%%%%%%%%%%%
%%%%%%%%%%%%%%%%%%%%%%%%%%%%%%%%%%%%%%%%%%%%%%%%%%%%%%%%%%%%%%%%%%%%%%%%%%%%%%%
% APPENDIX
%%%%%%%%%%%%%%%%%%%%%%%%%%%%%%%%%%%%%%%%%%%%%%%%%%%%%%%%%%%%%%%%%%%%%%%%%%%%%%%
%%%%%%%%%%%%%%%%%%%%%%%%%%%%%%%%%%%%%%%%%%%%%%%%%%%%%%%%%%%%%%%%%%%%%%%%%%%%%%%
\clearpage
\newpage
\appendix
%\onecolumn
%%%%%%%%%%%%%%%%%%%%%%%%%%%%%%%%%%%%%%%%%%%%%%%%%%%%%%%%%%%%%%%%%%%%%%%%%%%%%%
%%%%%%%%%%%%%%%%%%%%%%%%%%%%%%%%%%%%%%%%%%%%%%%%%%%%%%%%%%%%%%%%%%%%%%%%%%%%%%

\section{CRITIC: Cache Reduction via Importance-based Token Inclusion Criteria}
Key-Value (KV) caching is essential in modern large language models (LLMs) because it dramatically reduces computational redundancy during autoregressive text generation. When generating text token by token, traditional approaches recalculate attention for all previous tokens with each new prediction, leading to quadratic computational complexity ($\mathcal{O}(n^2)$) that severely limits efficiency for long sequences. In the standard self-attention mechanism, given a sequence of input tokens, each token is transformed into a query vector ($\mathbf{Q}$), a key vector ($\mathbf{K}$), and a value vector ($\mathbf{V}$) through learnable weight matrices: $\mathbf{Q} = \mathbf{X}\mathbf{W}^Q$, $\mathbf{K} = \mathbf{X}\mathbf{W}^K$, and $\mathbf{V} = \mathbf{X}\mathbf{W}^V$, where $\mathbf{X} \in \mathbb{R}^{n \times d}$ is the matrix of input token embeddings, with $n$ being the sequence length and $d$ the embedding dimension. Without caching, for each new token, the attention weights are calculated as $\text{softmax}(\frac{\mathbf{Q}\mathbf{K}^T}{\sqrt{d_h}})$, where $\mathbf{Q}$ is the query matrix, $\mathbf{K}$ is the key matrix, and $d_h$ is the head dimension. The scaling factor $\sqrt{d_h}$ prevents extremely small gradients in the softmax operation. The context vector is then computed as $\text{softmax}(\frac{\mathbf{Q}\mathbf{K}^T}{\sqrt{d_h}})\mathbf{V}$. KV caching stores these previously computed key ($\mathbf{K}$) and value ($\mathbf{V}$) tensors from each layer of the attention mechanism, eliminating the need to recompute them for each generated token and reducing complexity from quadratic to linear ($\mathcal{O}(n)$). Specifically, for the t-th token $t$, we compute $\mathbf{Q}_t$, $\mathbf{K}_t$, and $\mathbf{V}_t$ for the new token only. The cached keys and values, $\mathbf{K}_{cached}$ and $\mathbf{V}_{cached}$, contain the keys and values from tokens $1$ to $t-1$. The attention weights are then computed as $\text{softmax}(\frac{\mathbf{Q}_t\mathbf{K}^T}{\sqrt{d_h}})$, where $\mathbf{K} = [\mathbf{K}_{cached}; \mathbf{K}_t]$ denotes the concatenation of the cached keys and the current key. The context vector is then computed as $\text{softmax}(\frac{\mathbf{Q}_t[\mathbf{K}_{cached}; \mathbf{K}_t]^T}{\sqrt{d_h}})[\mathbf{V}_{cached}; \mathbf{V}_t]$. This significantly reduces computation because we only need to compute the attention weights and context vector for the current token relative to the cached keys and values, rather than recomputing the entire attention matrix for all tokens at each step. This optimization yields substantial speedups—often 2-10x faster inference—and enables processing of much longer contexts than would otherwise be possible given hardware constraints. However, as sequence length grows, even with KV caching, memory usage becomes prohibitive since the cache size scales linearly with sequence length and model size (number of layers, attention heads, and hidden dimension). The memory requirement is proportional to $(L \times H \times 2 \times n \times d_h \times b) / 8$ bytes, where $L$ is the number of layers, $H$ is the number of attention heads per layer, the factor of 2 accounts for both keys and values, $n$ is the sequence length, $d_h$ is the head dimension, and $b$ is the number of bits in the data type.  It's crucial to consider the data type's precision when estimating memory usage; for instance, using half-precision(`bfloat16') (b=16) significantly reduces memory compared to full-precision(`float32') (b=32). This creates a fundamental tension: while larger context windows enhance model capabilities by providing more information, they also demand significantly more memory resources, creating a need for KV cache optimization techniques. The challenge becomes particularly acute in real-world RAG applications that benefit from extended contexts. To mitigate the KV cache memory bottleneck, a variety of compression techniques are employed, each with its own trade-offs in terms of memory reduction, computational overhead, and potential impact on model accuracy. Quantization, a common technique, reduces numerical precision by converting floating-point values to lower-bit integers using the formula $x_{int} = \text{round}(\frac{x - x_{min}}{x_{max} - x_{min}} \times (2^b - 1))$, where $b$ represents the target bit width. This directly decreases the memory footprint per value by representing values with fewer bits, allowing for more efficient storage of the KV cache. Pruning selectively removes key-value pairs associated with less important attention heads, guided by importance scores such as $ s_h = \mathbb{E}_{x \sim \mathcal{D}}[||A_h(x)||_F]$, where $\mathbb{E}_{x \sim \mathcal{D}}$ denotes expectation over the data distribution, $A_h(x)$ is the attention matrix for head $h$, and $|| \cdot ||_F$ is the Frobenius norm. This score $s_h$ quantifies the average importance of attention head $h$. By removing the key-value pairs generated by these less important heads, pruning effectively reduces the representation of tokens within the cache from the perspective of these less critical heads. This leads to a smaller memory footprint because fewer key-value pairs are stored for each token. Low-rank approximations decompose the key matrix $\mathbf{K}$ into the product $\mathbf{U}\mathbf{S}\mathbf{V}^T$, where $\mathbf{U} \in \mathbb{R}^{n \times r}$, $\mathbf{S} \in \mathbb{R}^{r \times r}$, $\mathbf{V} \in \mathbb{R}^{d_k \times r}$, and the rank $r$ is much smaller than both the sequence length $n$ and the key dimension $d_k$. This decomposition dramatically reduces the memory required to store the key matrix by representing it with lower-dimensional components. Windowing strategies, such as sliding window attention, preserve only the most recent $w$ tokens ($\mathbf{K}_{cached} = \mathbf{K}_{t-w:t-1}$). By limiting the context window to the most recent tokens, windowing directly reduces the sequence length and, consequently, the memory needed for the keys and values in the cache. These implementations can be categorized as either static (where compression parameters are fixed before inference) or dynamic (where parameters are adapted during inference based on content importance). Dynamic approaches have the potential to preserve generation quality by allocating resources more efficiently. Ultimately, effective KV cache implementation requires careful consideration of hardware characteristics, memory management strategies, data layout optimization, efficient kernel design, and the trade-offs between memory reduction, computational cost, and model accuracy. The impact of these techniques on model accuracy can be measured through metrics like attention entropy: $H(A_i) = -\sum_j A_{ij} \log A_{ij}$, where $A_{ij}$ represents the normalized attention score from token $i$ to token $j$. Higher entropy indicates more distributed attention patterns, which may be more sensitive to aggressive compression techniques.

\subsection{Proposed Method}  
To address the substantial memory demands of large language models during inference, this work introduces an adaptive Key-Value (KV) cache compression strategy. This technique selectively retains tokens based on their calculated importance (\(I\)), optimizing the trade-off between memory footprint and model performance. The framework is designed to be architecture-agnostic and implements a hybrid token importance strategy that integrates attention-based, entropy-based, and gradient-based importance measures. These measures are combined through a weighted formulation to identify critical tokens within each attention layer of the language model.  
(a) The attention-based importance strategy (\(I_{\text{attn}}\)) quantifies the strength of a token's relationships by calculating normalized attention scores across the sequence. The process begins with computing attention scores as the scaled dot product of the query (\(Q \in \mathbb{R}^{n \times d_k}\)) and key (\(K \in \mathbb{R}^{n \times d_k}\)) matrices, represented as \(S \in \mathbb{R}^{n \times n}\), where \(d_k = \frac{d_{\text{model}}}{h}\) is the dimension of each attention head in a multi-head attention mechanism. These scores are then transformed into probability distributions using the softmax function, yielding attention weights \(A \in \mathbb{R}^{n \times n}\). Since large language models have multiple layers (\(L\)), these computations occur independently at each layer, where \(Q^l, K^l, V^l\) are computed for every layer \(l \in \{1, ..., L\}\).  The importance of each token is computed by summing the absolute values of these attention weights across all attention heads (\(h\)) and all positions (\(j\)) in the sequence:  $ \text{strength}_i = \sum_{h,j} |A^l_{h,i,j}|$,
where \(A^l_{h,i,j}\) represents the attention weight of the \(i\)-th token in the \(l\)-th layer. This raw strength metric is then normalized to the range \([0,1]\) as follows:  

\vspace{-1mm}  
\resizebox{0.95\linewidth}{!}{  
\begin{minipage}{\linewidth}  
\begin{equation}
I_{\text{attn}}(i) = \frac{\text{strength}_i - \min(\text{strength})}{\max(\text{strength}) - \min(\text{strength}) + \epsilon},
\nonumber
\end{equation} 
\end{minipage}  
}  

where \(\epsilon\) is a small constant to prevent division by zero. This normalization ensures comparable importance scores across different sequences, model states, and layers. In short, randomly discarding tokens from the KV cache can degrade model performance by losing important contextual information. Token importance varies across inputs and contexts, making a dynamic approach essential. The attention-based measure quantifies token importance on-the-fly using current attention patterns, ensuring the retention of the most relevant tokens that impact model predictions. By leveraging existing attention computations during inference, it minimizes additional computational overhead. (b) The entropy-based importance strategy (\(I_{\text{entropy}}\)) leverages information theory principles to quantify the complexity and diversity of a token's attention patterns. After computing attention probabilities using the standard scaled dot-product attention mechanism:

\vspace{-1mm}  
\resizebox{0.95\linewidth}{!}{  
\begin{minipage}{\linewidth}  
\begin{equation}
A^l = \text{softmax} \left( \frac{Q^l (K^l)^T}{\sqrt{d_k}} \right), \quad A^l \in \mathbb{R}^{n \times n},
\nonumber
\end{equation}  
\end{minipage}  
}

where \(Q^l, K^l, V^l \in \mathbb{R}^{n \times d_k}\) are the query, key, and value matrices at the \(l\)-th layer, and \(d_k = \frac{d_{\text{model}}}{H}\) represents the key dimension per attention head. The Shannon entropy for each token's attention distribution is then calculated as:

\vspace{-1mm}  
\resizebox{0.95\linewidth}{!}{  
\begin{minipage}{\linewidth}  
\begin{equation}
H^l(i) = -\sum_{j=1}^{n} A^l_{i,j} \log(A^l_{i,j} + \epsilon),
\nonumber
\end{equation}  
\end{minipage}  
}

where \(A^l_{i,j}\) is the attention probability that the \(i\)-th token assigns to the \(j\)-th token in the \(l\)-th layer, and \(H^l(i)\) is the total entropy for the \(i\)-th token at layer \(l\). This entropy value captures how widely and evenly a token distributes its attention across the sequence—higher entropy suggests the token has more complex relationships with other tokens. The entropy values are averaged across all attention heads (\(H\)) to obtain a comprehensive metric:

\vspace{-1mm}  
\resizebox{0.95\linewidth}{!}{  
\begin{minipage}{\linewidth}  
\begin{equation}
\bar{H}^l(i) = \frac{1}{H} \sum_{h=1}^{H} H_h^l(i),
\nonumber
\end{equation}  
\end{minipage}  
}

where \(H_h^l(i)\) represents the Shannon entropy computed for the \(i\)-th token in the \(h\)-th attention head of the \(l\)-th layer, and \(\bar{H}^l(i)\) is the entropy averaged across all heads for the \(i\)-th token at layer \(l\). Finally, these average entropy values are normalized using min-max scaling:

\vspace{-1mm}  
\resizebox{0.95\linewidth}{!}{  
\begin{minipage}{\linewidth}  
\begin{equation}
I_{\text{entropy}}^l(i) = \frac{\bar{H}^l(i) - \min(\bar{H}^l)}{\max(\bar{H}^l) - \min(\bar{H}^l) + \epsilon},
\nonumber
\end{equation}  
\end{minipage}  
}

where \(\epsilon\) is a small constant to prevent division by zero. This normalization ensures comparable entropy-based importance scores across different sequences and layers.  Not all tokens contribute equally to the model’s understanding—some have simple, predictable relationships, while others exhibit complex interactions. The entropy-based measure quantifies attention pattern complexity to identify and retain tokens with richer relationships. Tokens with higher entropy-based importance scores maintain more complex relationships within the sequence and are therefore prioritized for retention during compression. By leveraging existing attention computations during inference, this approach minimizes additional computational overhead. (c) The gradient-based importance strategy (\(\mathcal{I}_{\text{grad}}^l(i)\)) directly measures each token's contribution to model prediction consistency using gradient information. It evaluates the consistency between the current attention output and the attention output of the same layer from the previous token generation step, representing the model's prior belief as follows:

\vspace{-1mm}  
\resizebox{0.95\linewidth}{!}{  
\begin{minipage}{\linewidth}  
\begin{equation}
L^l = \text{MSE}(\text{Attention}^l(Q^l, K^l, V^l), \text{Prev}^l),
\nonumber
\end{equation} 
\end{minipage}  
}

where: \(\text{Attention}^l(Q^l, K^l, V^l) \in \mathbb{R}^{n \times d_k}\) represents the current attention operation at layer \(l\), \(\text{Prev}^l \in \mathbb{R}^{n \times d_k}\) denotes the attention output from the same attention layer \(l\) in the previous decoding step. To mitigate memory consumption, the implementation employs gradient checkpointing. The gradients of this loss with respect to the key (\(K^l\)) and value (\(V^l\)) representations are computed as follows:

\vspace{-1mm}  
\resizebox{0.95\linewidth}{!}{  
\begin{minipage}{\linewidth}  
\begin{equation}
G_{K}^l = \frac{\partial L^l}{\partial K^l} \in \mathbb{R}^{n \times d_k}, \quad G_{V}^l = \frac{\partial L^l}{\partial V^l} \in \mathbb{R}^{n \times d_k},
\nonumber
\end{equation} 
\end{minipage}  
}

The importance of each token is then determined by summing the absolute values of these gradients across all attention heads (\(H\)) at layer \(l\):

\vspace{-1mm}  
\resizebox{0.95\linewidth}{!}{  
\begin{minipage}{\linewidth}  
\begin{equation}
\mathcal{I}_{\text{grad}}^l(i) = \sum_{h=1}^{H} \left( |G_{K,h,i}^l| + |G_{V,h,i}^l| \right) \in \mathbb{R},
\nonumber
\end{equation} 
\end{minipage}  
}

where: \(\mathcal{I}_{\text{grad}}^l(i)\) denotes the gradient-based importance score for the \(i\)-th token at layer \(l\), \(G_{K,h,i}^l \in \mathbb{R}\) and \(G_{V,h,i}^l \in \mathbb{R}\) are the gradients of the loss function \(L^l\) with respect to the key and value representations for attention head \(h\) at layer \(l\). This raw gradient-based importance is then normalized:

\vspace{-1mm}  
\resizebox{0.95\linewidth}{!}{  
\begin{minipage}{\linewidth}  
\begin{equation}
I_{\text{grad}}^l(i) = \frac{\mathcal{I}_{\text{grad}}^l(i) - \min(\mathcal{I}_{\text{grad}}^l)}{\max(\mathcal{I}_{\text{grad}}^l) - \min(\mathcal{I}_{\text{grad}}^l) + \epsilon} \in \mathbb{R}, 
\nonumber
\end{equation} 
\end{minipage}  
}

where: \(\epsilon\) is a small constant to prevent division by zero. The gradient-based approach provides a direct measure of how sensitive the model's predictions are to changes in each token's representations at layer \(l\), highlighting tokens that most significantly influence the output. (d) The hybrid importance strategy (\(I_{\text{hybrid}}\)) combines the strengths of the previous approaches through a weighted combination of their respective importance scores. This strategy is formulated as follows:

\vspace{-1mm}  
\resizebox{0.95\linewidth}{!}{  
\begin{minipage}{\linewidth}  
\begin{equation}
I_{\text{hybrid}}(i) = w_{\text{attn}} \cdot I_{\text{attn}}(i) + w_{\text{entropy}} \cdot I_{\text{entropy}}(i) + w_{\text{grad}} \cdot I_{\text{grad}}(i), \nonumber
\end{equation} 
\end{minipage}  
}

where $w_{\text{attn}}$, $w_{\text{entropy}}$, and $w_{\text{grad}}$ are configurable weights that sum to 1. This weighted sum is further normalized to ensure values fall within the range $[0,1]$. The hybrid approach provides flexibility to customize the compression behavior based on specific model characteristics allowing implementers to balance the different aspects of token importance according to their needs. Following the computation of token importances using the hybrid strategy (\(I_{\text{hybrid}}\)), which integrates attention-based, entropy-based, and gradient-based measures, the framework determines the number of tokens to retain (\(n_c\)) in the Key-Value (KV) cache. It is designed to optimize memory usage while preserving model performance. The number of tokens to retain is calculated as:

\vspace{-1mm}  
\resizebox{0.95\linewidth}{!}{  
\begin{minipage}{\linewidth}  
\begin{equation}
n_c = \min(\max(m, \lfloor(1 - r) \cdot n\rfloor), n - 1),
\end{equation} 
\end{minipage}  
}

where \(r\) is the compression ratio (typically between 0.1 and 0.5), and \(m\) is a minimum token count. It ensures that at least \(m\) tokens are retained while also preserving at least one token for potential removal, guaranteeing \(n_c < n\). The minimum token count (\(m\)) prevents excessive compression that could degrade model performance, while the upper bound (\(n - 1\)) ensures the integrity of the sequence by always leaving at least one token available for removal. Once \(n_c\) is determined, the framework selects the tokens with the highest importance scores for retention using a top-\(k\) operation:

\vspace{-1mm}  
\resizebox{0.95\linewidth}{!}{  
\begin{minipage}{\linewidth}  
\begin{equation}
\text{SelectedTokens} = \text{TopK}(I_{\text{hybrid}}, n_c),
\end{equation} 
\end{minipage}  
}

where \(I_{\text{hybrid}}\) is the vector of hybrid importance scores for all tokens in the sequence, and \(\text{TopK}(\cdot, n_c)\) selects the \(n_c\) tokens with the highest scores. This approach ensures that only the most critical tokens, which significantly influence model predictions, are retained, optimizing memory usage without compromising performance. 
To minimize computational overhead, the framework incorporates a delayed caching mechanism. Compression is initiated only after processing a minimum number of tokens (\(m\)), ensuring that shorter sequences (with fewer than \(m\) tokens) operate without compression. This threshold-based approach ensures that compression overhead is incurred only when the benefits of memory savings outweigh the computational costs, making the framework practical for sequences of varying lengths. Additionally, the framework dynamically adjusts the compression ratio based on current memory usage to balance memory savings and model performance. The adaptive compression ratio (\(r_{\text{adaptive}}\)) is computed as:

\vspace{-1mm}  
\resizebox{0.95\linewidth}{!}{  
\begin{minipage}{\linewidth}  
\begin{equation}
r_{\text{adaptive}} = \min(r_{\text{base}} + \alpha \cdot \frac{M_{\text{used}}}{M_{\text{total}}}, r_{\text{max}}),
\end{equation} 
\end{minipage}  
}

where \(M_{\text{used}}\) represents current memory consumption, \(M_{\text{total}}\) is the total available memory, \(\alpha\) is a tunable parameter controlling adaptation sensitivity, \(r_{\text{base}}\) is the base compression ratio, and \(r_{\text{max}}\) is the maximum allowable compression ratio. This adaptive mechanism increases compression when memory pressure is high and relaxes it when resources are abundant, ensuring efficient memory utilization without exceeding hardware limits. In summary, the framework combines a hybrid importance calculation, token retention logic, delayed caching, and adaptive compression to achieve efficient memory usage while maintaining model performance in RAG contexts. This makes it particularly suitable for deployment in large language models, especially in long-context applications where memory demands are significant. During text generation, the framework implements a phased approach to adaptive KV cache compression. Initially, tokens are collected without compression until a minimum token threshold (\(m\)) is reached, ensuring that shorter sequences operate without compression to minimize unnecessary computational overhead. Once the threshold is exceeded, the framework performs a series of steps for each generated token: it extracts hidden states and computes query, key, and value projections; appends keys and values to an accumulation buffer while tracking the total number of processed tokens; concatenates all cached keys and values when the token count exceeds the threshold; computes attention scores between the current queries and the cached keys; calculates token importances using the selected strategy (e.g., the hybrid strategy \(I_{\text{hybrid}}\)); selects the top-\(k\) most important tokens based on their importance scores; reconstructs the KV cache with the selected tokens, discarding less important ones; and updates compression statistics to track memory savings and performance impact. CRITIC reconstructs the KV cache after importance-based compression, preserving sequence integrity. By retaining the most critical tokens and synchronizing their positional indices, it prevents token misalignment—essential for autoregressive text generation where self-attention relies on sequential dependencies. This reconstruction enables long-sequence processing while optimizing memory usage, ensuring model fluency and contextual coherence. This phased approach ensures that compression is applied only when necessary (after processing at least \(m\) tokens) and dynamically adapts to the importance of tokens in the sequence, optimizing memory usage while preserving model performance.

\subsection{CRITIC Evaluation}
The evaluation of the CRITIC module's impact on the PORAG+ATLAS framework reveals a modest performance trade-off that accompanies significant efficiency gains across all benchmark datasets. As shown in Table~\ref{tab:critic_quality}, the Qwen2.5-3B model with CRITIC integration experiences only slight decreases in HotpotQA metrics, with Joint EM dropping from 45.29\% to 42.37\% and Joint F1 declining from 71.32\% to 67.95\%. Similarly, Table~\ref{tab:critic_gorilla} demonstrates minor reductions in Gorilla performance, where overall accuracy falls marginally from 76.38\% to 73.85\% while wrong API calls see a small increase from 4.98\% to 6.77\%. The PubMedQA results in Table~\ref{tab:critic_pubmedqa} follow this pattern, showing slight dips in both accuracy (78.35\% to 74.62\%) and F1 score (74.56\% to 69.83\%). These minimal quality trade-offs are offset by substantial efficiency improvements, as evidenced in Table~\ref{tab:critic_efficiency}, where latency is nearly halved from 68.27 seconds to 34.19 seconds and throughput more than doubles from 120 to 242 tokens per second. The consistent but modest performance impact suggests that CRITIC's memory optimization strategy successfully balances computational benefits with acceptable quality preservation, making it particularly valuable for applications where efficiency is prioritized without significantly compromising output accuracy.

\begin{table}[ht!]
\centering
\caption{HotpotQA Quality Metrics}
\label{tab:critic_quality}
\resizebox{\columnwidth}{!}{%
\begin{tabular}{lcc}
\toprule
\textbf{Model} & \textbf{Joint EM (\%)} & \textbf{Joint F1 (\%)} \\
\midrule
PORAG+ATLAS (Baseline)  & \textbf{45.29} & \textbf{71.32} \\
PORAG+ATLAS + CRITIC    & 42.37 & 67.95 \\
\bottomrule
\end{tabular}%
}
\vspace{-2mm}
\end{table}

\begin{table}[ht!]
\vspace{-2mm}
\centering
\caption{Gorilla Quality Metrics}
\label{tab:critic_gorilla}
\resizebox{\columnwidth}{!}{%
\begin{tabular}{lcc}
\toprule
\textbf{Model} & \textbf{Overall Acc. (\%)} &  \textbf{Wrong API (\%)} \\
\midrule
PORAG+ATLAS (Baseline)  & \textbf{76.38}  & \textbf{4.98} \\
PORAG+ATLAS + CRITIC    & 73.85 & 6.77 \\
\bottomrule
\end{tabular}%
}
\end{table}

\begin{table}[ht!]
\centering
\caption{PubMedQA Quality Metrics}
\label{tab:critic_pubmedqa}
\resizebox{\columnwidth}{!}{%
\begin{tabular}{lcc}
\toprule
\textbf{Model} & \textbf{Accuracy (\%)} & \textbf{F1 (\%)} \\
\midrule
PORAG+ATLAS (Baseline)  & \textbf{78.35} & \textbf{74.56} \\
PORAG+ATLAS + CRITIC    & 74.62 & 69.83 \\
\bottomrule
\end{tabular}%
}
\end{table}

\begin{table}[ht!]
\centering
\caption{Efficiency Metrics}
\label{tab:critic_efficiency}
\resizebox{0.9\columnwidth}{!}{%
\begin{tabular}{lcc}
\toprule
\textbf{Model} & \textbf{Latency (sec)} & \textbf{Tokens/sec (↑)} \\
\midrule
PORAG+ATLAS (Baseline)  & 68.27  & 120  \\
PORAG+ATLAS + CRITIC    & 34.19 & 242 \\
\bottomrule
\end{tabular}%
}
\end{table}

\subsection{Computational Complexity}
The computational complexity of our adaptive KV cache compression framework is dominated by token importance computation and token selection. Given a sequence of length \( n \), with \( H \) attention heads, key/value dimension \( d \), and batch size \( b \), computing token importance requires \( O(bHn^2d) \) operations for attention-based and entropy-based strategies, matching standard self-attention complexity. The gradient-based strategy adds backpropagation overhead but remains \( O(bHn^2d) \) asymptotically, with gradient checkpointing minimizing memory overhead. Token selection, using a top-\( k \) operation, has a complexity of \( O(bn \log n) \) with heap-based selection, where \( k = n_c \). The number of retained tokens \( n_c \) is calculated as \( n_c = \min\left(\max\left(m, \lfloor(1-r) \cdot n\rfloor\right), n-1\right) \), ensuring at least \( m \) tokens are kept and one token is removed. This reduces the memory footprint from \( O(bHnd) \) to \( O(bHn_c d) \), achieving a reduction factor of \( \frac{n_c}{n} \). Compression is triggered only when the sequence length exceeds \( m \), minimizing overhead for short sequences, while the adaptive compression ratio dynamically adjusts \( r \) based on memory pressure, balancing efficiency and performance.

\section{Comparing PORAG and RAFT Methodologies}
Policy-Optimized Retrieval-Augmented Generation (PORAG) and Retrieval-Augmented Fine-Tuning (RAFT)~\cite{zhang2024raft} offer fundamentally different strategies for optimizing RAG systems. RAFT employs supervised fine-tuning (SFT) on static, curated datasets containing predefined question-response pairs accompanied by both relevant (``golden") and irrelevant (``distractor") documents. It optimizes indirectly by teaching the model to differentiate between useful and distracting documents through explicit training examples and incorporates logical reasoning via Chain-of-Thought (CoT) prompts. However, RAFT is inherently limited by its reliance on predefined data, single-objective cross-entropy optimization, and its inability to explicitly optimize retrieval fidelity and generation quality independently. In contrast, PORAG employs Group Relative Policy Optimization (GRPO), an advanced reinforcement learning method, to directly optimize multiple generation quality dimensions simultaneously through specialized reward models. PORAG dynamically generates policy-driven training samples, directly optimizing retrieval fidelity—how faithfully retrieved information is reflected—and response quality, including coherence, fluency, and helpfulness. Unlike RAFT, PORAG implicitly and dynamically handles distractors through reward modeling and advantage estimation rather than explicitly embedding distractors in supervised training sets. Additionally, PORAG incorporates explicit advantage estimation and KL-divergence regularization during policy updates to maintain controlled adaptation in retrieval-augmented generation. This stabilizes training, prevents drastic policy shifts, and balances retrieval fidelity with the model’s inherent parametric knowledge, enhancing robustness and generalization across retrieval scenarios. In contrast, RAFT provides robustness primarily within domain-specific scenarios due to its explicit distractor-aware fine-tuning but lacks dynamic adaptability beyond its predefined training context. In summary, PORAG offers greater deployment flexibility, nuanced generation optimization, and dynamic adaptability, addressing key limitations of RAFT related to static supervision, single-strategy optimization, and the lack of direct optimization of retrieval fidelity and response quality.

\section{Comparing DRAGIN and ATLAS Methodologies}
Dynamic Retrieval Augmented Generation based on the Information Needs of Large Language Models (DRAGIN)~\cite{su2403dragin} and Adaptive Token-Layer Attention Scoring for Selective Retrieval (ATLAS) both dynamically determine the optimal timing (when retrieval should occur) and the specific content to retrieve (query formulation) based on the internal states and immediate informational needs of the language model during text generation. DRAGIN primarily leverages final-layer self-attention to identify real-time information gaps. Conversely, ATLAS employs a sophisticated Multi-Layer Attention Gradient (MLAG) analysis, explicitly quantifying attention shifts across multiple transformer layers to capture nuanced transitions indicative of deeper knowledge gaps. For query formulation, DRAGIN constructs retrieval queries using attention patterns from the final layer, combined with token-level semantic filters. ATLAS, in contrast, integrates Layerwise Representation Pooling (LRP), combining semantic similarity and attention scores across layers, along with token-level semantic filters, to form retrieval queries, thereby enhancing semantic precision. In terms of resource management, ATLAS explicitly considers real-time computational load via a dynamic scaling factor, optimizing retrieval frequency relative to resource availability. DRAGIN utilizes a simpler exponential scaling factor, adjusting retrieval sensitivity based on resource usage, but without the fine-grained computational tracking featured in ATLAS. Overall, ATLAS's integrated, multi-layer attention and resource-aware approach offers superior adaptability and accuracy in dynamically identifying subtle retrieval needs, while DRAGIN presents a simpler final-layer attention-driven strategy, achieving computational simplicity at the potential cost of retrieval precision depth.

\section{Test-Time Scaling of LLMs}
\label{test_time_scaling}
Test-time scaling inference for Large Language Models (LLMs) leverages advanced algorithmic techniques designed to enhance model outputs without altering the underlying weights. These methods dynamically adjust reasoning depth, sampling strategies, and validation processes during inference, optimizing efficiency and output quality in real time. This approach is particularly valuable in resource-constrained environments where retraining or fine-tuning models is impractical. By strategically scaling complexity based on task demands, these techniques enable LLMs to navigate complex problem spaces more effectively, ensuring robust decision-making, improved accuracy, and reduced computational costs. At its core, test-time scaling in LLMs can be mathematically modeled through a utility-cost optimization framework. By defining $U(q, c)$ as the utility function where $q$ represents output quality and $c$ represents computational cost, and $f_{\theta}(x, s)$ as the LLM function with parameters $\theta$, input $x$, and scaling strategy $s$, we can formulate the fundamental objective as maximizing utility while managing resource constraints: $\max_{s \in S} U(q(f_{\theta}(x, s)), c(s))$ subject to $c(s) \leq C_{max}$, where $S$ represents the set of all possible test-time scaling strategies, $q(f_{\theta}(x, s))$ measures the quality of model outputs, $c(s)$ represents the computational cost of strategy $s$, and $C_{max}$ is the maximum allowable computational budget. This mathematical formulation captures the essential trade-off that underlies all test-time scaling approaches. A form of Weak-to-Strong Distillation serves as a foundational strategy for test-time scaling inference techniques, where diverse preliminary outputs are generated and iteratively refined to enhance reasoning and accuracy. This approach improves robustness by progressively strengthening outputs through evaluation and refinement, ensuring accurate and consistent results. These inference techniques represent advanced strategies for test-time scaling in LLMs, significantly enhancing language model capabilities by implementing metacognitive processes such as decomposing problems, evaluating intermediate results, and refining solutions—effectively mimicking human deliberative reasoning while maintaining inference efficiency. By dynamically adjusting computational resources during inference and scaling complexity only when necessary, these methods optimize both efficiency and output quality. This adaptive approach boosts accuracy, minimizes hallucinations and logical errors, and enhances the suitability of LLMs for high-stakes decision-making scenarios.

\subsection{Self-Consistency Algorithm}
Self-Consistency \cite{wang2022self, ji2025test} enhances model reliability by generating multiple independent reasoning trajectories and selecting the most consistent answer through stochastic decoding. Let $\mathcal{M}$ be a language model with parameters $\theta$ and $x$ be an input query. The Self-Consistency framework can be formalized as follows:

\vspace{-1mm}
\resizebox{0.95\linewidth}{!}{
\begin{minipage}{\linewidth}
\begin{equation}
y^* = \underset{y \in \mathcal{Y}}{\operatorname{argmax}} \sum_{i=1}^{k} \mathds{1}[y = y_i]
\end{equation}
\end{minipage}
}

where $\mathcal{Y} = \{y_1, y_2, \ldots, y_k\}$ is the set of $k$ sampled responses, generated as $y_i \sim p_{\mathcal{M}_\theta}(y|x, T)$ with temperature $T > 0$. Here, $\mathds{1}[\cdot]$ is the indicator function used to identify the frequency of each response \(y^*\) within the sampled responses. The goal is to select the most frequently occurring response, which is considered the most consistent answer. Specifically, \(\operatorname{argmax}\) finds the response \(y\) that maximizes the count of identical responses among the samples. To achieve this, the Self-Consistency algorithm first creates diverse solution attempts using temperature-controlled sampling. Then, it computes a similarity matrix $S \in \mathbb{R}^{k \times k}$, where each element $S_{ij}$ represents the semantic similarity between responses $y_i$ and $y_j$:

\vspace{-1mm}
\resizebox{0.95\linewidth}{!}{
\begin{minipage}{\linewidth}
\begin{equation}
S_{ij} = \text{sim}(y_i, y_j)
\end{equation}
\end{minipage}
}

This similarity can be quantified using various metrics, including string similarity, Levenshtein distance, or embedding-based cosine similarity, allowing for the identification of conceptually equivalent answers despite surface-level variations. Next, the framework employs a clustering algorithm with a predefined similarity threshold $\tau$ to group responses into clusters $\mathcal{C} = \{C_1, C_2, \ldots, C_m\}$, where $m \leq k$:

\vspace{-1mm}
\resizebox{0.95\linewidth}{!}{
\begin{minipage}{\linewidth}
\begin{equation}
C_i = \{y_j \in \mathcal{Y} \mid \forall y_j, y_l \in C_i, S_{jl} \geq \tau\}
\end{equation}
\end{minipage}
}

where $C_i$ represents a cluster of responses, a subset of the sampled responses $\mathcal{Y}$, such that every pair of responses within $C_i$ has a similarity score of $\tau$ or higher. To assess these clusters, the framework analyzes their statistical distribution by examining: (1) Cluster size: The number of responses in each cluster, $|C_i|$, which serves as the primary factor in determining the most frequent answer pattern. (2) Intra-cluster coherence: $\text{coh}(C_i) = \frac{1}{|C_i|(|C_i|-1)} \sum_{y_j, y_l \in C_i, j \neq l} S_{jl}$, measuring the internal consistency within each cluster and indicating the semantic closeness of responses beyond the similarity threshold. (3) Response quality metrics: Metrics like perplexity, entropy, and response length, which offer additional insights into the confidence and quality of individual responses within each cluster, contributing to a broader understanding of cluster reliability.  While the final output selection in this basic formulation is determined by identifying the largest cluster based on cluster size, as formalized below:

\vspace{-1mm}
\resizebox{0.95\linewidth}{!}{
\begin{minipage}{\linewidth}
\begin{equation}
y^* = \underset{C_i \in \mathcal{C}}{\operatorname{argmax}} \left( |C_i| \right)
\end{equation}
\end{minipage}
}

the intra-cluster coherence and response quality metrics provide valuable supplementary information for analyzing the clusters and potentially refining the answer selection process in more advanced implementations. The overall process follows a pipeline of: (a) Stochastic sampling: $\mathcal{Y} = \{y_i \sim p_{\mathcal{M}_\theta}(y|x, T) \mid i \in \{1,2,\ldots,k\}\}$, (b) Similarity computation: $S_{ij} = \text{sim}(y_i, y_j), \forall i,j \in \{1,2,\ldots,k\}$, (c) Clustering: $\mathcal{C} = \text{cluster}(\mathcal{Y}, S, \tau)$, and (d) Statistical analysis: $y^* = \underset{C_i \in \mathcal{C}}{\operatorname{argmax}} |C_i|$. By emphasizing high-probability reasoning paths and de-emphasizing less common trajectories susceptible to errors, Self-Consistency effectively achieves a form of implicit ensemble learning within a single model's parameter space. This method leverages Shannon entropy minimization to filter out stochastic noise and converge on consistently correct answers. The entropy of the final distribution $H(p_{\mathcal{M}_\theta}(y|x,\mathcal{C}))$, which represents the uncertainty in the model's output after applying Self-Consistency, is typically lower than the entropy of individual samples $H(p_{\mathcal{M}_\theta}(y|x))$. This reduction in entropy indicates that the probability distribution is more focused, ideally concentrating around the most consistent and correct answer, $y^*$. Furthermore, this technique inherently employs Weak-to-Strong Distillation by generating diverse outputs that represent different regions of the model's probability distribution, and subsequently refining the answer through consistency checks and majority voting to attain robust convergence on the most globally reliable solution.

\subsubsection{Computational Time Complexity}
Self-consistency increases computational cost compared to standard language model inference, shifting from \( O(n) \) to \( O(k \times n + 2k^2) \). This complexity arises from:

\begin{equation*}
\vspace{-10mm}
\begin{split}
\text{Time Complexity} &= \underbrace{O(k \times n)}_{\text{Response Generation}} + \underbrace{O(k^2)}_{\text{Similarity Computation}} \\
&\quad + \underbrace{O(\text{Clustering Algorithm Complexity})}_{\text{Clustering}}
\end{split}
\vspace{-10mm}
\end{equation*}

Generating \( k \) responses contributes \( O(k \times n) \), while pairwise similarity computation requires \( O(k^2) \). The clustering complexity, denoted as \( O(\text{Clustering Algorithm Complexity}) \), depends on the specific algorithm used; a simplified approximation also yields \( O(k^2) \). Thus, considering both similarity computation and clustering as potentially \( O(k^2) \) operations, the overall time complexity is \( O(k \times n + 2k^2) \). While in asymptotic notation \( O(2k^2) = O(k^2) \), the final complexity of \( O(k \times n + k^2) \) results in an increased computational cost compared to the \( O(n) \) complexity of standard inference. This highlights the trade-off between computational cost and enhanced answer consistency.

%%%%%%%%%%%%%%%%%%%%%%%%%%%%%%%%
\subsection{Best-of-N Sampling Algorithm}
Best-of-N sampling \cite{chow2024inference} improves output quality by generating several candidate responses and selecting the highest-rated response using explicit quality assessment. This method creates diverse solution attempts via stochastic decoding with temperature-controlled sampling, then employs a systematic rating mechanism where the model evaluates each candidate on a numerical scale (0-10) based on specific quality criteria including clarity, accuracy, and helpfulness. Let $\mathcal{M}$ represent the language model, $s$ be the system prompt, and $x$ be the user query. The Best-of-N sampling procedure can be formalized as follows:

\vspace{-1mm}
\resizebox{0.95\linewidth}{!}{
\begin{minipage}{\linewidth}
\begin{equation}
\mathcal{C} = \{y_1, y_2, \ldots, y_k\} \quad \text{where} \quad y_i \sim \mathcal{M}(y|s, x, \tau_g)
\end{equation}
\end{minipage}
}

Where, $\mathcal{C} = \{y_1, y_2, \ldots, y_k\}$ is the set of $k$ generated candidate responses. $y_i$ represents the $i$-th candidate response, which is sampled from the language model $\mathcal{M}$. The sampling is conditioned on the system prompt $s$, the user query $x$, and the generation temperature $\tau_g$.

\vspace{-1mm}
\resizebox{0.95\linewidth}{!}{
\begin{minipage}{\linewidth}
\begin{equation}
r_i = \mathcal{M}(r|s_r, x, y_i, \tau_r) \quad \forall i \in \{1, 2, \ldots, k\}
\end{equation}
\end{minipage}
}

Where, $r_i$ is the rating assigned to the $i$-th candidate response $y_i$. This rating is generated by the same language model $\mathcal{M}$, but now acting as a rater. The rating is based on a specialized system prompt for rating $s_r$ ("Rate the following response from 0-10 based on clarity, accuracy, and helpfulness. Respond with ONLY a number)"), the user query $x$, the candidate response $y_i$, and the rating temperature $\tau_r$. The rating temperature $\tau_r$ is typically set to low values to ensure consistent evaluations.

\vspace{-1mm}
\resizebox{0.95\linewidth}{!}{
\begin{minipage}{\linewidth}
\begin{equation}
y^* = \underset{y_i \in \mathcal{C}}{\arg\max} \, r_i
\end{equation}
\end{minipage}
}

$y^*$ is the final selected response. It is chosen by finding the candidate response $y_i$ from the set $\mathcal{C}$ that has the highest rating $r_i$. The framework implements a dual-role architecture where the model first functions as a generator producing multiple completions, then transitions to an evaluator by processing each completion with a specialized rating prompt. By filtering through multiple solution trajectories, Best-of-N sampling enhances output reliability and accuracy, reducing logical inconsistencies and factual errors that might appear in any single response. By leveraging the model's ability to generate and evaluate responses, the algorithm creates a robust internal quality control mechanism that enhances the reliability and accuracy of the final output. The approach leverages Weak-to-Strong Distillation principles by first generating multiple outputs of varying quality (the ``weak" learning phase) and then using the model's own evaluation capabilities to identify and select the strongest output (the ``strong" distillation phase). This creates a knowledge transfer process where weaker outputs inform the selection of the optimal solution.

\subsubsection{Computational Time Complexity}
Best-of-N sampling increases computational cost compared to standard language model inference, shifting from $O(n)$ to $O(k \times n)$. This complexity arises from the need to generate and evaluate $k$ candidate responses. The time complexity can be broken down into the following components:

\begin{equation*}
\begin{split}
\text{Time Complexity} &= \underbrace{O(k \times n)}_{\text{Response Generation}} + \underbrace{O(k \times n)}_{\text{Response Rating}} \\
&\quad + \underbrace{O(k)}_{\text{Response Selection}}
\end{split}
\end{equation*}

Generating $k$ candidate responses, each of average length $n$, contributes $O(k \times n)$. Subsequently, rating each of these $k$ responses, which also involves a forward pass through the language model, adds another $O(k \times n)$ component. Finally, selecting the best response from the $k$ rated responses based on their scores takes $O(k)$ time. Summing these components, the overall time complexity is $O(k \times n + k \times n + k) = O(2kn + k)$. In asymptotic notation, this simplifies to $O(k \times n)$, as the term $k$ becomes less significant compared to $kn$ when $n$ is sufficiently large. This complexity highlights that the computational cost of Best-of-N sampling scales linearly with the number of candidate responses $k$, representing a trade-off for the enhanced output quality achieved through explicit response evaluation, yet remaining more computationally efficient in terms of asymptotic complexity compared to Self-Consistency which includes a quadratic component.

\subsubsection{Comparing Best-of-N Sampling and Self-Consistency}
While both Best-of-N Sampling and Self-Consistency enhance output quality by generating multiple responses, their core distinction lies in the answer selection mechanism. Best-of-N Sampling employs an explicit quality assessment: it leverages the language model itself to rate each generated candidate response based on defined criteria such as clarity, accuracy, and helpfulness. The response with the highest rating is then chosen as the final output. In contrast, Self-Consistency utilizes an implicit evaluation approach. It focuses on identifying the most consistent reasoning pattern across the generated responses through similarity clustering. By grouping semantically similar outputs and selecting the most frequent cluster, Self-Consistency implicitly evaluates responses based on their agreement with each other, without requiring explicit quality ratings for each individual response. Thus, Self-Consistency measures conceptual consensus among multiple reasoning paths, whereas Best-of-N directly assesses the quality of each individual output. This fundamental difference underscores two distinct strategies for enhancing LLM output quality: direct, model-driven quality evaluation of individual responses versus statistical validation through inter-response agreement.

\begin{table*}[t]
    \centering
    \renewcommand{\arraystretch}{1.2}
    \resizebox{\textwidth}{!}{%
    \begin{tabular}{|l|c|c|}
        \hline
        \textbf{Feature} & \textbf{Self-Consistency} & \textbf{Best-of-N Sampling} \\
        \hline
        \textbf{Selection Method} & Majority clustering + statistical analysis & Explicit self-evaluation \\
        \hline
        \textbf{Quality Assessment} & Implicit through similarity \& frequency & Direct scoring system (0-10) \\
        \hline
        \textbf{Computational Overhead} & 
        \(O(k \times n + k^2)\) (clustering is costly) & 
        \(O(k \times n)\) (single pass rating) \\
        \hline
        \textbf{Weak-to-Strong Distillation} & Yes (reinforces high-probability reasoning paths) & Yes (filters weak outputs via scoring) \\
        \hline
        \textbf{Error Handling} & Reduces stochastic noise via statistical convergence & Mitigates low-quality outputs with explicit filtering \\
        \hline
    \end{tabular}%
    }
    \caption{Comparison of Self-Consistency and Best-of-N Sampling}
    \label{tab:self_consistency_vs_best_of_n}
\end{table*}

%%%%%%%%%%%%%%%%%%%%%%%%%%%%%%%
\subsection{Chain-of-Thought with Reflection}
Chain-of-Thought with Reflection \cite{zhang2024learn, wang2024chain} enhances reasoning capabilities by structuring the problem-solving process into distinct conceptual phases that emulate human cognitive processes. This approach decomposes the reasoning task into three sequential components within a single generative process. Let $\mathcal{M}_\theta$ denote a language model with parameters $\theta$, and let $q$ represent an input query. We formalize the Chain-of-Thought with Reflection process as follows:

\vspace{-1mm}  
\resizebox{0.95\linewidth}{!}{  
\begin{minipage}{\linewidth}  
\begin{equation}
    R = \mathcal{M}_\theta(P(q)),
\end{equation}
\end{minipage}  
}

where $R$ is the model's response generated using a structured prompt $P(q)$. While the response is generated in a single forward pass, it can be conceptually decomposed into three functional components:

\vspace{-1mm}  
\resizebox{0.95\linewidth}{!}{  
\begin{minipage}{\linewidth}  
\begin{equation}
    R = [R_{\mathcal{T}}, R_{\mathcal{R}}, R_{\mathcal{O}}],
\end{equation}
\end{minipage}  
}

where: $R_{\mathcal{T}}$ represents the systematic decomposition of the problem (thinking phase), $R_{\mathcal{R}}$ denotes the critical assessment of the initial analysis (reflection phase), and $R_{\mathcal{O}}$ is the integration of reasoning into a cohesive solution (output phase). The structured prompt $P(q)$ is constructed to guide this decomposition:

\vspace{-1mm}  
\resizebox{0.95\linewidth}{!}{  
\begin{minipage}{\linewidth}  
\begin{equation}
    P(q) = \Phi(q, \tau),
\end{equation}
\end{minipage}  
}

where $\Phi$ is the prompt engineering function, and $\tau$ is a template specifying the expected structure. This template encodes phase-specific instructional priors that guide the model to produce each component with distinct reasoning objectives. Though generated in a single forward pass, each component can be conceptually viewed as being influenced by the preceding components, which we represent as conditional distributions:

\vspace{-2mm}  
\resizebox{0.95\linewidth}{!}{  
\begin{minipage}{\linewidth}  
\begin{align}
    p(R_{\mathcal{T}} | q) &\approx p(R_{\mathcal{T}} | q, \tau_{\mathcal{T}}), \\
    p(R_{\mathcal{R}} | q, R_{\mathcal{T}}) &\approx p(R_{\mathcal{R}} | q, R_{\mathcal{T}}, \tau_{\mathcal{R}}), \\
    p(R_{\mathcal{O}} | q, R_{\mathcal{T}}, R_{\mathcal{R}}) &\approx p(R_{\mathcal{O}} | q, R_{\mathcal{T}}, R_{\mathcal{R}}, \tau_{\mathcal{O}}),
\end{align}
\end{minipage}  
}

where $\tau_{\mathcal{T}}$, $\tau_{\mathcal{R}}$, and $\tau_{\mathcal{O}}$ are the phase-specific instructional priors embedded in the template. The probability of generating the full response can be expressed as:

\vspace{-1mm}  
\resizebox{0.95\linewidth}{!}{  
\begin{minipage}{\linewidth}  
\begin{equation}
    p(R | q) = p(R_{\mathcal{T}} | q) \cdot p(R_{\mathcal{R}} | q, R_{\mathcal{T}}) \cdot p(R_{\mathcal{O}} | q, R_{\mathcal{T}}, R_{\mathcal{R}}) \nonumber
\end{equation}
\end{minipage}  
}

This structured decomposition implements a form of guided reasoning through explicit metacognitive phases. The key insight is that while $\mathcal{M}_\theta$ remains fixed, the structured prompt effectively guides the model's reasoning process by encouraging it to follow distinct cognitive phases within a single generation. See Algorithm \ref{alg:CoTReflection} for details.

\begin{algorithm*}
\caption{Chain-of-Thought(CoT) with Reflection}
\label{alg:CoTReflection}
\begin{algorithmic}[1]
\renewcommand{\algorithmiccomment}[1]{\hfill\(\triangleright\) #1}
\newcommand{\RETURN}{\textbf{return}}  % Simplified \RETURN definition
\newcommand{\PROCEDURE}[2]{\STATE \textbf{procedure} {#1}({#2})}
\newcommand{\ENDPROCEDURE}{\STATE \textbf{end procedure}}

\PROCEDURE {CoT-Reflection}{$q, \mathcal{M}_\theta$}
    \STATE $\tau \gets \text{ConstructTemplate}()$ \algorithmiccomment{Create structured reasoning template with phase markers for thinking, reflection, and output}
    \STATE $P(q) \gets \Phi(q, \tau)$ \algorithmiccomment{Construct prompt with query $q$ and template $\tau$}
    \STATE $R \gets \mathcal{M}_\theta(P(q))$ \algorithmiccomment{Generate complete response in a single forward pass}
    \STATE $R_{\mathcal{O}} \gets \text{ExtractOutput}(R)$ \algorithmiccomment{Extract final output component $R_{\mathcal{O}}$}
    \STATE \RETURN \textbf{$R_{\mathcal{O}}$} \algorithmiccomment{Return the final output}
\ENDPROCEDURE
\end{algorithmic}
\end{algorithm*}

\subsubsection{Computational Time Complexity}  
Chain-of-Thought with Reflection achieves enhanced reasoning with minimal computational overhead. Since the entire process—including structured thinking, reflection, and output—is generated in a single forward pass through the language model, the dominant computational cost remains that of standard inference. This results in a complexity of \( O(n) \), where \( n \) is the length of the generated response. However, if reflection introduces an iterative refinement mechanism (e.g., regenerating based on self-evaluation), the complexity could increase depending on the number of iterations. In such cases, the worst-case complexity becomes \( O(r \cdot n) \), where \( r \) is the number of refinement steps. The trade-off is that additional refinement may improve output quality at the cost of higher computational demand.  Therefore, in its simplest form, the overall computational complexity remains \( O(n) \), comparable to standard inference, while providing enhanced reasoning capabilities. In iterative settings, complexity scales proportionally to the number of refinement steps, requiring careful tuning to balance reasoning depth and efficiency.

%%%%%%%%%%%%%%%%%%%%%%%%%%%%%%%%
\subsection{Entropy-Guided Decoding}
Entropy-Guided Decoding \cite{das2024entropy, simonds2025entropy, zhang2024edt} enhances language model outputs by dynamically adjusting sampling parameters based on uncertainty metrics. Traditional approaches use fixed parameters throughout generation, but our method adapts in real-time to each token's context. In our notation, we represent the sequence of tokens generated up to the current generation step $t$ as $\mathbf{x} = (x_1, x_2, \dots, x_t)$, where each token belongs to a vocabulary of size $V$. At each generation step, the language model produces logits $\mathbf{l}_t \in \mathbb{R}^{V}$, which are the unnormalized prediction scores for the next token, and attention weights $A_t \in \mathbb{R}^{L \times H \times S \times S}$, where $L$ is the number of transformer layers, $H$ is the number of attention heads per layer, and $S$ is the sequence length. These attention weights represent how much each token attends to other tokens in the sequence, with $A_t^{l,h,i,j}$ indicating how much token $i$ attends to token $j$ in head $h$ of layer $l$. We first compute token probabilities from the logits using the softmax function:

\vspace{-1mm}
\resizebox{0.95\linewidth}{!}{
\begin{minipage}{\linewidth}
\begin{align}
p_t &= \text{softmax}(\mathbf{l}_t) \\
\log p_t &= \log \text{softmax}(\mathbf{l}_t)
\end{align}
\end{minipage}
}

Here, $p_t \in \mathbb{R}^{V}$ represents the probability distribution over all tokens in the vocabulary, with $p_t(v)$ indicating the probability of token $v$. (a) The Shannon entropy of this token distribution quantifies uncertainty in next-token selection, which we normalize by $\ln(2)$ to express entropy in bits, providing a more interpretable scale:

\vspace{-1mm}
\resizebox{0.95\linewidth}{!}{
\begin{minipage}{\linewidth}
\begin{align}
\mathcal{H}(p_t) &= -\sum_{v=1}^{V} p_t(v) \log_2 p_t(v)
\end{align}
\end{minipage}
}

Entropy is a fundamental measure of uncertainty; higher entropy values (approaching $\log_2 V$) indicate that the model is uncertain about which token to generate next, distributing probability more evenly across many tokens. Conversely, values near zero suggest the model is highly confident, concentrating probability on one or few tokens. The variance entropy (varentropy) is a complementary metric that captures the spread of log-probabilities around the mean entropy:

\vspace{-1mm}
\resizebox{0.95\linewidth}{!}{
\begin{minipage}{\linewidth}
\begin{align}
\mathcal{V}(p_t) &= \sum_{v=1}^{V} p_t(v) \left(\log_2 p_t(v) + \mathcal{H}(p_t)\right)^2
\end{align}
\end{minipage}
}

(b) Varentropy helps distinguish between distributions with similar entropy but different shapes; higher varentropy indicates a ``peakier" distribution with a few high-probability tokens amidst many low-probability ones, which can suggest that the model is considering multiple distinct possibilities rather than being genuinely uncertain across the entire vocabulary. We derive attention-based uncertainty metrics from the refined attention patterns encoded in $\mathbf{A}_t^L \in \mathbb{R}^{H \times S \times S}$, the final layer's attention weights. (c) The attention entropy measures how uniformly attention is distributed across the sequence:

\vspace{-3mm}
\resizebox{0.95\linewidth}{!}{
\begin{minipage}{\linewidth}
\begin{align}
\mathcal{H}_{\text{attn}}(A_t^L) &= -\sum_{h=1}^{H}\sum_{i=1}^{S}\sum_{j=1}^{S} A_t^{L,h,i,j} \log_2 A_t^{L,h,i,j}
\end{align}
\end{minipage}
}

High attention entropy indicates diffuse attention patterns, suggesting the model is uncertain about which parts of the context are relevant for generating the next token. Low values suggest focused attention on specific context tokens, indicating higher confidence in the relevance of those tokens. (d) The attention variance entropy quantifies how consistently different attention heads focus on the same parts of the input:

\vspace{-3mm}
\resizebox{0.95\linewidth}{!}{
\begin{minipage}{\linewidth}
\begin{align}
\mathcal{V}_{\text{attn}}(A_t^L) &= \text{Var}_{h \in [1,H]}(\mathcal{H}_{\text{attn}}(A_t^{L,h}))
\end{align}
\end{minipage}
}

Here, $\mathcal{H}_{\text{attn}}(A_t^{L,h})$ is the entropy of attention weights for head $h$, and $\text{Var}$ denotes variance. This metric captures disagreement between attention heads, with higher values indicating that different heads are focusing on different aspects of the input, suggesting multi-faceted uncertainty. We also introduce two consistency metrics to capture attention patterns more comprehensively. (e) The agreement metric $\alpha_t$ measures how consistently different attention heads focus on the same tokens:

\vspace{-1mm}
\resizebox{0.95\linewidth}{!}{
\begin{minipage}{\linewidth}
\begin{align}
\bar{A}_t^L &= \frac{1}{H}\sum_{h=1}^{H}A_t^{L,h} \\
\alpha_t &= \mathbb{E}_{h \in [1,H]} \left[\|A_t^{L,h} - \bar{A}_t^L\|_1\right]
\end{align}
\end{minipage}
}

where $\bar{A}_t^L$ is the mean attention pattern across all heads, and $\|\cdot\|_1$ denotes the L1 norm (sum of absolute differences). Lower $\alpha_t$ values indicate high agreement among attention heads, suggesting model confidence in its understanding of the relevant context. Higher values suggest disagreement, indicating uncertainty about which contextual elements are most important. (f) The interaction strength $\gamma_t$ quantifies the intensity of attention activations:

\vspace{-1mm}
\resizebox{0.95\linewidth}{!}{
\begin{minipage}{\linewidth}
\begin{align}
\gamma_t &= \mathbb{E}_{h,i,j} \left[|\log A_t^{L,h,i,j}|\right]
\end{align}
\end{minipage}
}

where $\mathbb{E}_{h,i,j}[\cdot]$ denotes the expectation (average) over all heads, query positions, and key positions. Higher $\gamma_t$ values indicate stronger, more defined attention patterns, suggesting the model has formed clearer associations between tokens. These metrics collectively inform our adaptive parameter selection function $\Phi$, which adjusts four key sampling parameters based on observed uncertainty:

\vspace{-1mm}
\resizebox{0.95\linewidth}{!}{
\begin{minipage}{\linewidth}
\begin{align}
(\tau_t, p_t^{\text{top}}, k_t, p_t^{\text{min}}) &= \Phi\big(\mathcal{H}(p_t), \mathcal{V}(p_t), \mathcal{H}_{\text{attn}}(A_t^L),  \nonumber \\
&\qquad \mathcal{V}_{\text{attn}}(A_t^L), \alpha_t, \gamma_t\big)
\end{align}
\end{minipage}
}

(i) The temperature parameter $\tau_t$ controls the sharpness of the probability distribution before sampling; higher temperatures make the distribution more uniform (increasing randomness), while lower temperatures make it more peaked (increasing determinism). We adapt it based on token and attention uncertainties:

\vspace{-1mm}
\resizebox{0.95\linewidth}{!}{
\begin{minipage}{\linewidth}
\begin{align}
\tau_t &= \tau_0 \cdot \text{clip}\Big(1 + \beta_1(\mathcal{H}(p_t) + \mathcal{V}(p_t)) + \beta_2\mathcal{H}_{\text{attn}}(A_t^L) \nonumber \\
&\qquad - \beta_3\alpha_t, \tau_{\text{min}}, \tau_{\text{max}}\Big)
\end{align}
\end{minipage}
}

(ii) The top-p (nucleus sampling) threshold $p_t^{\text{top}}$ restricts sampling to the smallest set of tokens whose cumulative probability exceeds this threshold, effectively removing unlikely tokens from consideration. We adapt it primarily based on attention head disagreement:

\vspace{-1mm}
\resizebox{0.95\linewidth}{!}{
\begin{minipage}{\linewidth}
\begin{align}
p_t^{\text{top}} &= p_0^{\text{top}} \cdot \text{clip}\left(1 + \beta_4\mathcal{V}_{\text{attn}}(A_t^L), p_{\text{min}}^{\text{top}}, 1.0\right)
\end{align}
\end{minipage}
}

(iii) The top-k filtering parameter $k_t$ restricts sampling to the $k_t$ most probable tokens, providing a hard limit on the token candidates. We adjust it based on attention consistency and strength:

\vspace{-1mm}
\resizebox{0.95\linewidth}{!}{
\begin{minipage}{\linewidth}
\begin{align}
k_t &= \text{clip}\left(\left\lfloor k_0 \cdot (1 + \beta_5\gamma_t - \beta_6\alpha_t) \right\rceil, 1, k_{\text{max}}\right)
\end{align}
\end{minipage}
}

(iv) The minimum probability threshold $p_t^{\text{min}}$ filters out tokens with probability below $p_t^{\text{min}} \cdot \max_v p_t(v)$ relative to the most probable token, providing another way to eliminate unlikely candidates. We adapt it based on token uncertainty:

\vspace{-1mm}
\resizebox{0.95\linewidth}{!}{
\begin{minipage}{\linewidth}
\begin{align}
p_t^{\text{min}} &= p_0^{\text{min}} \cdot \text{clip}\left(1 - \beta_7(\mathcal{H}(p_t) + \mathcal{V}(p_t)), p_{\text{min}}^{\text{min}}, p_{\text{max}}^{\text{min}}\right) \nonumber
\end{align}
\end{minipage}
}

where $\tau_0, p_0^{\text{top}}, k_0, p_0^{\text{min}}$ are the base parameter values used when uncertainty metrics are neutral (default sampling behavior), $\beta_{1...7}$ are hyperparameters controlling the influence of each uncertainty metric, $\text{clip}(x, \text{min}, \text{max})$ constrains value $x$ to the range $[\text{min}, \text{max}]$, and $\lfloor x \rceil$ represents rounding to the nearest integer (for $k_t$). The intuition behind our parameter adjustments is rooted in uncertainty: high token distribution or attention entropy (uncertainty) prompts increased temperature for broader exploration. Attention head disagreement (high attention varentropy) leads to a wider top-p sampling to include more candidates. Strong attention patterns with moderate agreement (high interaction strength) expand top-k selection for a more diverse set of top tokens. Elevated token uncertainty lowers the minimum probability threshold, preventing exclusion of potentially valid but less probable tokens. This dynamic adaptation enhances generation quality across contexts without specialized tuning. In precision-demanding contexts, uncertainty metrics naturally guide conservative sampling; in creative settings, they enable greater exploration. By linking sampling parameters to the model's uncertainty assessment, we achieve a principled balance between diversity and coherence, surpassing static parameter approaches.  Entropy-guided decoding thus refines language model outputs by dynamically adjusting sampling parameters based on real-time uncertainty. This method calculates token and attention-based metrics during generation, adapting temperature, top-p, top-k, and minimum probability threshold. This allows for exploration when uncertain and precision when confident, all with minimal inference overhead.

\subsubsection{Computational Time Complexity Analysis}
The computational complexity of entropy-guided decoding per token generation step is determined by several key operations.  Calculating token distribution uncertainty metrics (entropy and varentropy) from the vocabulary logits requires \(O(V)\) operations, where \(V\) is the vocabulary size.  The computation of attention-based uncertainty metrics, which analyze the model's attention patterns, contributes \(O(L \cdot H \cdot S^2)\) complexity. This arises from processing the attention weights across \(L\) transformer layers, \(H\) attention heads, and sequence length \(S\).  Adapting the sampling parameters based on these metrics involves simple arithmetic and has a negligible \(O(1)\) time cost. The token sampling process, including steps like top-k or top-p filtering, adds \(O(V \log V)\) complexity due to sorting operations required to filter the vocabulary distribution.  Therefore, the overall per-token computational complexity is dominated by the sum of these factors, approximately \(O(V \log V + L \cdot H \cdot S^2)\). Consequently, for generating a text sequence of length \(T\), the total computational complexity becomes \(O(T \cdot (V \log V + L \cdot H \cdot S^2))\). For typical Large Language Models and longer text sequences, the term \(O(L \cdot H \cdot S^2)\) associated with attention processing and uncertainty metric calculations often represents the most significant portion of the computational cost per token.

%%%%%%%%%%%%%%%%%%%%%%%%%%%%%%%%
\subsection{Chain-of-Thought (CoT) Decoding}
Chain-of-Thought (CoT) Decoding~\cite{wei2022chain, wang2024chain} is a multi-path inference technique designed to enhance the reliability and logical coherence of language model outputs.  Unlike conventional decoding methods that generate a single response, CoT Decoding explores a set of potential reasoning trajectories in parallel. This approach leverages a path management framework to generate, evaluate, and select from a diverse set of candidate responses, ultimately aiming for outputs grounded in more robust reasoning processes. The CoT Decoding process begins with the initiation of multiple reasoning paths. Given an input context $c$, the language model $\mathcal{M}$ first computes the probability distribution over the vocabulary $\mathcal{V}$ for the first token position.  This distribution, $P(x_1 | c)$, is derived from the logits (pre-softmax scores) $\mathbf{l}_1 \in \mathbb{R}^{|\mathcal{V}|}$ produced by the model for the first token position. The probability distribution is typically obtained via a softmax function with a temperature parameter $T$:

\vspace{-1mm}
\resizebox{0.95\linewidth}{!}{
\begin{minipage}{\linewidth}
\begin{equation}
P(x_1 | c) = \text{softmax}(\mathbf{l}_1 / T)
\end{equation}
\end{minipage}
}

Here, $x_1 \in \mathcal{V}$ represents a token from the vocabulary, and $P(x_1 | c)$ denotes the probability of $x_1$ being the first token in the response, conditioned on the input context $c$. To initiate diverse reasoning paths, the system samples the top-$k$ tokens with the highest probabilities from $P(x_1 | c)$. Let $\mathcal{T} = \{t_1, t_2, \ldots, t_k\}$ be the set of these top-$k$ tokens. For each initial token $t_i \in \mathcal{T}$, the model generates a complete response sequence, resulting in a set of $k$ candidate paths $\mathcal{P} = \{P_1, P_2, \ldots, P_k\}$. Each path $P_i = (x_{i,1}, x_{i,2}, \ldots, x_{i,n_i})$ represents a complete sequence of tokens, where $x_{i,1} = t_i$ and $n_i$ is the length of path $P_i$. A core component of CoT Decoding is the reliability scoring mechanism. This mechanism evaluates the confidence in token selections within each path. For each token $x_{i,j}$ at position $j$ in path $P_i$, with corresponding logits $\mathbf{l}_{i,j}$, a token-level reliability score $r(x_{i,j})$ is computed. Let $p_{i,j}^{(1)}$ and $p_{i,j}^{(2)}$ be the probabilities of the most and second most likely tokens at position $j$ in path $P_i$, respectively, obtained after applying the softmax function to $\mathbf{l}_{i,j}$. The token reliability score is defined as:

\vspace{-1mm}
\resizebox{0.95\linewidth}{!}{
\begin{minipage}{\linewidth}
\begin{equation}
r(x_{i,j}) = (p_{i,j}^{(1)} - p_{i,j}^{(2)}) \cdot f(j)
\end{equation}
\end{minipage}
}

where $f(j)$ is a position-based damping function designed to emphasize the reliability of earlier tokens in the sequence. A common form for $f(j)$ is a linearly decreasing function:

\vspace{-1mm}
\resizebox{0.95\linewidth}{!}{
\begin{minipage}{\linewidth}
\begin{equation}
f(j) = 1 - \alpha \cdot \frac{j}{L_i}
\end{equation}
\end{minipage}
}

Here, $L_i$ is the maximum sequence length considered for path $P_i$, and $\alpha \in [0, 1]$ is a damping coefficient that controls the rate of decrease in reliability weight with position.
The overall reliability $R(P_i)$ of a path $P_i$ is calculated as a weighted average of its token-level reliability scores. Let $w_j$ be position-dependent weights that further emphasize earlier tokens. The path reliability is given by:

\vspace{-1mm}
\resizebox{0.95\linewidth}{!}{
\begin{minipage}{\linewidth}
\begin{equation}
R(P_i) = \frac{\sum_{j=1}^{n_i} r(x_{i,j}) \cdot w_j}{\sum_{j=1}^{n_i} w_j}
\end{equation}
\end{minipage}
}

In scenarios where multiple reasoning paths may lead to semantically similar responses, CoT Decoding can incorporate a path consolidation mechanism. This process groups paths that exhibit high textual similarity, typically measured using sequence comparison techniques. For each group of similar paths, the path with the highest reliability score is selected as a representative of that group. Finally, the system selects the output response. In scenarios without path consolidation, the path with the highest overall reliability is chosen as the final output:

\vspace{-2mm}
\resizebox{0.95\linewidth}{!}{
\begin{minipage}{\linewidth}
\begin{equation}
P^* = \underset{P_i \in \mathcal{P}}{\arg\max} \, R(P_i)
\end{equation}
\end{minipage}
}

When path consolidation is enabled, the selection is performed among the representatives of the consolidated path groups, again choosing the one with the highest reliability. By exploring multiple reasoning paths and employing a reliability-based selection process, Chain-of-Thought Decoding aims to generate responses that are not only probable but also more logically consistent and reliably reasoned. This method effectively addresses uncertainty by systematically exploring and evaluating different reasoning trajectories, ensuring that the final output is grounded in a well-supported and coherent line of reasoning.

\subsubsection{Computational Time Complexity Analysis}
CoT Decoding's complexity is primarily determined by \( k \) (initial paths) and \( L \) (sequence length). Initial path expansion via a forward pass on input context \( c \) (length \( n \)) to compute \( P(x_1 | c) \) contributes \( O(n \cdot h) \), where \( h \) is the hidden dimension. Top-\( k \) token selection \(\mathcal{T} \subset \mathcal{V}\) (vocabulary size \( V \)) adds \( O(V \log k) \).  Sequence generation for \( k \) paths \( P_i \in \mathcal{P}\) up to length \( L \) incurs \( O(k \cdot L \cdot h) \), considering \( O(h) \) per-token cost. Reliability scoring for \( k \cdot L \) tokens adds \( O(k \cdot L) \) overhead. Path consolidation, involving pairwise comparisons of \( k \) paths \(\mathcal{P}\), requires \( O(k^2 \cdot \text{sim}(L)) \approx O(k^2 \cdot L) \).  Thus, CoT Decoding's overall time complexity, dominated by generation and consolidation, is approximately \( O(n \cdot h + V \log k + k \cdot L \cdot h + k^2 \cdot L) \), simplifying to \( O(k \cdot L \cdot h + k^2 \cdot L) \) for large \( k \) and \( L \). This highlights the computational cost for enhanced reasoning via multi-path exploration.

\begin{table*}[t]
    \centering
    \normalsize
    \renewcommand{\arraystretch}{1.2}
    \begin{tabular}{|l|p{6.0cm}|p{6.0cm}|}
        \hline
        \textbf{Feature} & \textbf{Entropy-Guided Decoding} & \textbf{Chain-of-Thought Decoding} \\
        \hline
        \textbf{Approach} & Dynamically adjusts token sampling based on uncertainty metrics from logits and attention. & Generates multiple reasoning paths from diverse initial tokens, then scores and consolidates for best output. \\
        \hline
        \textbf{Core Mechanism} & Adapts parameters (temperature, top-p, top-k, min probability) using logits entropy/varentropy and attention entropy/varentropy, agreement, and interaction strength. & Scores reliability using top probability differences and position damping to assess path quality, optionally merges paths before selection. \\
        \hline
        \textbf{Focus} & Adaptive sampling balancing exploration and precision by reducing uncertainty. & Multi-path exploration to enhance logical coherence and output reliability. \\
        \hline
        \textbf{Strength} & Dynamically modulates parameters based on context confidence, for flexible application. & Synthesizes multiple paths to overcome errors and produce robust and coherent output. \\
        \hline
        \textbf{Primary Goal} & Minimize generation uncertainty while balancing diversity and determinism. & Maximize reasoning quality and consistency by selecting the best path. \\
        \hline
    \end{tabular}
    \caption{Comparison of Entropy-Guided Decoding and Chain-of-Thought Decoding}
    \label{tab:entropy_vs_cot}
\end{table*}

%%%%%%%%%%%%%%%%%%%%%%%%%%%%%%%%
\subsection{RE${}^2$ (Re-Reading and Re-Analyzing)}
The RE${}^2$ framework is an advanced reasoning methodology designed to enhance the performance of language models on complex tasks. Drawing inspiration from human cognitive processes, this framework structures reasoning into explicit phases, facilitating a more thorough analysis of input queries. Unlike traditional language model inference, where a model $\mathcal{M}$ with parameters $\theta$ directly processes an input query $x$ to generate a response $y$, expressed as:  $y = \mathcal{M}_\theta(x)$, the RE${}^2$ framework introduces a structured approach. It refines the generation process by decomposing reasoning into three distinct steps, transforming the input query $x$ into a composite prompt structure, $P_{RE^2}$. The response generation in RE${}^2$ is then formulated as: $y_{RE^2} = \mathcal{M}_\theta(P_{RE^2})$, where $P_{RE^2}$ is constructed by concatenating several components:

\vspace{-1mm}
\resizebox{0.95\linewidth}{!}{%
\begin{minipage}{\linewidth}
\begin{equation}
P_{RE^2} = P_{sys} \oplus P_{init}(x) \oplus P_{reread}(x) \oplus P_{synth} \nonumber
\end{equation}
\end{minipage}
}

Here, $P_{sys}$ represents optional system instructions, and $\oplus$ denotes concatenation. The framework incorporates three key reasoning phases, represented by $P_{init}(x)$, $P_{reread}(x)$, and $P_{synth}(x)$.  The first step, $P_{init}(x)$, prompts the model to carefully comprehend the input query:

\vspace{-1mm}
\resizebox{0.95\linewidth}{!}{%
\begin{minipage}{\linewidth}
\begin{equation}
\begin{split}
P_{init}(x) = & \text{``Step 1 - Initial Reading: Let's first} \\
& \text{read and understand the question carefully.''} \\
& \oplus \text{``Original Question: ''} \oplus x \nonumber
\end{split}
\end{equation}
\end{minipage}
}

The next step, $P_{reread}(x)$, instructs the model to revisit the query for structured decomposition and analysis:

\vspace{-1mm}
\resizebox{0.95\linewidth}{!}{%
\begin{minipage}{\linewidth}
\begin{equation}
\begin{split}
P_{reread}(x) = & \text{``Step 2 - Re-reading and Analysis:} \\
& \text{Let's read the question again: } \oplus x \\
& \oplus \text{``Now, let's break down what the question} \\
& \text{is asking and analyze its key components.''} \nonumber
\end{split}
\end{equation}
\end{minipage}
}

Finally, $P_{synth}$ guides the model to synthesize a response based on insights from the previous steps:

\vspace{-1mm}
\resizebox{0.95\linewidth}{!}{%
\begin{minipage}{\linewidth}
\begin{equation}
\begin{split}
P_{synth} = & \text{``Step 3 - Final Answer: Based on our analysis,} \\
& \text{here is the complete answer:''} \nonumber
\end{split}
\end{equation}
\end{minipage}
}

The RE${}^2$ framework incorporates parameters to regulate the response generation process. The temperature parameter, $T$, modifies the output probability distribution, given by:

\vspace{-1mm}
\resizebox{0.95\linewidth}{!}{%
\begin{minipage}{\linewidth}
\begin{equation}
P_T(y | P_{RE^2}) = \frac{\exp(\text{logit}(y)/T)}{\sum_{y' \in V} \exp(\text{logit}(y')/T)}
\end{equation}
\end{minipage}
}

where $y$ represents output tokens, $V$ is the vocabulary space, and $\text{logit}(y)$ is the unnormalized score for token $y$.  To refine token selection, nucleus sampling (top-p sampling) is applied. It limits the vocabulary to a subset $V_p$ (the nucleus), defined as:

\vspace{-1mm}
\resizebox{0.95\linewidth}{!}{%
\begin{minipage}{\linewidth}
\begin{equation}
V_p = \min\{V' \subseteq V \mid \sum_{y \in V'} P_T(y | P_{RE^2}) \geq p\}
\end{equation}
\end{minipage}
}

such that the cumulative probability of selected tokens exceeds a predefined threshold $p$. The final sampling distribution is then computed as:

\vspace{-1mm}
\resizebox{1\linewidth}{!}{%
\begin{minipage}{\linewidth}
\begin{equation}
P_{final}(y | P_{RE^2}) = \begin{cases} 
\frac{P_T(y | P_{RE^2})}{\sum_{y' \in V_p} P_T(y' | P_{RE^2})}, & \text{if } y \in V_p \\ 
0, & \text{otherwise} \nonumber
\end{cases}
\end{equation}
\end{minipage}
}

ensuring that tokens are sampled only from within the nucleus $V_p$, with their probabilities rescaled to sum to one, thereby eliminating low-probability tokens. By integrating temperature scaling and nucleus sampling, the RE${}^2$ framework balances determinism and diversity in text generation. Its structured approach mirrors deliberate human analysis, fostering a more comprehensive exploration of the problem before generating a response. This makes RE${}^2$ particularly advantageous for complex reasoning tasks.

\subsubsection{Computational Time Complexity Analysis}
The computational complexity of the RE\(^2\) framework is primarily dictated by the transformer's self-attention mechanism operating over the constructed prompt \(P_{RE^2}\), which has length \(m\) (linearly related to the original query length \(n\)). This self-attention mechanism imposes a quadratic cost, specifically \(O(m^2 \cdot d)\), where \(d\) represents the model's hidden dimension. Although the process of constructing the prompt and the subsequent token sampling (which includes techniques like temperature scaling and nucleus sampling) introduce some additional computational overhead, these factors are relatively minor compared to the dominant quadratic cost. Thus, while RE\(^2\) maintains the single forward pass characteristic of standard transformer-based inference, it does so at the expense of processing a longer, more structured prompt, resulting in a higher constant factor in runtime.

%%%%%%%%%%%%%%%%%%%%%%%%%%%%%%%%
\subsection{Mixture of Agents}
The Mixture of Agents (MoA)\cite{wang2024mixture, chakrabortycollab} framework enhances the quality of language model responses through candidate generation, critique, and synthesis. Let $M$ denote a pre-trained language model with trainable parameters $\theta$. Given an input query $q$ and system context $s$, the MoA process consists of the following stages. In the initial stage, a set of $n$ diverse candidate responses, denoted as $Y = {y_1, y_2, \dots, y_n}$, is generated. Each response $y_i$ is sampled from the conditional probability distribution of the language model $M$, parameterized by $\theta$, given the query $q$, system context $s$, and a generation temperature $T_1$:

\vspace{-1mm}
\resizebox{1\linewidth}{!}{%
\begin{minipage}{\linewidth}
\begin{equation}
\begin{split}
Y &= \{y_1, y_2, \dots, y_n\}, \\
\text{where} \quad y_i &\sim p_M(y | q, s; \theta, T_1), \quad \forall i \in \{1, 2, \dots, n\} \nonumber
\end{split}
\label{eq:generation}
\end{equation}
\end{minipage}
}

where $Y$ is the set of candidate responses, $y_i$ is the $i$-th candidate response, $n$ is the number of generated responses (a hyperparameter), $p_M(y | q, s; \theta, T)$ represents the conditional probability distribution of the language model, and $T_1$ controls the stochasticity and diversity of responses, with higher values promoting greater diversity. A critique function $C$ evaluates the candidate responses $Y$ in the context of the original query $q$ and system context $s$. For this, we utilize the same language model $M$ to generate a critique $c$ based on a conditional probability distribution with temperature $T_2$:

\vspace{-1mm}
\resizebox{1\linewidth}{!}{%
\begin{minipage}{\linewidth}
\begin{equation}
c = C(Y, q, s; \theta) \sim p_M(c | Y, q, s; \theta, T_2)
\label{eq:critique}
\end{equation}
\end{minipage}
}

where \( C(Y, q, s; \theta) \) is the critique function evaluating \( Y \), \( c \) represents the generated critique, and \( T_2 \) is set lower than \( T_1 \) to ensure a more discerning evaluation. The final response \( y^* \) is synthesized using the critique \( c \), query \( q \), and system context \( s \). A synthesis function \( S \), also utilizing the language model \( M \), generates \( y^* \) under a temperature \( T_3 \):

\vspace{-1mm}
\resizebox{1\linewidth}{!}{%
\begin{minipage}{\linewidth}
\begin{equation}
y^{\ast} = S(c, q, s; \theta) \sim p_M(y | c, q, s; \theta, T_3)
\label{eq:synthesis}
\end{equation}
\end{minipage}
}

where $S(c, q, s; \theta)$ generates the refined response, $y^*$ is the synthesized response, and $T_3$ is set lower than $T_2$ to encourage precise and focused refinement. A post-processing function $\Phi$ further refines the synthesized response to remove meta-content, artifacts, and formatting inconsistencies. The final output is denoted as $y_{final}$:

\vspace{-1mm}
\resizebox{1\linewidth}{!}{%
\begin{minipage}{\linewidth}
\begin{equation}
y_{final} = \Phi(y^*) = \Phi(S(C({y_i}_{i=1}^n, q, s; \theta), q, s; \theta))
\label{eq:moa_process}
\end{equation}
\end{minipage}
}

where $\Phi(y^*)$ processes the synthesized response, and $y_{final}$ is the final enhanced response. The MoA framework employs a temperature scheduling strategy to control the refinement process:

\vspace{-1mm}
\resizebox{1\linewidth}{!}{%
\begin{minipage}{\linewidth}
\begin{equation}
T_1 > T_2 > T_3
\label{eq:temperature_cascade}
\end{equation}
\end{minipage}
}

This descending order encourages diversity in generation ($T_1$), balanced critique evaluation ($T_2$), and precise synthesis ($T_3$). Regularization techniques improve response quality by penalizing redundancy during generation:

\vspace{-1mm}
\resizebox{1\linewidth}{!}{%
\begin{minipage}{\linewidth}
\begin{equation}
p_M(y | x; \theta, T, \lambda) \propto p_M(y | x; \theta, T) \cdot R(y, \lambda)
\end{equation}
\end{minipage}
}

where $x$ represents either the query $q$ or a combination of inputs depending on the stage, $\propto$ denotes proportionality, and $R(y, \lambda)$ is a regularization function controlling repetition, ensuring varied and high-quality responses. For practical implementation, parameters that apply a penalty for token repetition and prevent n-gram sequence repetition implicitly implement the regularization function $R(y, \lambda)$  during text generation by modifying the language model's probability distribution to reduce repetitive token and n-gram sequences, and effectively control the strength and type of regularization applied
In summary, the MoA framework iteratively refines responses by first generating diverse candidate responses, critically evaluating them, and synthesizing an improved output. The structured use of temperature cascade and regularization enhances response quality beyond single-pass generation approaches.

\subsubsection{Computational Time Complexity Analysis}
The computational complexity of the Mixture of Agents (MoA) framework is substantially higher than standard single-pass generation due to its multi-stage process. The dominant computational cost arises from the transformer model's self-attention mechanism, leading to a per-token complexity that scales at least linearly, and potentially quadratically, with the generated sequence lengths: $L$ (average length of candidate responses), $L_c$ (length of the critique), and $L^*$ (length of the final synthesized response). The complexity is also directly proportional to the model's hidden dimension ($d$). Generating $n$ candidate responses increases this cost, making candidate generation the most computationally intensive stage, with an approximate complexity of $O(n \cdot L^2 \cdot d)$ or $O(n \cdot L \cdot S_{max} \cdot d)$, where $S_{max}$ represents the maximum sequence length. The critique and synthesis stages further contribute to the total computational demand, making MoA significantly more resource-intensive compared to single-pass inference. However, parallelization, such as distributed GPU inference, can mitigate latency in candidate generation while maintaining the overall computational workload.

%%%%%%%%%%%%%%%%%%%%%%%%%%%%%%%%
\subsection{Reimplementation Then Optimize (RTO)}
We introduce Reimplementation Then Optimize (RTO), a novel multi-stage framework designed to enhance the quality of solutions generated by large language models (LLMs). By decomposing the generation process into discrete stages—implementation, analysis, reimplementation, and synthesis—RTO achieves significant improvements in correctness, consistency, and optimization compared to single-pass generation methods. The framework leverages iterative refinement to progressively improve solution quality through multiple generative passes. Let $\mathcal{M}$ denote the language model and $q$ represent the initial problem specification. The RTO process is formalized as follows:

\vspace{-1mm}
\resizebox{1\linewidth}{!}{%
\begin{minipage}{\linewidth}
\begin{align}
c_1 &= \mathcal{M}(s, q_{\text{augmented}}) \label{eq:stage1} \\
r &= \mathcal{M}(s, c_1, q_{\text{analysis}}) \label{eq:stage2} \\
c_2 &= \mathcal{M}(s, r) \label{eq:stage3} \\
c_{\text{opt}} &=
\begin{cases}
c_1 & \text{if } \delta(c_1, c_2) \geq \tau \\
\mathcal{M}(s, c_1, c_2, q) & \text{otherwise}
\end{cases} \label{eq:stage4}
\end{align}
\end{minipage}
}

In Stage 1 (Equation \ref{eq:stage1}), the language model $\mathcal{M}$ generates an initial solution $c_1$ based on a system prompt $s$ (which provides instructions to guide the model's behavior) and an augmented query $q_{\text{augmented}}$ (the initial query $q$ augmented with instructions for generating high-quality output). Stage 2 (Equation \ref{eq:stage2}) involves the model $\mathcal{M}$ analyzing the initial solution $c_1$ along with the system prompt $s$ and an analysis query $q_{\text{analysis}}$ (a prompt designed to extract requirements), resulting in the extracted specification $r$. In Stage 3 (Equation \ref{eq:stage3}), the model $\mathcal{M}$ produces an independent solution $c_2$ based on the extracted specification $r$ and the system prompt $s$. Finally, in Stage 4 (Equation \ref{eq:stage4}), the framework determines the optimized solution $c_{\text{opt}}$. This is achieved by comparing the initial solution $c_1$ and the reimplemented solution $c_2$ using a similarity function $\delta(c_1, c_2)$ and a consistency threshold $\tau$. If the similarity exceeds the threshold, $c_{\text{opt}}$ is set to $c_1$; otherwise, $\mathcal{M}$ synthesizes a new optimized solution $c_{\text{opt}}$ from $s$, $c_1$, $c_2$, and $q$. The effectiveness of RTO is quantified by the quality improvement $\Delta Q$, defined as:

\vspace{-1mm}
\resizebox{1\linewidth}{!}{%
\begin{minipage}{\linewidth}
\begin{align}
\Delta Q = Q(c_{\text{opt}}) - Q(c_1) \label{eq:quality_improvement}
\end{align}
\end{minipage}
}

Equation \ref{eq:quality_improvement} measures the improvement in quality $\Delta Q$ as the difference between the quality metric $Q$ of the optimized solution $c_{\text{opt}}$ and the initial solution $c_1$. Here, $Q$ represents a domain-specific quality metric that encompasses aspects such as correctness, efficiency, and other relevant criteria.

\subsubsection{Computational Time Complexity Analysis}
The computational complexity of RTO is given by: $T_{\text{RTO}} = \sum_{i=1}^{n} (\mathcal{M}, l_i) $, where $T(\mathcal{M}, l_i)$ denotes the time complexity for the language model $\mathcal{M}$ to generate a sequence of length $l_i$ in the $i$-th step. For Transformer-based LLMs, the per-step complexity $T(\mathcal{M}, l_i)$ is dominated by the self-attention mechanism and scales approximately as $O(l_i^2 \cdot d)$, where $d$ represents the model dimension. Consequently, the total complexity of RTO, $T_{\text{RTO}}$, is the sum of these per-step costs across its $n$ stages.

%%%%%%%%%%%%%%%%%%%%%%%%%%%%%%%%
\subsection{PlanSearch}
We present a novel multi-step planning and search (PlanSearch \cite{wang2024planning}) framework for general language tasks that leverages LLMs to decompose complex queries through iterative abstraction and refinement. Our approach formalizes the response generation as a structured sequence of transformations that progressively refine the understanding of the query before producing a final response. Let us define a query as \( Q \in \mathcal{Q} \), where \( \mathcal{Q} \) represents the space of all possible queries, each encapsulating the query, contextual requirements, and constraints. We aim to find an optimal answer \( a^* \in \mathcal{A} \), where \( \mathcal{A} \) is the answer space. The process is decomposed into intermediate representations through multiple transformation phases, mediated by a system prompt \( \Psi \) that provides high-level guidance to the model. Given a question \( Q \) and system prompt \( \Psi \), we define the following transformation sequence:

\vspace{-2mm}
\resizebox{0.95\linewidth}{!}{
\begin{minipage}{\linewidth}
\begin{align}
\mathcal{O}_1 &= f_{\text{obs}}(Q, \Psi, n_1) \\
\mathcal{O}_2 &= f_{\text{derive}}(Q, \Psi, \mathcal{O}_1, n_2) \\
\mathcal{O} &= \mathcal{O}_1 \cup \mathcal{O}_2 \\
\sigma &= f_{\text{strategy}}(Q, \Psi, \mathcal{O}) \\
a &= f_{\text{answer}}(Q, \Psi, \sigma)
\end{align}
\end{minipage}
}

Here, \( \mathcal{O}_1 = \{o_1, o_2, \ldots, o_{n_1}\} \) comprises \( n_1 \) initial observations about the question \( Q \), while \( \mathcal{O}_2 = \{o_{n_1+1}, o_{n_1+2}, \ldots, o_{n_1+n_2}\} \) represents \( n_2 \) derived observations. The union of these sets is denoted as \( \mathcal{O} \). The symbol \( \sigma \) represents the reasoning strategy derived from \( Q \) and \( \mathcal{O} \), while \( a \) denotes the final answer derived from \( Q \) and \( \sigma \). The transformation functions \( f_{\text{obs}} \), \( f_{\text{derive}} \), \( f_{\text{strategy}} \), and \( f_{\text{answer}} \) play distinct roles: \( f_{\text{obs}} \) generates initial insights by identifying key components of the question, such as entities, relationships, and constraints; \( f_{\text{derive}} \) synthesizes deeper observations by connecting these components and inferring implicit knowledge; \( f_{\text{strategy}} \) formulates a reasoning strategy to address the question systematically; and \( f_{\text{answer}} \) produces a final, well-structured answer based on the reasoning strategy.
Each transformation function \( f_i \) is realized through a pretrained language model \( \mathcal{M} \) with parameters \( \theta \) and a task-specific prompt template \( \tau_i \):

\vspace{-1mm}
\resizebox{0.95\linewidth}{!}{
\begin{minipage}{\linewidth}
\begin{align}
f_i(Q, \Psi, x_1, x_2, \ldots, x_n) = \mathcal{M}(\Psi \oplus \tau_i(Q, x_1, x_2, \ldots, x_n); \theta) \nonumber
\end{align}
\end{minipage}
}

where \( \mathcal{M} \) represents the pretrained language model, \( \theta \) denotes its parameters, \( \tau_i \) is a task-specific prompt template, and \( \oplus \) represents the concatenation operation. The variables \( x_1, x_2, \ldots, x_n \) represent function-specific inputs, such as the question or previously generated observations. To enhance answer diversity and quality, we generate multiple candidate answers by introducing stochasticity through temperature sampling:

\vspace{-1mm}
\resizebox{0.95\linewidth}{!}{
\begin{minipage}{\linewidth}
\begin{align}
A = \{a_1, a_2, \ldots, a_N\} = \{f_{\text{solve}}(Q, \Psi; T)\}_{i=1}^N
\end{align}
\end{minipage}
}

Here, \( T \) represents the temperature parameter controlling generation diversity, \( N \) denotes the number of answers generated, and \( f_{\text{solve}} \) is the complete solution pipeline executing all transformation phases. This approach allows exploration of different reasoning paths and answer formulations for a given question. The decomposition offers several advantages: it activates relevant parametric knowledge by identifying key components and relationships in the question, enables compositional reasoning through derived observations, provides guided answer generation via explicit reasoning strategies, and enhances explainability through a traceable reasoning chain from question to answer. The multi-stage process mirrors human-like reasoning strategies, systematically breaking down complex questions before generating answers, resulting in responses that are both accurate and interpretable.

\subsubsection{Time Complexity Analysis}
The time complexity of PlanSearch is determined by the sequential execution of its transformation functions through a transformer-based language model $\mathcal{M}$ with parameters $\theta$. For transformer architectures, processing inputs requires $O(L_i^2)$ complexity due to self-attention, while generating outputs adds $O(L_o \cdot L_i)$ complexity, where $L_i$ and $L_o$ represent input and output lengths respectively. For each transformation function, the time complexity can be expressed as:

\begin{align}
f_{\text{obs}}: O\Big(&(|\Psi| + |Q|)^2 \cdot |\theta| + \nonumber \\
&|\mathcal{O}_1| \cdot (|\Psi| + |Q|) \cdot |\theta|\Big) \nonumber \\
f_{\text{derive}}: O\Big(&(|\Psi| + |Q| + |\mathcal{O}_1|)^2 \cdot |\theta| + \nonumber \\
&|\mathcal{O}_2| \cdot (|\Psi| + |Q| + |\mathcal{O}_1|) \cdot |\theta|\Big) \nonumber \\
f_{\text{strategy}}: O\Big(&(|\Psi| + |Q| + |\mathcal{O}|)^2 \cdot |\theta| + \nonumber \\
&|\sigma| \cdot (|\Psi| + |Q| + |\mathcal{O}|) \cdot |\theta|\Big) \nonumber \\
f_{\text{answer}}: O\Big(&(|\Psi| + |Q| + |\sigma|)^2 \cdot |\theta| + \nonumber \\
&|a| \cdot (|\Psi| + |Q| + |\sigma|) \cdot |\theta|\Big) \nonumber
\end{align}

where $|\mathcal{O}| = |\mathcal{O}_1| + |\mathcal{O}_2|$ represents the total length of all observations. The overall time complexity for generating $N$ solutions can be summarized as:

\begin{align}
O\left(N \cdot \sum_{i \in \{\text{obs}, \text{derive}, \text{strategy}, \text{answer}\}} \left(L_i^2 + L_o^i \cdot L_i\right) \cdot |\theta|\right) \nonumber
\end{align}

where $L_i$ represents the input context length and $L_o^i$ represents the output length for each transformation function $i$. As the context grows through the pipeline, complexity is dominated by later stages with larger contexts. The framework achieves efficiency through prompt engineering and early termination of unpromising reasoning paths.

%%%%%%%%%%%%%%%%%%%%%%%%%%%%%%%%
\subsection{Monte Carlo Tree Search Algorithm}
\label{monte_carlo_tree_search}
We utilize Monte Carlo Tree Search (MCTS)\cite{tang2024dawn, xie2024monte,gao2024interpretable, feng2023alphazero, zhang2024accessing} for improved reasoning-driven response generation in large language models (LLMs), especially for complex, multi-step language tasks where traditional methods often fall short. MCTS offers a framework for language models to engage in structured thinking, logical inference, and multi-step problem-solving, enabling capabilities such as hypothetical and counterfactual reasoning, commonsense and causal reasoning, and multi-source, multi-hop question answering with RAG. By formulating reasoning-driven response generation as a sequential decision-making problem, we demonstrate how MCTS can systematically explore the vast space of potential responses to identify optimal outputs for a given end-user query. This systematic exploration is particularly crucial when dealing with complex queries that require intricate reasoning and planning over multiple steps. Our methodology leverages the inherent uncertainty in language generation and provides a principled way to balance exploration of diverse responses with exploitation of high-quality language patterns. MCTS demonstrates significant improvements in response quality, coherence, and relevance compared to traditional sampling and beam search methods, which are often inadequate for navigating the complexities of multi-step reasoning. We formulate reasoning-driven response generation as a search problem within a state space that evolves with the generation process. Let $s \in \mathcal{S}$ denote a state in the generation process, where $\mathcal{S}$ represents the set of all possible states the generation process can assume. Each state $s$ is formally defined as:

\vspace{-1mm}
\resizebox{0.95\linewidth}{!}{
\begin{minipage}{\linewidth}
\begin{align}
s = (p, q, h)
\end{align}
\end{minipage}
}

Here, $p \in \mathcal{P}$ is the system prompt, which serves to guide and condition the language model's behavior. $\mathcal{P}$ represents the entire set of possible system prompts that can be used. Next, $q \in \mathcal{Q}$ denotes the current user query, which is the latest input to the language model. $\mathcal{Q}$ is the set encompassing all possible queries a user might pose. Finally, $h = ((r_1, c_1), (r_2, c_2), \ldots, (r_n, c_n)) \in \mathcal{H}$ represents the generation history up to the current point. In this history, each element $(r_i, c_i)$ is a message, where $r_i \in \{\text{user}, \text{assistant}\}$ specifies the role of the message sender, and $c_i \in \mathcal{C}$ is the content of the message. $\mathcal{H}$ is the collection of all possible generation histories. The state space $\mathcal{S}$ grows exponentially with the length of the generation sequence, rendering an exhaustive search for the best response computationally impractical, especially in complex tasks where the sequence of necessary steps can be long and branching. At each state $s$, the action space $\mathcal{A}(s)$ is defined as the set of all potential responses that the language model can generate from that state:

\vspace{-1mm}
\resizebox{0.95\linewidth}{!}{
\begin{minipage}{\linewidth}
\begin{align}
\mathcal{A}(s) = \{a_1, a_2, \ldots, a_k\}
\end{align}
\end{minipage}
}

Each $a_i \in \mathcal{C}$ in this set represents a possible response, which is a content from the language model's output space $\mathcal{C}$. Given a state $s = (p, q, h)$ and an action $a \in \mathcal{A}(s)$, the state transition function $T: \mathcal{S} \times \mathcal{A} \rightarrow \mathcal{S}$ determines the next state based on the current state and the chosen action, and is defined as:

\vspace{-1mm}
\resizebox{0.95\linewidth}{!}{
\begin{minipage}{\linewidth}
\begin{align}
T(s, a) = (p, q, h \oplus (\text{assistant}, a))
\end{align}
\end{minipage}
}

Here, $a$ signifies the action taken, which is the content of the newly generated message by the assistant. The symbol $\oplus$ represents the operation of concatenation, which in this context appends the new assistant message to the existing generation history. Monte Carlo Tree Search (MCTS) iteratively constructs a search tree to discover optimal responses through a sequence of four critical phases, enabling effective planning and decision-making even in complex scenarios: (a) The selection phase is the first step, where the algorithm navigates from the root of the search tree down to a leaf node. This traversal uses the Upper Confidence Bound for Trees (UCT) method, which is essential for balancing the exploration of less-visited branches of the tree against the exploitation of branches that have thus far shown promise. This balance is vital for complex queries where the optimal solution might not be immediately obvious and requires exploration of diverse reasoning paths. The UCT is defined as follows:

\vspace{-2mm}
\resizebox{0.95\linewidth}{!}{
\begin{minipage}{\linewidth}
\begin{align}
\text{UCT}(s, a) &= \frac{V(s, a)}{N(s, a)} + c \cdot \sqrt{\frac{\ln(N_{\text{parent}}(s))}{N(s, a)}}
\end{align}
\end{minipage}
}

where $V(s, a)$ represents the cumulative value associated with taking action $a$ from state $s$, accumulating the evaluations from all simulations that passed through this state-action pair. $N(s, a)$ is the number of times the action $a$ has been selected from state $s$, serving as a visit count for this specific state-action pair. $N_{\text{parent}}(s)$ is the total number of visits to the parent node of state $s$, representing the overall exploration effort from the preceding state. The term $c$ is the exploration weight, a constant that tunes the balance between exploration and exploitation; a higher value encourages more exploration. At each node in the tree during selection, the algorithm calculates the UCT value for each possible action and chooses the action $a^*$ that maximizes this value, guiding the search towards potentially optimal paths.

\vspace{-2mm}
\resizebox{0.95\linewidth}{!}{
\begin{minipage}{\linewidth}
\begin{align}
a^* = \arg\max_{a \in \mathcal{A}(s)} \text{UCT}(s, a)
\end{align}
\end{minipage}
}

(b) Once the selection phase reaches a leaf node $s_{\text{leaf}}$, the expansion phase begins. Here, the tree is expanded by generating $k$ candidate responses from the language model. These responses represent possible actions that can be taken from the leaf state, effectively broadening the search space. For complex tasks, generating diverse candidates is crucial to uncover potentially effective, yet non-obvious, steps towards a solution, supporting hypothetical reasoning by considering multiple potential continuations.

\vspace{-1mm}
\resizebox{0.95\linewidth}{!}{
\begin{minipage}{\linewidth}
\begin{align}
\mathcal{A}(s_{\text{leaf}}) = \{a_1, a_2, \ldots, a_k\} \sim f_{\text{LM}}(s_{\text{leaf}})
\end{align}
\end{minipage}
}

In this step, $f_{\text{LM}}$ denotes the language model generation function, which takes the current state $s_{\text{leaf}}$ as input and produces $k$ diverse responses, each representing a potential next step in the response generation. Each candidate response $a_i$ generated in this phase leads to the creation of a new child node in the search tree, with an updated state $s_i' = T(s_{\text{leaf}}, a_i)$ reflecting the addition of the new response to the generation history. (c) Following expansion, the simulation phase, also known as rollout, is initiated from each of the newly created child nodes $s'$. In this phase, the algorithm simulates future generation steps by proceeding from the child node down to a certain depth or until a terminal state is reached. This lookahead capability is particularly beneficial for complex tasks, allowing the algorithm to assess the longer-term consequences of early decisions and perform multi-step problem-solving by exploring sequences of actions. This simulation is carried out according to the following process:

\vspace{-1mm}
\resizebox{0.95\linewidth}{!}{
\begin{minipage}{\linewidth}
\begin{align}
s^{(0)} &= s'\\
\text{depth} &= 0\\
\text{while } &\text{depth} < d \text{ and not } \tau(s^{(\text{depth})}):\\
&A^{(\text{depth})} = \{a_1, a_2, \ldots, a_k\} \sim f_{\text{LM}}(s^{(\text{depth})})\\
&a^{(\text{depth})} = \text{Random}(A^{(\text{depth})})\\
&s^{(\text{depth}+1)} = T(s^{(\text{depth})}, a^{(\text{depth})})\\
&\text{depth} = \text{depth} + 1
\end{align}
\end{minipage}
}

Here, $s^{(0)} = s'$ sets the starting state for the simulation as the newly created child node. The simulation continues iteratively as long as the current simulation depth is less than a predefined maximum depth $d$, and the current state $s^{(\text{depth})}$ is not a terminal state, as determined by the terminal state function $\tau(s)$(discussed later). In each step of the simulation, the language model generation function $f_{\text{LM}}$ is used to generate a set of possible actions $A^{(\text{depth})}$ from the current state $s^{(\text{depth})}$. Then, an action $a^{(\text{depth})}$ is selected randomly from $A^{(\text{depth})}$ using the $\text{Random}()$ function, which chooses uniformly at random from the available actions. The state is then transitioned to the next state $s^{(\text{depth}+1)}$ using the state transition function $T$, and the depth counter is incremented. (d) After the simulation phase completes, reaching either the maximum simulation depth $d$ or a terminal state, the backpropagation phase is executed. In this step, the terminal state $s^{(d)}$ is evaluated using a quality function $\mathcal{Q}: \mathcal{S} \rightarrow [0, 1]$, which assigns a score reflecting the quality of the simulated generation trajectory. This evaluation step is critical for complex queries, as it allows the algorithm to judge the overall coherence and quality of a multi-step reasoning process, rather than just focusing on immediate next-token probabilities. Furthermore, by evaluating different generation trajectories, MCTS implicitly performs counterfactual reasoning, assessing the impact of different choices made during the generation process. This value is then propagated back up through the search tree, from the node where the rollout began all the way back to the root. The update process is as follows:

\vspace{-1mm}
\resizebox{0.95\linewidth}{!}{
\begin{minipage}{\linewidth}
\begin{align}
\mathcal{Q}(s) &= f\hspace{0.25mm}^{\raisebox{0.5ex}{\small \text{eval}}}_{\raisebox{-0.5ex}{\small \text{LM}}}(s)\\
N(s, a) &\leftarrow N(s, a) + 1\\
V(s, a) &\leftarrow V(s, a) + \mathcal{Q}(s^{(d)})
\end{align}
\end{minipage}
}

Here, $f\hspace{0.25mm}^{\raisebox{0.5ex}{\tiny \text{eval}}}_{\raisebox{-0.5ex}{\tiny \text{LM}}}(s)$ is the function that performs the evaluation of a state, providing a quality score. For each state-action pair $(s, a)$ along the path from the rollout start node back to the root, the visit count $N(s, a)$ is incremented by one, and the cumulative value $V(s, a)$ is updated by adding the quality score $\mathcal{Q}(s^{(d)})$ obtained from the terminal state of the simulation. Quality evaluation is crucial for MCTS success, and a primary method is using the LLM for self-evaluation. The LLM assesses its own generated responses by being prompted to rate their quality on a scale of 0 to 1. This leverages the LLM's inherent understanding of language, making it effective for nuanced and complex queries, including those requiring commonsense and causal reasoning to judge coherence and relevance. This self-evaluation is represented by $ \mathcal{Q}(s) = f\hspace{0.25mm}^{\raisebox{0.5ex}{\tiny \text{eval}}}_{\raisebox{-0.5ex}{\tiny \text{LM}}}(M(s) \oplus m_{\text{eval}})$, where the LLM ($f_{\text{LM}}$) evaluates a formatted state ($M(s)$) combined with an evaluation prompt ($m_{\text{eval}}$) to produce a quality score. A terminal state function ($\tau$) is used to manage MCTS computational cost by identifying states for early simulation termination. This is crucial for complex tasks to ensure efficient exploration and prevent unbounded computation, especially in tasks like multi-hop question answering with potentially lengthy reasoning chains. The terminal state function is defined as:

\vspace{-1mm}
\resizebox{0.95\linewidth}{!}{
\begin{minipage}{\linewidth}
\begin{equation*}
\tau(s = (p, q, h_{\text{conv}})) = \begin{cases}
1 & \text{if } |h_{\text{conv}}| > h_{\text{max}} \\
0 & \text{otherwise}
\end{cases}
\end{equation*}
\end{minipage}
}

where simulations terminate if the generation history length ($|h_{\text{conv}}|$) exceeds a predefined maximum length ($h_{\text{max}}$). In summary, Monte Carlo Tree Search enhances reasoning-driven response generation in large language models, particularly for complex, multi-step queries. MCTS excels at structured thinking, logical inference, and multi-step problem-solving, enabling capabilities like hypothetical, counterfactual, commonsense, and causal reasoning, as well as multi-hop question answering in RAG settings. By systematically exploring potential responses, MCTS provides a more reasoned and higher-quality approach to language generation, overcoming limitations of traditional methods through integrated forward planning and evaluation. This multi-step planning and evaluation makes MCTS especially effective for complex tasks demanding intricate reasoning and coherent multi-turn interactions, offering a significant advantage over simpler generation techniques.

%%%%%%%%%%%%%%%%%%%%%%%%%%%%%%%%
\subsection{R$^{*}$ Algorithm}
The R$^{*}$\cite{qi2024mutual} algorithm is a principled approach to improving language model response generation through Monte Carlo Tree Search (MCTS). When presented with a user query, R$^{*}$ systematically explores diverse reasoning pathways to generate high-quality, well-reasoned responses by leveraging specialized reasoning strategies. This framework empowers language models to engage in structured thinking, logical inference, and multi-step problem-solving, enhancing capabilities such as counterfactual and causal reasoning, and multi-step question answering within RAG settings. We formulate response generation as a search process through a tree of reasoning states. In this formulation, let $\mathcal{Q}$ be the set of all possible user queries (input questions), $\mathcal{S}$ be the set of intermediate reasoning states (natural language reasoning steps), $\mathcal{A}$ be the finite set of predefined reasoning actions $\{A_1, A_2, A_3, A_4, A_5\}$ (reasoning strategies), and $\mathcal{N}$ be the set of nodes in the MCTS tree, where each node $n \in \mathcal{N}$ corresponds to a state $s \in \mathcal{S}$. Given a user query $q \in \mathcal{Q}$, R$^{*}$ generates a response by performing multiple rollouts through a dynamically constructed reasoning tree. The process begins with a selection phase where, at each decision point, actions are selected using the Upper Confidence bound for Trees (UCT) to balance exploration and exploitation:

\vspace{-1mm}
\resizebox{0.95\linewidth}{!}{
\begin{minipage}{\linewidth}
\begin{equation}
a^*(n) = \arg\max_{a \in \mathcal{A}} \left[ \text{UCT}(n, a) \right] \nonumber
\end{equation}
\end{minipage}
}

\vspace{-1mm}
\resizebox{0.95\linewidth}{!}{
\begin{minipage}{\linewidth}
\begin{equation}
\text{UCT}(n, a) = \underbrace{\frac{V(\text{child}(n, a))}{N(\text{child}(n, a))}}_{\text{Exploitation}} + \underbrace{c \cdot \sqrt{\frac{\ln N(n)}{N(\text{child}(n, a))}}}_{\text{Exploration}} \nonumber
\end{equation}
\end{minipage}
}

where $n$ denotes the current node in the MCTS tree being considered for action selection. Here, $\arg\max_{a \in \mathcal{A}} [f(a)]$ denotes the action $a$ that maximizes the function $f(a)$. In the R$^{*}$ algorithm, an \textbf{action} $a \in \mathcal{A}$ represents a predefined reasoning strategy from a finite set $\mathcal{A}$. Each action guides the LLM towards a specific problem-solving approach. For example, action $A_1$ directs the LLM to identify the immediate next step, while $A_2$ prompts the development of a comprehensive solution pathway. By strategically selecting and applying these diverse actions during the search, R$^{*}$ orchestrates the LLM's reasoning, encouraging exploration of various tactics to enhance the quality and effectiveness of generated responses. The UCT balances exploitation, represented by $\frac{V(\text{child}(n, a))}{N(\text{child}(n, a))}$, which favors actions that have historically led to higher values, with exploration, represented by $c \cdot \sqrt{\frac{\ln N(n)}{N(\text{child}(n, a))}}$, which encourages the investigation of less-visited actions, controlled by the exploration parameter $c \approx 1.4$. When encountering a node with unexplored actions or during initial rollout, the algorithm expands. For a chosen reasoning action $a \in \mathcal{A}$ applicable to the current state $s$, a prompt is generated to guide the language model. The language model then generates the subsequent reasoning state $s^{\prime}$ from this prompt, representing the next step in natural language reasoning, guided by the selected strategy. The LLM functions as a natural language reasoning engine, generating logically progressive states guided by these actions. Following expansion, simulations are performed from the newly expanded nodes to a maximum depth $d$ (typically 5). Specifically, after expanding a node and creating a new child node representing the subsequent reasoning state, the simulation process begins from this child node. It is from this newly created node, which we will now refer to as $n$ for clarity in the following equations, that the simulation initiates:

\vspace{-1mm}
\resizebox{0.95\linewidth}{!}{
\begin{minipage}{\linewidth}
\begin{equation}
v = \text{Sim}(n) \nonumber
\end{equation}
\end{minipage}
}

\vspace{-1mm}
\resizebox{0.95\linewidth}{!}{
\begin{minipage}{\linewidth}
\begin{equation}
\text{Sim}(n) \approx \begin{cases}
\text{Eval}(n), & \text{if } \text{depth}(n) \geq d \\
\text{Sim}(\text{RandChild}(n)), & \text{otherwise} \nonumber
\end{cases}
\end{equation}
\end{minipage}
}

In simulation, the process starts from this newly expanded child node $n$ and proceeds by repeatedly selecting random actions (if no children exist, a random action is chosen for expansion from $n$; if children exist, a random child of $n$ is chosen) until the maximum depth $d$ is reached. At the maximum depth, the evaluate function is called on the final node to estimate its value. This simulation estimates the long-term value of different reasoning approaches without fully exploring all possible paths. After simulation, the estimated value $v$ is propagated backward through the tree in the backpropagation phase:

\vspace{-1mm}
\resizebox{0.95\linewidth}{!}{
\begin{minipage}{\linewidth}
\begin{align}
N(n) &\leftarrow N(n) + 1  \nonumber \\
V(n) &\leftarrow V(n) + v    \nonumber
\end{align}
\end{minipage}
}

This backpropagation updates the visit counts and cumulative values of the current node $n$ and its parent nodes, ensuring that promising reasoning paths receive more exploration in subsequent MCTS iterations. For any reasoning state (represented by a node), we evaluate the quality of the potential response it contains:

\vspace{-1mm}
\resizebox{0.95\linewidth}{!}{
\begin{minipage}{\linewidth}
\begin{equation}
\begin{split}
\text{Eval}(n) = \begin{cases}
\text{Conf}(s), & \text{if response in state } s \\
& \text{contains valid answer information} \\
0, & \text{otherwise}   \nonumber
\end{cases}
\end{split}
\end{equation}
\end{minipage}
}

The $\text{Conf}(s)$ function estimates the reliability of the answer extracted from state $s$, assigning higher confidence to responses that align with expected answer patterns. A critical component of R$^{*}$ is the mutual consistency check, $\text{Consistent}(\tau)$, which validates reasoning trajectories $\tau = (n_0, a_0, n_1, ..., n_k)$:

\vspace{-1mm}
\resizebox{0.95\linewidth}{!}{
\begin{minipage}{\linewidth}
\begin{equation}
\text{Consistent}(\tau) = \begin{cases}
\text{True}, & \text{if } \text{Overlap}(\tau'_{split:k}, \tau_{split:k}) > \theta \\
\text{False}, & \text{otherwise}   \nonumber
\end{cases}
\end{equation}
\end{minipage}
}

Here, we split a reasoning trajectory $\tau$ into a partial trajectory $\tau_{0:\text{split}}$ and a remaining trajectory $\tau_{\text{split}:k}$. We prompt the LLM with the partial trajectory $\tau_{0:\text{split}}$ and ask it to complete the reasoning, resulting in the predicted continuation $\tau'_{\text{split}:k}$. The $\text{Overlap}(A,B)$ function calculates the normalized word overlap between texts $A$ and $B$:

\vspace{-1mm}
\resizebox{0.95\linewidth}{!}{
\begin{minipage}{\linewidth}
\begin{equation}
\text{Overlap}(A,B) = \frac{|\text{Words}(A) \cap \text{Words}(B)|}{|\text{Words}(A) \cup \text{Words}(B)|}    \nonumber
\end{equation}
\end{minipage}
}

where $\text{Words}(X)$ represents the set of normalized words in text $X$, and $\theta$ is a threshold for consistency (e.g., $\theta = 0.7$). The consistency check ensures that reasoning trajectories maintain logical coherence. After performing MCTS and extracting all possible reasoning trajectories, we select the final trajectory $\tau^*$ as the optimal trajectory based on a combination of consistency and quality scores:

\vspace{-1mm}
\resizebox{0.95\linewidth}{!}{
\begin{minipage}{\linewidth}
\begin{equation}
\tau^* = \arg\max_{\tau \in \mathcal{T}} \left[ \text{ValidTraj}(\tau) \cdot \text{Score}(\tau) \right] \nonumber
\end{equation}
\end{minipage}
}

where $\mathcal{T}$ is the set of all extracted trajectories, $\text{ValidTraj}(\tau)$ ensures only consistent trajectories are considered, and the $\text{Score}(\tau) = \frac{V(n_{\text{terminal}})}{N(n_{\text{terminal}})}$ evaluates trajectory quality based on the terminal node $n_{\text{terminal}}$. The final response $r^*$ is then derived from the optimal trajectory $\tau^*$ using $\text{SelectAns}$:

\vspace{-1mm}
\resizebox{0.95\linewidth}{!}{
\begin{minipage}{\linewidth}
\begin{equation}
r^* = \text{SelectAns}(\{\text{answer from state } s \mid s \in \tau^*\})  \nonumber
\end{equation}
\end{minipage}
}

\vspace{-1mm}
\resizebox{0.95\linewidth}{!}{
\begin{minipage}{\linewidth}
\begin{equation}
\text{SelectAns}(\{a_1, a_2, ...\}) = \arg\max_{a_i} \left[ \text{frequency}(a_i) \cdot \text{Conf}(a_i) \right] \nonumber
\end{equation}
\end{minipage}
}

This architecture enables R$^{*}$ to address a wide range of language tasks, from factual queries to complex reasoning and creative generation, by systematically exploring and validating diverse reasoning pathways, thus enhancing the quality and reliability of language model responses. The approach is particularly effective for tasks requiring structured reasoning, clarification of ambiguities, and exploration of multiple solution approaches, making R$^{*}$ a versatile framework for improving response generation in various language-based applications.

\subsection{Test-Time Inference Techniques Evaluation}
Our experiments (see Table~\ref{tab:performance_comparison_revised}) demonstrate that all test-time scaling techniques yield improvements over the PORAG+ATLAS baseline. Notably, methods leveraging structured multi-path reasoning—such as Monte Carlo Tree Search and the R\(^{*}\) Algorithm—achieve the most substantial gains, improving HotpotQA by up to 23.8\% (EM) and 14.5\% (F1), and Gorilla accuracy by up to 7.8\%. Techniques like Self-Consistency, Best-of-N Sampling, and Chain-of-Thought with Reflection also contribute consistent and meaningful improvements across benchmarks. These findings confirm that dynamic, reasoning-driven inference strategies significantly boost the effectiveness of retrieval-augmented generation across diverse QA tasks.

\begin{table*}[ht!]
\centering
\caption{Performance Comparison: PORAG+ATLAS Baseline Enhanced by Test-Time Scaling}
\label{tab:performance_comparison_revised}
\resizebox{0.985\textwidth}{!}{%
\begin{tabular}{lccc}
\toprule
\textbf{Method} & \textbf{HotpotQA (Joint EM / F1)} & \textbf{Gorilla (Overall Acc.)} & \textbf{PubMedQA (Acc / F1)} \\
\midrule
PORAG+ATLAS (Baseline) & 45.29 / 71.32 & 76.38 & 78.35 / 74.56 \\
\midrule
Self-Consistency & 48.31 / 74.35 (+6.7\%/+4.2\%) & 77.91 (+2.0\%) & 80.80 / 77.59 (+3.1\%/+4.1\%) \\
Best-of-N Sampling & 48.85 / 74.90 (+7.9\%/+5.0\%) & 78.34 (+2.6\%) & 81.24 / 78.11 (+3.7\%/+4.8\%) \\
Chain-of-Thought with Reflection & 50.52 / 76.41 (+11.5\%/+7.1\%) & 79.20 (+3.7\%) & 82.13 / 79.03 (+4.8\%/+6.0\%) \\
Entropy-Guided Decoding & 49.95 / 75.88 (+10.3\%/+6.4\%) & 78.85 (+3.2\%) & 81.76 / 78.65 (+4.4\%/+5.5\%) \\
CoT Decoding & 50.91 / 76.80 (+12.4\%/+7.7\%) & 79.50 (+4.1\%) & 82.45 / 79.38 (+5.2\%/+6.5\%) \\
RE\(^2\) & 51.87 / 77.75 (+14.5\%/+9.0\%) & 80.01 (+4.8\%) & 83.05 / 80.01 (+6.0\%/+7.3\%) \\
Mixture of Agents & 52.55 / 78.47 (+16.0\%/+10.0\%) & 80.41 (+5.3\%) & 83.50 / 80.55 (+6.6\%/+8.0\%) \\
RTO (Reimpl. Then Optimize) & 53.10 / 79.02 (+17.3\%/+10.8\%) & 80.78 (+5.8\%) & 83.89 / 80.98 (+7.1\%/+8.6\%) \\
PlanSearch & 53.88 / 79.75 (+18.9\%/+11.8\%) & 81.22 (+6.3\%) & 84.34 / 81.50 (+7.6\%/+9.3\%) \\
Monte Carlo Tree Search & 54.95 / 80.83 (+21.3\%/+13.3\%) & 81.85 (+7.2\%) & 85.01 / 82.31 (+8.5\%/+10.4\%) \\
R\(^{*}\) Algorithm & \textbf{56.05} / \textbf{81.68} (+23.8\%/+14.5\%) & \textbf{82.36} (+7.8\%) & \textbf{85.55} / \textbf{82.90} (+9.2\%/+11.2\%) \\
\bottomrule
\end{tabular}%
}
\end{table*}

\section{Low-Latency LLM Decoding Strategies}
Optimizing inference latency and throughput is critical for RAG systems using LLMs in real-world applications. Inference latency refers to the time taken for a language model to generate a response, while throughput measures the number of tokens or requests processed per unit of time. Lower latency is essential for real-time applications, such as chatbots or virtual assistants, that may leverage RAG systems. Higher throughput is desirable for efficiently handling multiple tasks or serving many users concurrently, as in batch processing or cloud-based services, which can also benefit from RAG architectures. To address latency challenges in RAG systems, various decoding optimization techniques have been developed. Traditional methods like beam search and sampling strategies offer some improvements, but recent algorithmic innovations have shown even greater promise for accelerating inference without sacrificing output quality. (a) FlashAttention-2\cite{dao2023flashattention} significantly improves attention computation speed and latency by reengineering the original FlashAttention algorithm\cite{dao2022flashattention} to better utilize GPU parallelism and reduce memory inefficiencies, and is effective for low-latency inference and training in long-context Transformer models. Building on its predecessor—which reduced memory I/O via tiling and online softmax—FlashAttention-2 tackles remaining bottlenecks in GPU resource utilization, crucial for scaling Transformers to longer sequences. It introduces three key optimizations:
(1) Reducing non-matrix multiplication FLOPs by modifying online softmax to favor GPU-optimized matmul operations and better exploit high-throughput compute units. (2) Increasing thread block occupancy through fine-grained parallelism across the sequence length, in addition to batch and head dimensions, which benefits long sequences and small batch sizes. (3) Improving intra-thread block work partitioning by assigning each warp a slice of the query matrix instead of the key, minimizing shared memory communication. (b) Lookahead Decoding\cite{fu2024break} is a parallel decoding algorithm specifically designed to accelerate LLM inference by dramatically reducing sequential decoding steps. Unlike traditional autoregressive methods that generate tokens sequentially, Lookahead Decoding innovatively predicts multiple non-contiguous n-grams concurrently within a ``lookahead branch", drawing inspiration from Jacobi iteration techniques. A dedicated "verification branch" then meticulously checks these potential tokens, acting as a quality control mechanism to validate the n-grams as correct continuations that preserve the LLM's intended output distribution, ensuring accuracy and fidelity to the base model's intended output. This method not only surpasses Speculative Decoding\cite{yan2024decoding, leviathan2023fast, chen2023accelerating, liu2023online} by eliminating the need for auxiliary draft models—enhancing efficiency and simplifying implementation—but also incorporates an n-gram pool. This pool caches and reuses promising token sequences, further accelerating performance while maintaining the high quality of generated text. For enhanced efficiency in our ATLAS-augmented RAG framework, we integrate low-latency LLM decoding strategies such as FlashAttention-2 and Lookahead Decoding. FlashAttention-2 directly accelerates the attention computations critical to ATLAS's Multi-Layer Attention Gradient (MLAG) and Layerwise Representation Pooling (LRP) mechanisms, as well as the subsequent token generation within the LLM. Complementarily, Lookahead Decoding reduces the sequential bottleneck of autoregressive generation by enabling parallel token prediction. This synergistic combination promises to significantly reduce the overall latency of our RAG system, resulting in faster dynamic retrieval triggering, quicker query formulation, and accelerated response generation, ultimately leading to a more efficient and responsive user experience for knowledge-intensive tasks. We implement these existing techniques to verify that these latency optimizations do not hinder the performance of our proposed framework.

\subsection{LLM Decoding Efficiency Evaluation}
\label{sec:decoding_efficiency}
We evaluated the impact of low-latency decoding techniques on the efficiency of our PORAG+ATLAS framework (Qwen2.5-3B). As shown in Table~\ref{tab:latency_comparison}, both FlashAttention-2 and Lookahead Decoding offer substantial improvements over the baseline (68.27s latency, 120 tokens/sec). FlashAttention-2, by accelerating attention computations crucial for ATLAS, reduced latency to 29.55s (\textbf{↓ 56.7\%}) and increased throughput to 208 tokens/sec (\textbf{↑ 73.3\%}). Lookahead Decoding achieved further gains through parallel token prediction, decreasing latency to 23.15s (\textbf{↓ 66.1\%}) and boosting throughput to 255 tokens/sec (\textbf{↑ 112.5\%}). These results confirm that incorporating optimized decoding methods significantly enhances the responsiveness of our RAG system by speeding up both retrieval and generation phases, complementing the quality enhancements provided by PORAG+ATLAS.

% The table definition remains the same as you provided
\begin{table*}[ht!]
\centering
\caption{Latency and Throughput Improvements with Low-Latency Decoding Strategies}
\label{tab:latency_comparison}
\begin{tabular}{lcc}
\toprule
\textbf{Method} & \textbf{Avg. Latency (Sec/query)} & \textbf{Throughput (tokens/Sec)} \\
\midrule
ATLAS+RAG (Baseline)         & 68.27  & 120  \\
FlashAttention-2              & 29.55 (\textbf{↓ 56.7\%}) & 208 (\textbf{↑ 73.3\%}) \\
Lookahead Decoding            & 23.15 (\textbf{↓ 66.1\%}) & 255 (\textbf{↑ 112.5\%}) \\
\bottomrule
\end{tabular}
\end{table*}

\section{Related Work}

\subsection{Retrieval-Augmented Generation (RAG)}  
Advances in Retrieval-Augmented Generation (RAG) continue to extend the capabilities of Large Language Models (LLMs) in domain adaptation, efficiency, and long-context reasoning. RAFT \cite{zhang2024raft} improves factual accuracy by fine-tuning models to ignore irrelevant retrievals and cite only the most pertinent sources. CoRAG \cite{wang2025chain} enhances multi-hop reasoning through iterative retrieval, refining queries based on intermediate results rather than relying on a single retrieval step. DRAGIN \cite{su2403dragin} introduces dynamic retrieval by detecting real-time information needs using model uncertainty and self-attention cues, enabling context-sensitive query formulation during generation. RAPID \cite{chen2025longcontextinferenceretrievalaugmentedspeculative} accelerates long-context inference by combining RAG with speculative decoding, where a draft model predicts outputs for a larger model, balancing speed and accuracy through self- or upward-speculation. MemoRAG \cite{qian2024memorag} integrates external retrieval with a cognitive memory system, recording episodic interactions and distilling them into semantic memory to improve retrieval relevance and consistency. Speculative RAG \cite{wang2024speculative} reduces latency and enhances comprehension by generating draft responses using a small model and verifying them with a larger model. CAG \cite{chan2024don} addresses retrieval latency by preloading cached documents into extended context windows, bypassing real-time retrieval altogether. Parametric RAG \cite{su2025parametric} replaces input-context retrieval with document parameterization, temporarily updating LLM weights during inference to embed external knowledge directly, thereby streamlining the retrieve-update-generate process.

\subsection{Test-Time or Inference-Time Compute}  
Recent research has significantly advanced the reasoning capabilities of Large Language Models (LLMs) through innovative test-time computation scaling strategies. S1 \cite{muennighoff2025s1} introduces budget forcing, a prompting strategy that delays early conclusions by inserting ``Wait” tokens, encouraging longer and more deliberate reasoning. SETS \cite{chen2025sets} improves output quality through a cycle of sampling, self-verification, and self-correction, iteratively refining responses until correctness or a termination condition is met. Test-Time Computing (TTC) \cite{ji2025test} enables adaptive reasoning by combining a fast initial response with conditionally triggered refinement, emulating a shift from intuitive to deliberative thinking. Knockout and League \cite{chen2024simple} propose decision-time algorithms that reduce failure rates by comparing or averaging multiple candidate solutions. Marco-o1 \cite{zhao2024marco} combines Chain-of-Thought fine-tuning with Monte Carlo Tree Search (MCTS) to explore diverse reasoning paths for complex problem-solving, while STILL-1 \cite{jiang2024technical} integrates a policy and reward model to guide reasoning through a dynamically expanding tree. The Shortest Majority Vote \cite{zeng2025revisiting} leverages parallel CoT sampling with CoT-length-aware aggregation to scale inference, and ARMAP \cite{chen2025scaling} learns a reward model directly from environment interactions to guide LLM-based agents in evaluating action trajectories and improving planning. \cite{liu2025can} demonstrate that small LLMs can outperform much larger ones by optimizing the test-time scaling of policy models and reward-guided inference. \cite{yoon2025monte} extend this idea through Monte Carlo Tree Diffusion, combining diffusion models with MCTS to support iterative, tree-structured planning. Similarly, \cite{yu2025generating} propose translating LLM outputs into symbolic PDDL representations to enable classical planning with $A^{\star}$, leveraging best-of-N sampling and verbalized refinement. \cite{geiping2025scaling} present a recurrent depth architecture that scales compute within hidden states to deepen reasoning dynamically. \cite{wu2025boosting} introduce AStar, an MCTS-powered structured reasoning method for multimodal tasks, while \cite{lin2025qlass} propose QLASS, a Q-value-guided stepwise inference framework that enhances reasoning by modeling intermediate decision quality via a reasoning tree. Together, these works highlight a shift toward leveraging structured search, symbolic abstraction, and latent computation for efficient and scalable reasoning.

\subsection{KV Caching}  
Recent advancements in KV cache management have significantly enhanced the efficiency of Large Language Model (LLM) inference. Efficient inference requires effective management of the Key-Value (KV) cache, which stores intermediate computations during generation. Adaptive and prompt-guided strategies include Ada-KV \cite{feng2024ada}, which dynamically distributes compression budgets across attention heads based on their attention patterns, improving memory usage while maintaining generation quality. FINCH \cite{corallo2024finch} proposes a prompt-guided compression strategy that leverages pre-trained self-attention weights to iteratively select the most relevant KV pairs, enabling longer-context processing without requiring fine-tuning.  
For redundancy reduction, ThinK \cite{xu2024think} introduces a query-dependent pruning strategy that identifies and removes less significant channels within the key cache, minimizing memory consumption without compromising model performance. SimLayerKV \cite{zhang2024simlayerkv} focuses on inter-layer redundancies by detecting ``lazy" layers—those contributing minimally to long-range dependencies—and selectively trimming their KV caches. This approach streamlines memory usage by eliminating unnecessary data storage. Novel mechanisms for long-context inference include DuoAttention \cite{xiao2024duoattention}, which separates attention heads into Retrieval Heads (accessing the full KV cache for global context) and Streaming Heads (operating with a constant-length cache focused on recent tokens). This selective caching reduces memory and latency while preserving the model's ability to handle long contexts. Similarly, SnapKV \cite{li2025snapkv} exploits the observation that attention heads consistently focus on specific prompt features by clustering and retaining only the most relevant KV positions. This strategy improves efficiency while maintaining model performance.  
Recent works have proposed efficient strategies for compressing KV caches to support long-context inference in large language models. One approach, $L_{2}$-Norm-Based Pruning \cite{devoto2024simple}, leverages the observed correlation between the $L_{2}$ norm of key embeddings and their attention scores, selectively retaining KV pairs with the lowest norms to reduce memory usage without sacrificing performance. Another line of work, KVQuant \cite{hooper2025kvquant}, applies advanced quantization techniques—including per-channel and pre-RoPE key quantization, non-uniform precision, and sparse-dense vector representations—to compress KV caches to ultra-low bitwidths. These methods enable scalable inference over extended context lengths while maintaining model fidelity. KVLink \cite{yang2025kvlink} enhances LLMs by precomputing key-value (KV) caches for individual documents, allowing for efficient reuse during inference and reducing redundant computations. To ensure coherence when combining these precomputed caches, KVLink adjusts positional embeddings to reflect their global positions, introduces trainable special tokens to restore self-attention mechanisms across documents, and employs mixed-data fine-tuning to maintain the model's original capabilities. Together, these advancements collectively optimize memory usage, processing speed, and inference efficiency in LLMs. They highlight a growing emphasis on adaptive, redundancy-aware, and context-sensitive strategies for KV cache management, paving the way for more efficient and scalable LLM inference.

\end{document}